\documentclass[11pt,letterpaper]{article}
\usepackage{booktabs,graphicx,longtable,wrapfig,placeins,needspace}
\usepackage{preprint}
\definecolor{epitotal}{HTML}{C62828}
\definecolor{epivisible}{HTML}{7B1FA2}
\definecolor{epirem}{HTML}{188038}
\definecolor{epiale}{HTML}{1565C0}
\definecolor{eploss}{HTML}{455A64}
\newif\ifcolorcoding \colorcodingtrue \newcommand{\ctot}[1]{\ifcolorcoding{\color{epitotal}#1}\else{#1}\fi}
\newcommand{\cvis}[1]{\ifcolorcoding{\color{epivisible}#1}\else{#1}\fi}
\newcommand{\cinv}[1]{\ifcolorcoding{\color{epirem}#1}\else{#1}\fi}
\newcommand{\cale}[1]{\ifcolorcoding{\color{epiale}#1}\else{#1}\fi}
\newcommand{\closs}[1]{\ifcolorcoding{\color{eploss}#1}\else{#1}\fi}
\hypersetup{pdftitle={Scaling Laws of Ensemble Uncertainty in Language Models},pdfauthor={Siamak Ravanbakhsh, David Nitchi, Khang Ngo}}
\newcommand{\KL}{\operatorname{KL}}
\newcommand{\EE}{\mathbb{E}}
\newcommand{\calE}{\mathcal E}
\title{Scaling Laws of Ensemble\\Uncertainty in Language Models}
\author{Siamak Ravanbakhsh, David Nitchi, Khang Ngo\\[2pt]
  \textmd{McGill University and Mila}\\[2pt]
  \textnormal{\texttt{\{siamak.ravanbakhsh,david.nitchi,khang.ngo\}@mila.quebec}}}
\date{}

\begin{document}
\begin{preprintabstract} 
Many use cases for large language models (LLMs)---from AI safety to exploration and human interaction---can benefit from epistemic uncertainty estimates.
We argue that existing Bayesian Deep Learning solutions, which are often restricted to a fixed model family, fail to capture the full epistemic uncertainty at practical LLM scales. This is because the error due to the bounded capacity of the model family remains invisible to disagreement among models supported by the prior. We relate this error to the well-known power-law scaling of LLM loss with model size.
We then show that a second power law quantifies the shrinking disagreement within ensembles as model size increases. Its slower decay relative to the capacity law above suggests that ensembles capture a growing share of epistemic uncertainty and become more competitive as an axis for scaling, compared to a single larger model.\\
Finally, a third power law quantifies the loss of ensemble diversity as members share a larger fraction of training before branching, and therefore identifies a further limitation, beyond the capacity error, for LoRA ensembles and optimization-based or second-order methods that approximate uncertainty around a single training solution.\\
At a high level, these findings suggest diverse ensembles can be a promising axis, both for scaling and uncertainty quantification at larger scales.

\end{preprintabstract}

\section{Introduction}

Epistemic uncertainty reflects a model's gap in knowledge that more data or higher capacity could reduce. Bayesian inference represents such uncertainty through a distribution over plausible models. This uncertainty can guide the selection of informative labels in active learning \citep{houlsby2011}, experimental design \citep{rainforth2024}, and autonomous materials discovery \citep{kusne2020}. Recognizing knowledge gaps can also guide safe exploration and decisions about when to seek human oversight under distribution shift \citep{amodei2016}, while uncertainty about human preferences can incentivize an agent to defer to human intervention \citep{hadfieldmenell2017,russell2019}.
 
Many Bayesian Deep Learning (BDL) methods, including variational inference, Monte Carlo dropout, Laplace approximations, optimization-based methods, and many variations of ensembles \citep{graves2011,dropout,laplace,swag,de}, approximate a distribution over weights within a fixed architecture. As a result, when optimal model(s) sit outside this family, their reported uncertainties become quite unreliable. The loss difference between the best prediction available in a family and the ``true'' predictor is the approximation error or capacity loss. This loss remains \emph{invisible} to the BDL method as it lies outside the prior's support. We distinguish the \cinv{invisible residual} from the portion accessible or \cvis{\emph{visible}} to BDL methods (Figure~\ref{fig:family-cartoon}).
Empirical language-model loss laws include a size-dependent term \citep{kaplan,chinchilla,bahri} that one could loosely identify as this capacity loss or the invisible portion.\footnote{
The power law with model size could be attributed to either visible plus invisible (i.e., the total reducible loss), or just the invisible portion. At the scale of our experiments, these two ways of modelling the loss curves remain basically indistinguishable; see Appendix~\ref{sec:scaling-comparison} and Figure~\ref{fig:scaling-law-comparison}.}

\begin{preprintfigure}[tbp]
\centering
\begin{minipage}[c]{.37\linewidth}
\centering
\includegraphics[width=\linewidth]{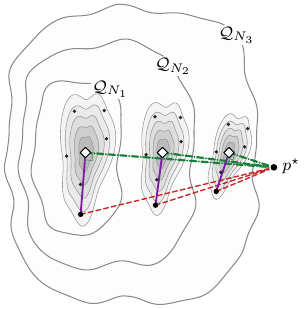}
\end{minipage}\hfill
\begin{minipage}[c]{.605\linewidth}
\captionsetup{font=footnotesize,justification=raggedright,singlelinecheck=false,skip=0pt}
\caption[Visible and invisible epistemic uncertainty]{\textbf{Visible and invisible epistemic uncertainty.}
The schematic shows increasingly expressive model families approaching the true next-token predictor $p^\star$.
Shaded contours are Bayesian posterior distributions; dots are ensemble members and diamonds their pooled predictions (which may not belong to the same family).
\textcolor{epitotal}{\textbf{Red dashed edges}} show total epistemic loss above the irreducible loss that decreases with a well-known power law with model size.
\textcolor{epivisible}{\textbf{Purple solid edges}} show the visible portion recovered by ensembling, which we find decreases rather slowly as a power law with model size.
\textcolor{epirem}{\textbf{Green dash--dot edges}} show the invisible portion, including errors shared by the members.
Their different decay rates imply the ratio purple-to-green increases with scale, and ensembles capture a bigger share of epistemic uncertainty.
We investigate this phenomenon and its implications.
Squared edge lengths illustrate an additive loss split.
}
\label{fig:family-cartoon}
\end{minipage}
\end{preprintfigure}

Now for the visible portion of the error or uncertainty, even under misspecification, Bayesian posteriors can concentrate around the best supported predictor, with posterior variance shrinking with $D^{-1}$ for $D$ independent observations \citep{kleijn2012}. However, this concentration result requires some assumptions that are violated in neural networks due to  symmetries and redundant parametrizations \citep{watanabe2013}. In a related work, \citet{rosso} study uncertainty scaling across data and model size, and report power-law trends in vision and weak data dependence in language-model fine-tuning. We posit that their results should be interpreted, not as scaling laws of \ctot{epistemic uncertainty}, but those of \cvis{visible portions}. In this paper we investigate the behaviour of these visible and invisible portions of the uncertainty in large language models (LLMs) using deep ensembles, and identify several implications. Concretely, we investigate:

\textbf{1. Scaling laws for the ensemble disagreement, in terms of data, model and context size.}
We empirically report a shallow power law for ensemble spread (the visible portion) across a wide range of open-weight models and our own family of character level models: the disagreement scales as $N^{-\gamma}$ with $\gamma\approx0.08$--$0.20$, where $N$ denotes model parameter count; see Figures~\ref{fig:champion}b and~\ref{fig:fresh-data-spread}. Figure~\ref{fig:visible} shows that this \cvis{ensemble spread} decreases more slowly than the \cinv{invisible portion} of epistemic loss. We compare this relationship across different ways of ensembling using geometric vs arithmetic pooling, training-token budgets and context length; see Figures~\ref{fig:fresh-data-spread} and~\ref{fig:pooling}. Beyond moderate training budgets, the visible portion changes little (Figure~\ref{fig:fresh-data-spread}a). Longer-context dependence is also weak beyond some moderate length (Figure~\ref{fig:fresh-data-spread}b; Appendix~\ref{sec:context-details}).

\textbf{2. The cost-benefits of ensembling at scale.}
Separate power laws for the visible and invisible portions predict a growing \emph{visible fraction} with model size 
 (Figure~\ref{fig:visible}). 
 The uncertainty is in part due to uncertainty in the irreducible-loss. The prediction, which is corroborated by similar trends we find using reference-to-member KL to estimate reducible loss (Figure~\ref{fig:noise-validation}), suggests that ensembling becomes more effective in capturing a growing portion of epistemic uncertainty at scale.
Another implication of this scaling law is that beyond uncertainty estimation,
with a fixed total number of parameters and fixed dataset sizes, ensembling becomes competitive with having a larger model at larger scales in terms of its loss; see Figure~\ref{fig:economics}.

\textbf{3. Scaling law for ensembling through late branching.}
A cost-effective approach to creating ensembles is late branching, where several models continue training starting with different orderings of training data, starting from the same partially trained model. The question is whether it confers similar benefits
as ``proper'' ensembles that have a completely separate history.
Our results suggest that later branching significantly reduces the seed spread, and does so in a predictable way, characterized by a simple power-law;
see Figure~\ref{fig:branching-law}. One implication is that ensembling techniques that produce variations around a single seed later in the training are significantly less diverse; this includes LoRA ensembles \citep{loraensembles}, or methods that approximate the posterior using the final portions of the training trajectory \citep{swag}. 

\begin{preprintfigure}[tp]
\centering\includegraphics[width=\linewidth]{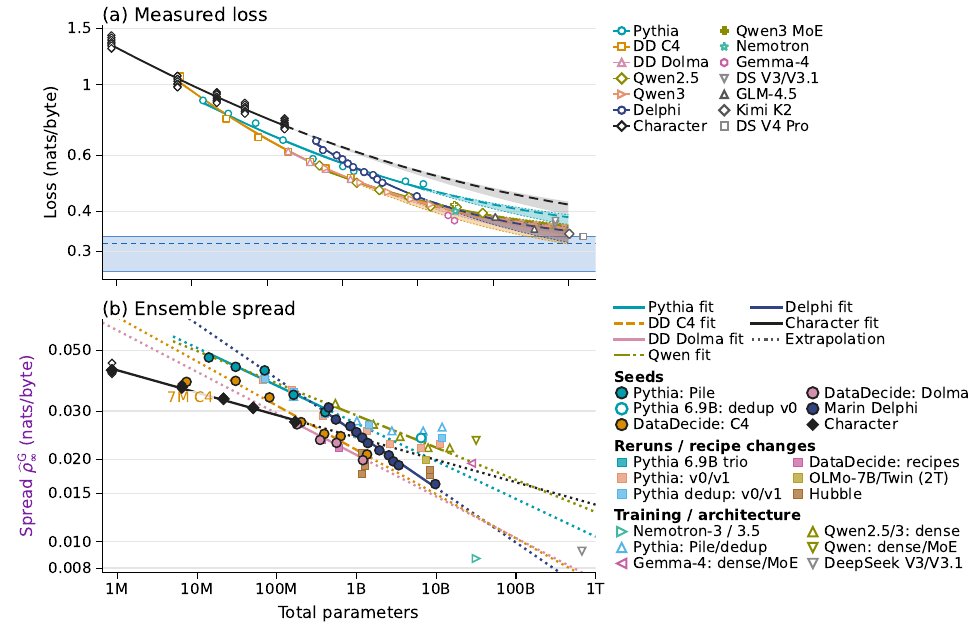}
\caption{\textbf{\closs{Model loss} and \cvis{ensemble spread} versus size.} (a) \closs{Measured losses} of 77 public and 98 character-level checkpoints on the same recent-text passages. The fits use $L(N)=E+A(N/10^9)^{-\eta}$ at \cale{$E=0.31727$} nats/byte. Colored bands refit the laws over the \cale{floor range $[0.26019,0.33373]$} (blue). (b) Corrected \cvis{geometric spread} for 72 public-model comparisons and 35 character-model settings, with fits $\widehat\rho_\infty^{\mathrm G}(N)=c(N/10^9)^{-\gamma}$. Dotted curves extrapolate. Fitted exponents $(\eta,\gamma)$ are Pythia $(0.191,0.138)$, DataDecide C4 $(0.253,0.162)$, DataDecide Dolma $(0.246,0.152)$, Qwen2.5 $(0.222,0.118)$, Qwen3 $(0.247,0.118)$, Delphi $(0.313,0.200)$, and character models $(0.164,0.082)$.  The spread fits are independent of $E$.}
\label{fig:champion}
\end{preprintfigure}

\section{Relation Between Loss, Epistemic Uncertainty and Ensembles}
\label{sec:theory}
This section first establishes the relationship between loss and epistemic uncertainty and then divides the epistemic part of the loss into two components based on their accessibility to an ensemble. This leads us to use geometric spread as a measure of disagreement in an ensemble. The derivations of this section are not limited to seed ensembles and remain true for more diverse ensembling techniques.

\subsection{Epistemic uncertainty as reducible loss}
Using upper-case for random variables, let $X$ be a context and $Y$ its next token in a finite vocabulary $\mathcal V$, with true conditional distribution $p_x$ and model prediction $q_x$. The mean cross-entropy loss splits into a model-independent irreducible part and a reducible part that vanishes only for $q=p$:
\begin{equation}
 \closs{L(q)=\EE_{X,Y}[-\log q_X(Y)]}
 =\cale{\underbrace{\EE_X H(p_X)}_{\text{irreducible loss } E=H(Y\mid X)}}
 +\ctot{\underbrace{\EE_X\KL(p_X\Vert q_X)}_{\text{reducible loss } \calE(q)}}.
 \label{eq:reducible}
\end{equation}
For a fully calibrated model, the total loss equals mean predictive entropy $\EE_X H(q_X)$. To see this, 
note that full calibration means the true token distribution conditional on the prediction $q_X$ is $q_X$ itself, that is $\Pr(Y=\cdot\mid q_X)=q_X$ -- i.e., whenever the model assigns, say, $70\%$ to a token, that token actually appears next in $70\%$ of the contexts where the model made that exact prediction.
Conditioning on $q_X$ and taking expectation, and using $\Pr(Y=\cdot\mid q_X)=q_X$ gives
\begin{equation}
 L(q)
 =\EE_X\!\left[-\sum_y \Pr(Y=y\mid q_X)\log q_X(y)\right]
 \overset{\text{calibration}}{=}\EE_X H(q_X).
 \label{eq:calibrated}
\end{equation}
A calibrated model can still miss information; for example, the true unigram prediction, which always predicts the marginal token frequencies regardless of the context, is fully calibrated.
Next we identify this missing information as epistemic uncertainty in the form of a conditional mutual information. 
First observe that for a fully calibrated model, by definition of calibration, we have 
$$H(Y\mid q_X)=\EE_X\, H\big(\Pr(Y=\cdot\mid q_X)\big)\overset{\text{calibration}}{=}\EE_X H(q_X).$$
Substituting this in the definition of conditional mutual information we get
\[
 I(Y;X\mid q_X)=H(Y\mid q_X)-H(Y\mid X, q_X)
 =\EE_X H(q_X)-H(Y\mid X).
\]
where we used $H(Y\mid X, q_X) = H(Y\mid X)$ because $q_X$ is  a function of $X$. 
Rearranging gives the separation of mean predictive entropy into aleatoric and epistemic uncertainty \citep{broecker2009}:\footnote{Without calibration, reducible loss also includes a \textit{calibration error}:
 $\EE_X\KL(c(q_X)\Vert q_X)$, where $c(q_X)=\Pr(Y=\cdot\mid q_X)$.}
\begin{equation}
 \closs{\underbrace{\EE_X H(q_X)}_{\substack{\text{mean predictive entropy}\\ \text{total loss } L(q) \text{, when calibrated}}}}
 =\cale{\underbrace{H(Y\mid X)}_{\substack{\text{aleatoric uncertainty}\\  \text{irreducible loss } E}}}
 +\ctot{\underbrace{I(Y;X\mid q_X)}_{\substack{\text{epistemic uncertainty}\\ \text{reducible loss } \calE(q) \text{, when calibrated}}}}
 \label{eq:calibration}
\end{equation}

This gives us a clean interpretation of the loss in LLMs; for calibrated models, the reducible loss is epistemic, in the sense that
it is concerned with information between the context and the next token that our model fails to capture; see also \citep{deup2023,kotelevskii2025}. 

As to whether the calibration assumption is justified in practical settings, empirical studies report good calibration of pretrained language models in several settings \citep{desai2020calibration,kadavath2022,openai2023gpt4,spiess2025calibration,semcal2026}.
Our recalibration tests find held-out log-loss gains below $1\%$ of estimated reducible loss for the character-level models and below $0.34\%$ of original log loss across 45 Pythia, DataDecide, Qwen, and Delphi checkpoints (Appendix~\ref{sec:calibration-evidence}).

\subsection{The Visible and Invisible Parts of Epistemic Uncertainty in Ensembles}
\label{sec:ensembles}

\paragraph{Geometric and arithmetic spreads.}
Let $S\geq2$ denote the number of ensemble members  $q_1,\ldots,q_S$, and let $\closs{\bar L_S}=S^{-1}\sum_s L(q_s)$ be their average loss. We  consider two standard ways to pool over the members, using geometric and arithmetic pooling:
\begin{equation}
 q_{S,x}^{\mathrm A}(y)=\frac1S\sum_s q_{s,x}(y),\qquad
 q_{S,x}^{\mathrm G}(y)=\frac{1}{{Z_x}} {\exp\{S^{-1}\sum_s\log q_{s,x}(y)\}}.
 \label{eq:pools}
\end{equation}
Here $Z_x=\sum_{y\in\mathcal V}\exp\{S^{-1}\sum_s\log q_{s,x}(y)\}$ normalizes the geometric pool. Arithmetic pooling averages probabilities, while geometric pooling averages logits or log probabilities and renormalizes. For either choice $P\in\{\mathrm A,\mathrm G\}$, the \cvis{\emph{spread}} is the difference between the average loss of members and the ensemble loss: $\cvis{\rho_S^P}=\bar L_S-L(q_S^P)$. The \ctot{epistemic loss of a member} (on average) can be decomposed as the sum of \cvis{spread} plus the \cinv{reducible loss of the ensemble}:
\begin{equation}
 \ctot{\underbrace{\bar L_S-E}_{\text{epistemic loss for a single member}}}
 =\cvis{\underbrace{\bar L_S-L(q_S^P)}_{\substack{\text{spread } \rho_S^P \text{ (visible)}}}}
 +\cinv{\underbrace{L(q_S^P)-E.}_{\substack{\text{epistemic loss of the ensemble (invisible)}}}}
 \label{eq:recovery}
\end{equation}
This equality separates the visible portion from the invisible portion. We use  $f_S^P=\cvis{\rho_S^P}/\ctot{(\bar L_S-E)}\in[0,1]$ for the ``fraction'' of the members' epistemic uncertainty visible via the ensemble. 
While arithmetic pooling is used more often in practice, the geometric pooling is a closer conceptual counterpart to mean-pooling with Gaussian likelihoods in regression; with geometric pooling the spread equals the mean divergence from the pool to its members  \footnote{For either pooling rule, simply substituting the definition of the $L$ terms, we see that the spread is the reduction in divergence from the true conditional distribution: $\rho_S^P=S^{-1}\sum_s\EE_X\KL(p_X\Vert q_{s,X})-\EE_X\KL(p_X\Vert q_{S,X}^P)$.} \citep{wood2023}:
\begingroup
\setlength{\belowdisplayskip}{0pt}
\setlength{\belowdisplayshortskip}{0pt}
\begin{equation}\label{eq:geometric}
\cvis{\rho_S^{\mathrm G}}=S^{-1}\sum_s\EE_X\KL(q_{S,X}^{\mathrm G}\Vert q_{s,X}) \end{equation}
\endgroup
This geometric spread plays the role of a variance-of-means with Gaussian ensembling: it is zero when all members agree and equals the loss recovered by the ensemble; its value for a specific context is independent of the observed next token and depends only on the predictions. 
In contrast, for arithmetic pooling, the loss reduction depends on the observed next token.

\paragraph{Bayesian active learning by disagreement (BALD).} BALD provides a third, and most common measure of disagreement in the ensembles \citep{houlsby2011}:
\begingroup
\setlength{\belowdisplayskip}{0pt}
\setlength{\belowdisplayshortskip}{0pt}
\begin{equation}\label{eq:bald}
 \closs{\underbrace{\EE_X H(q_{S,X}^{\mathrm A})}_{\text{pooled predictive entropy}}}
 =\closs{\underbrace{\frac1S\sum_s\EE_X H(q_{s,X})}_{\text{mean member entropy}}}
 +\underbrace{\frac1S\sum_s\EE_X\KL(q_{s,X}\Vert q_{S,X}^{\mathrm A})}_{\text{ensemble disagreement } J_S\text{ (BALD)}}.
\end{equation}
\endgroup
One way to connect BALD to spread is to note that relative to the average member, arithmetic pooling lowers log loss by the spread and raises entropy by BALD. 
While in some settings BALD is interpreted as epistemic uncertainty, in the above equation, mean member entropy includes epistemic uncertainty shared by all members (as in capacity loss) and unlike the epistemic term in Equation~\eqref{eq:calibration}, BALD does not measure the true epistemic uncertainty.
Given the direct connection between spreads and epistemic loss in \eqref{eq:recovery}, 
our focus is mainly on this measure of ensemble diversity, and in particular on the geometric spread due to its simpler form and its connection to variance in ensembling for regression.
However, one could also see the similarity of BALD and geometric ensembling in their KL forms in \eqref{eq:bald} and~\eqref{eq:geometric}; for nearby predictions, BALD and geometric spread agree to leading quadratic order \citep{nielsen2014chi} and our evaluations show similar qualitative behaviour for all three measures.

\textbf{Putting different ensemble sizes on the same scale.}
In experiments we would like to normalize quantities to account for the ensemble size.
Variation around the mean of $S$ IID samples, divided by $S$, has an expectation of $(S-1)/S$ times the population variance. For nearby predictions, both spreads and BALD have leading quadratic variance terms \citep{oehlert1992delta,nielsen2014chi}. We therefore multiply these measures by $S/(S-1)$ and use the subscript $\infty$ for the approximate population estimate. For example, $\widehat\rho_\infty^{\mathrm G}=\frac{2}{1}\rho_2^{\mathrm G}$ for two members. Note that our ensembles do not generally have iid members, and the justification above does not exactly hold in practice.

\begin{preprintfigure}[t]
\centering
\includegraphics[width=.8\linewidth]{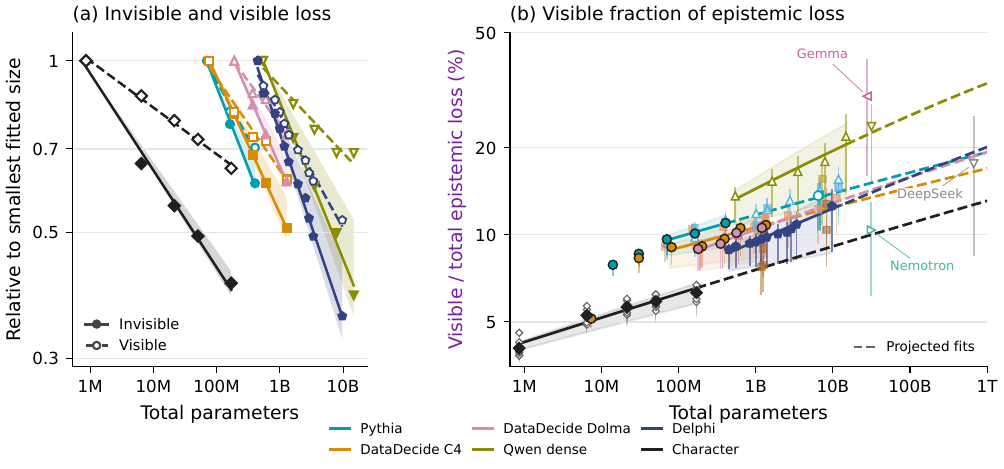}
\caption{\textbf{A larger portion of the epistemic uncertainty is visible at larger scales.} (a) Solid lines and filled markers show invisible loss; dashed lines and open markers show visible loss. Colors identify families. Spread and invisible-loss exponents $(\gamma,\alpha)$ are $(0.200,0.285)$ for Pythia, $(0.162,0.240)$ for DataDecide C4, $(0.152,0.256)$ for DataDecide Dolma, $(0.118,0.276)$ for dense Qwen, $(0.200,0.326)$ for Delphi, and $(0.082,0.170)$ for character models. Each component is normalized at its family's smallest fitted size. (b) Visible fractions for 72 public comparisons and 35 character settings on recent text. Uncertainty bands reflect the floor range.}
\label{fig:visible}
\end{preprintfigure}

\Needspace{19\baselineskip}
\section{The Empirical Observations and Their Implications}
\label{sec:results}

\subsection{The Ensembles Used in Experiments}
\label{sec:setup}

\begin{wrapfigure}{r}{.34\linewidth}
\begin{preprintfigureframe}
\centering
\includegraphics[width=\linewidth]{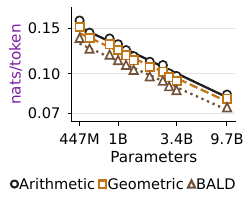}
\captionsetup{font=footnotesize,justification=raggedright,skip=3pt}
\caption{\textbf{Ensemble disagreement measures scale similarly in Delphi seed ensembles.}}
\label{fig:measures-representative}
\end{preprintfigureframe}
\end{wrapfigure}

We use existing open-weight models and a new character-level family, grouped by how closely ensemble members are related. Details and an explicit list of models are in Appendix~\ref{sec:model-families} and Table~\ref{tab:ensemble-families}.

\textbf{Independent seeds.} These are models that differ only in their seed; our experiments specifically use Pythia, DataDecide, and Delphi families, where DataDecide includes two families that differ in the choice of the datasets, C4 and Dolma.
Our character-level transformers also use independent seeds. For this family, we vary model size, training budget, and shared training history; see its training details in Appendix~\ref{sec:byte-details}.

\textbf{Release reruns and small recipe changes.} These keep the architecture fixed but allow small changes to the training recipe or corpus, as well as reruns across releases.

\textbf{Larger training or architecture changes.} These pair models of similar size with aligned token vocabularies, but allow larger changes in training data, training procedure, or architecture. Some also share early training. A family of ensembles that we use extensively from this group comprises pairs of similar-sized Qwen 2.5 and Qwen 3 models. This group also includes dense/MoE comparisons.

\FloatBarrier
\subsection{Scaling of Visible Epistemic Uncertainty with Model, Data and Context}\label{sec:seed-spread}

We use recent text (August 2026 Common Crawl) for the 72 public-model comparisons to reduce pretraining overlap when bounding irreducible loss. Appendix~\ref{sec:c4-validation} gives the corresponding results on C4 validation (2019 Common Crawl).

\textbf{Scaling with model size.} On recent text, fits of $\cvis{\widehat\rho_\infty^{\mathrm G}}\propto N^{-\gamma}$ give $\gamma=0.118$--$0.200$ across the fitted public-model families (Figure~\ref{fig:champion}b). The public-model loss exponents are $\eta=0.191$--$0.313$ at the shared floor (panel a). Appendix~\ref{sec:spread-fits} and Tables~\ref{tab:size-fits} and~\ref{tab:ensemble-measure-scaling} give the shared fitting protocol and family coefficients.
Our character-level models show the same shallow size trend on held-out FineWeb-Edu text, with exponent $\gamma=0.118$ when counting total parameters (Appendix~\ref{sec:byte-scaling}; Figure~\ref{fig:character-spread-invisible}). Figure~\ref{fig:visible}a shows that the \cinv{invisible component} decreases more sharply than the \cvis{spread} across families on recent text.
Geometric spread, arithmetic spread, and BALD show similar shallow size trends (Figure~\ref{fig:measures-representative}; Appendix~\ref{sec:measure-family-scaling}).

\begin{preprintfigure}[t]
\centering\includegraphics[width=\linewidth]{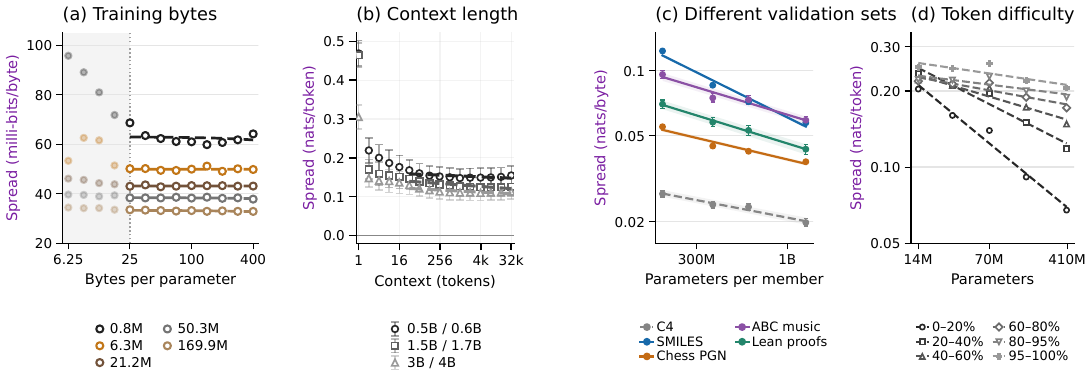}
\caption{\textbf{Spread by training budget, context, evaluation domain, and difficulty.} (a) Character-level model spread as we increase the dataset size, from 25 bytes per parameter onward, changes little. (b) Three Qwen2.5/Qwen3 pairs' spread changes are relatively small with context length beyond 16 tokens. (c) Corrected geometric spread for four DataDecide-Dolma sizes on C4 (gray) and four specialized/OOD domains, which show a higher spread but maintain a power-law. (d) Spread for tokens binned based on their difficulty (entropy), for Pythia. The spread resolves more slowly for more difficult tokens.}
\label{fig:fresh-data-spread}
\label{fig:spread-difficulty}
\end{preprintfigure}

\label{sec:data}
\textbf{Scaling with data.} Since for multi-seed open-weight model families we do not have clean checkpoints after cooldown and the learning rate varies during the training due to (e.g., cosine) scheduling, we cannot reliably measure spread with number of tokens in those model families. However, for our character-level family we could do so, and we see that beyond some initial phase up to 25 bytes per parameter the spread changes are small (Figure~\ref{fig:fresh-data-spread}a). Appendix~\ref{sec:byte-details} has the details.

\textbf{Scaling with context length.}
Context length is another axis along which language-model loss can decrease approximately as a power law  \citep{kaplan,xiong2024context}. Since our character-level model has a relatively short context length, we consider Pythia, DataDecide, Delphi, and Qwen families. Figure~\ref{fig:fresh-data-spread}b shows geometric spread for three Qwen2.5/Qwen3 base-model pairs on fixed targets as context grows to 32,768 tokens. After a substantial initial contraction, spread depends only weakly on additional context; see Appendix~\ref{sec:context-details} for details and similar results on BALD and arithmetic spread, as well as similar trends in Pythia, DataDecide, and Delphi ensembles.

\textbf{Change with dataset.} Figure~\ref{fig:fresh-data-spread}c shows that shifting the evaluation domain to specialized datasets, including ABC music, chemical SMILES, chess games, and Lean proofs, can change both the offset and the slope of the spread law. Ensemble spread still follows an approximate power law in model size across these domains. Appendix~\ref{sec:spread-fits} describes the evaluation sets and gives results for other model families in Figure~\ref{fig:ood-seed-spread}.

\textbf{Change of ensemble spread based on token difficulty.} Intuitively, the spread, which quantifies the disagreement within the ensemble, should depend on the difficulty of the next token.
We evaluate this hypothesis by binning tokens based on their difficulty level (prediction entropy) according to 
some model. 
We use a fixed reference model so that each difficulty bin contains the same tokens at every model size. Figure~\ref{fig:spread-difficulty}d shows the trend for Pythia seed ensembles: we see a power law trend where the decay exponent depends on the difficulty bin. Difficult positions account for a growing share of the spread, and their slow resolution with model size can intuitively explain the mild scaling of the spread; Appendix~\ref{sec:pooling-full-windows} has the details. 

\subsection{A Larger Share of Epistemic Uncertainty is Visible at Scale}
\label{sec:visible}

\begin{preprintfigure}[t]
\centering\includegraphics[width=\linewidth]{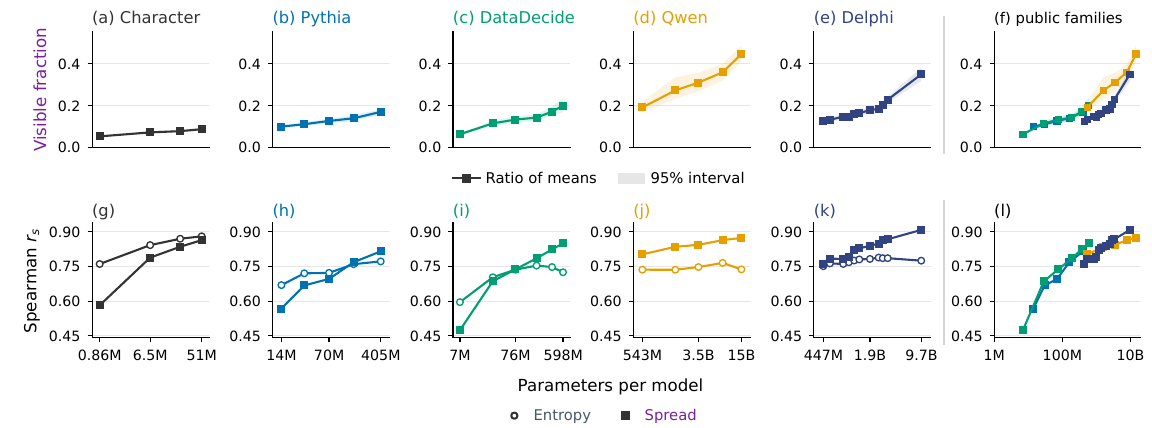}
\caption{\textbf{Context-dependent uncertainty across all families.} (a--e) The analogue of Figure~\ref{fig:visible}(b) using a reference-based estimate of the epistemic uncertainty for each context. Mean corrected \cvis{spread} divided by mean reference-to-member KL. Bands show $95\%$ intervals from resampling held-out windows. (g--k) Spearman correlations of single-model \closs{entropy} and ensemble \cvis{spread} with the reference-to-member KL estimate of reducible loss. (f,l) Visible fractions and spread correlations for the four public families on the same 40 recent-text passages. Character models use held-out FineWeb-Edu text. Within each family, all sizes use the same test contexts, vocabulary and reference.}
\label{fig:noise-validation}
\end{preprintfigure}

Given the three terms for visible, invisible and total epistemic loss, and two power-laws, an open question is which of these terms exactly have a power-law scaling.
A direct power law for total epistemic loss, $\ctot{\bar L_S-E}\simeq KN^{-\eta}$, combined with spread $\cvis{\rho_\infty^P}\simeq c_PN^{-\gamma_P}$ predicts a visible fraction which is their ratio $f_\infty^P\simeq(c_P/K)N^{\eta-\gamma_P}$. Since the former is steeper, according to our empirical results ($\eta>\gamma_P$), this extrapolation eventually exceeds $100\%$. 
Alternatively, separate power laws for the \cvis{visible} and \cinv{invisible} terms of Equation~\eqref{eq:recovery} give
\begin{equation}
 \cvis{\rho_\infty^P(N)}=c_PN^{-\gamma_P},\qquad \cinv{L(q_\infty^P)-E}=AN^{-\alpha},\qquad
 f_\infty^P(N)=\frac{\cvis{\rho_\infty^P(N)}}{\cinv{L(q_\infty^P)-E}+\cvis{\rho_\infty^P(N)}}.
 \label{eq:component-share}
\end{equation}

Here, the fraction stays below $100\%$, but the sum is generally not a power law.
The question becomes which one of these two alternatives better explains the data. Comparisons across several families of ensembles at scales we have seen so far give similar fit quality for the two approaches, with no consistent advantage (Appendix~\ref{sec:scaling-comparison} and in particular Figure~\ref{fig:scaling-law-comparison}). Identifying the correct model requires larger families, where predictions for the visible fraction clearly diverge because the second model saturates.
Figure~\ref{fig:visible}b shows the separate-component fits and their sensitivity to the irreducible loss. At the shared central floor, the invisible component decays faster than spread in all six fitted series, so the estimated visible fraction increases with size (Table~\ref{tab:component-share}).
These results suggest that at larger scales, the ensemble covers a larger fraction of the total epistemic uncertainty and ensembling becomes more effective in estimating the
uncertainty. However, depending on the floor estimate the slope for the increasing visible fraction can be small. 
Similar qualitative results appear in \citep{rosso,kirsch} in different settings, where they observe that ensemble disagreement decreases with scale; however, they do not consider the role of capacity and its steeper scaling law in this context. 
By doing so, we show that the recovered fraction of epistemic uncertainty using the ensemble grows with scale despite its shrinking spread. 
Appendix~\ref{sec:visible-details} has the details, limitations and additional results.

\textbf{The Visible Fraction Across Contexts.}
We estimate reducible loss at individual contexts by the forward KL from a fixed reference (Equation~\eqref{eq:reducible}). The references are BLT for character models, MPT-30B for Pythia, OLMo-7B for DataDecide, Qwen2.5-72B for Qwen, and Delphi 25B for Delphi.

The estimated \cvis{visible fraction} increases overall with size in all five families (Figure~\ref{fig:noise-validation}a--f). Its magnitude depends on the chosen reference. Figure~\ref{fig:visible}b instead measures loss above a common floor.

Figure~\ref{fig:noise-validation}g--k compares \closs{entropy} and \cvis{spread} as scores for ranking contexts by reference-based \ctot{reducible loss}. Appendix~\ref{sec:context-recovered-share} gives the common estimators.
At larger public-model sizes, \cvis{spread} also predicts reference-based \ctot{epistemic uncertainty} more effectively than \closs{entropy} or a hidden-state linear probe trained with Adam following \citet{ahdritz}; see Figure~\ref{fig:context-uncertainty-prediction} and Appendix~\ref{sec:uncertainty-protocol}.

\begin{preprintfigure}[t]
\centering
\includegraphics[width=\linewidth]{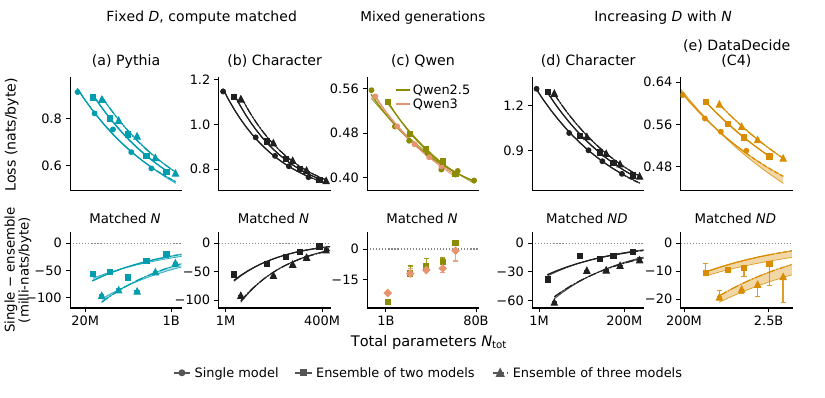}
\captionsetup{font=footnotesize,justification=raggedright,singlelinecheck=false}
\caption{\textbf{Single models, pairs, and triples across five settings.} Top: loss versus total parameters for single models and ensembles. Bottom: fitted single-model loss minus measured ensemble loss (markers), with predictions from separately fitted loss and spread laws (curves). Shading and vertical intervals show sensitivity to the corpus-specific floor ranges, with the loss laws refitted at each floor. (a,b) Pythia and character-level models match training data and total parameters. Character uses 8.493B bytes per model. (c) Qwen matches parameters but mixes generations with different training budgets. In (d,e), each gap compares the ensemble with a single model at the same total \emph{training compute cost} $ND$. Character models use FineWeb-Edu; public models use recent text. Delphi is excluded because its released checkpoints satisfy neither the fixed-data, matched-parameter constraint nor the matched-training-compute constraint. For out-of-distribution data we observe cross-overs; see Appendix Figure~\ref{fig:economics-specialized}}
\label{fig:economics}
\end{preprintfigure}

\subsection{Ensembling as an Alternative to a Larger Model}
\label{sec:ensemble-economics}
The fact that a growing fraction of reducible loss is explained by ensemble spread suggests that ensembling can provide a new axis for scaling at larger model sizes. 
Here, we compare different ensemble sizes at matched total parameters or training compute, treating a single model as an ensemble of size one.
For compute-optimal training, assuming (approximately) equal exponents $\alpha=\beta$ in the Chinchilla loss law $L(N,D)=E+aN^{-\alpha}+bD^{-\beta}$ and minimizing loss at a fixed training budget gives $D=\kappa N$, where $\kappa=(b/a)^{1/\alpha}$ \citep{chinchilla}. Since training compute is proportional to both model and data sizes $ND$, it grows as $N^2$ for compute-optimal training.
We can therefore compare an ensemble of $S$ models, each with $N$ parameters and trained on $D$ tokens, with a larger single model in two ways.

\textbf{(1) Matching total parameters and compute.} In this setting, we train a single model with $S N$ parameters on the same $D$ tokens. This also matches the ensemble's training and inference compute, to leading order for dense models.

\textbf{(2) Matching compute under compute-optimality.}  In this setting, both ensemble members and the larger model are assumed compute optimal, with the same total training compute. This means the larger model has $\sqrt{S}N$ parameters and it is trained on $\sqrt{S}D$ tokens. This model uses more data than each member, but fewer parameters than the total.

To see when ensembling is preferable, recall that its reduction in loss of its average member is exactly the \cvis{spread} (Equation~\eqref{eq:recovery}). The ensemble therefore outperforms the larger single model when this reduction exceeds the improvement obtained from the larger model:
\begin{equation}
 \cvis{\rho_S^P}>\closs{\bar L_S}-\closs{L(q_{\mathrm{big}})}.
 \label{eq:economics}
\end{equation}
We model the mean member loss by $L(N,D)$ and the spread by $\rho_S^P=(1-S^{-1})c_PN^{-\gamma_P}$, ignoring its weak dependence on training data beyond the initial regime (Figure~\ref{fig:fresh-data-spread}a). The following claim extends the comparison to any two fixed ensemble sizes under these two scaling laws, where $S=1$ corresponds to a single model. 
\begin{preprintclaim}
\setlength{\parskip}{2pt}
\textbf{Claim (An ensemble with more members outperforms at matched budget beyond some scale).}
Suppose the same Chinchilla and spread laws hold for two ensembles of sizes $1\le S_1<S_2$, with $a,b,c_P>0$ and $\alpha=\beta>\gamma_P>0$. Define $r=S_2/S_1$ and $u_i=1-S_i^{-1}$. Let each member of the $S_2$-member ensemble be compute optimal, with $N$ parameters and $D$ training tokens. Compare it with an $S_1$-member ensemble whose members use $(rN,D)$ in scenario~(1), matching training data and total parameters, or $(\sqrt r\,N,\sqrt r\,D)$ in scenario~(2), matching total training compute with compute-optimal members.

The $S_2$-member ensemble has lower loss in scenario~$1$ when $N>N_1^\star(S_1,S_2)$, and in scenario~$2$ when $N>N_2^\star(S_1,S_2)$, where
\begin{equation}
 \begin{aligned}
 N_1^\star(S_1,S_2)
 =\left[\frac{a(1-r^{-\alpha})}{c_P(u_2-u_1r^{-\gamma_P})}\right]^{1/(\alpha-\gamma_P)},
 N_2^\star(S_1,S_2)
 =\left[\frac{2a(1-r^{-\alpha/2})}{c_P(u_2-u_1r^{-\gamma_P/2})}\right]^{1/(\alpha-\gamma_P)}.
 \end{aligned}
 \label{eq:threshold}
\end{equation}
Both thresholds are finite, and $N_2^\star>N_1^\star$. Setting $S_1=1$ recovers the single-model comparison; setting $S_1=2,S_2=3$ gives the triplet--pair crossover. The proof is in Appendix~\ref{sec:ensemble-crossover-proof}.
\end{preprintclaim}
So for any two fixed ensemble sizes, the scaling laws predict a scale beyond which allocating the same budget to more ensemble members gives lower loss, provided the laws remain valid at that scale.
Figure~\ref{fig:economics} supports the predicted increase in competitiveness with model size. We see that the loss gap between ensembles and larger single models generally becomes smaller, both at fixed training data as in the first scenario (a,b), and when matching total training cost $ND$ as in the second scenario (d,e). Despite the trend, most ensembles still have higher loss than the larger model.
The bottom row curves are produced using two separately fitted laws: the gain from ensemble spread power law, and the improvement from a larger single model, based on Chinchilla. We fit a shared spread law to pairs and triples, then subtract the larger model's predicted improvement, matching parameters at fixed $D$ or total training cost $ND$. Qwen is left without a lower curve because its members come from different generations.
Appendix~\ref{sec:specialized-economics} presents analogous results on specialized data that are not in-distribution for some models (domains include music, math proofs, chemical SMILES and chess), where the spread can be several times higher and the crossover can be seen for different model families.  
Appendix~\ref{sec:implication-details} gives details, numerical crossover ranges, and additional results that show crossover instances observed in training checkpoints. 

These findings complement \citet{kim}, who show that ensembling
can outperform larger single language models under extensive data reuse and tuned regularization; see also \citep{lobacheva}. Our analysis derives a crossover
criterion from the relative scaling of loss and ensemble spread.
The criterion assumes our two scaling laws,
without requiring a particular regularization scheme, and, in spirit, its implications for ensembling also go against \citet{abe2022}, who argue that larger single models can match ensemble performance.

\begin{preprintfigure}[t]
\centering
\includegraphics[width=\linewidth]{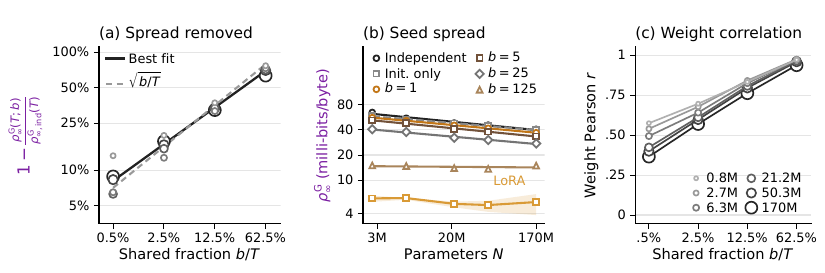}
\captionsetup{font=footnotesize,justification=raggedright,singlelinecheck=false,skip=4pt}
\caption{\textbf{Shared training reduces ensemble spread.} (a) Fraction of independent-seed spread lost versus the shared fraction of training, $b/T$, where $b$ is the branching point and $T$ the final training budget. The loss of spread follows an approximate square-root law. (b) Geometric spread versus model size for independent seeds, late branching, and LoRA ensembles. Panels (a,b) use recent text. (c) Pearson correlation between the complete final parameter vectors of parent and branch at $T=200$ bytes per analytic parameter. The size key applies to (a,c); (c) includes all six widths. Appendix~\ref{sec:branching-weight-correlation} gives calculation details and the corresponding spread comparison.}
\label{fig:branching-law}
\end{preprintfigure}

\subsection{Ensemble Spread with Late Branching and LoRA Ensembles}
\label{sec:branching}
Late branching reduces ensemble cost by separating ensemble-member training at a later checkpoint. Ensemble members will therefore have the same initialization and early updates. We first study this type of ensemble with full-parameter continuations, then with LoRA ensembles, where training continues on low-rank adaptors for the weights.

\paragraph{The spread follows an approximate square-root law.}
Using the FineWeb-Edu byte models, we branch after $b\in\{1,5,25,125\}$ bytes per parameter and evaluate at a final budget $T=200$. Different continuations from the same parent start from the parent's parameters but see the same remaining data in different orders; all parameters  are updated during training. Independent references for this section change both initialization and data order. \footnote{We see that compared to ensembles that differ only in initialization, changing the data order increases the spread slightly, on average only by about $4\%$ across the five fitted sizes.} 
For different branching points we measure how much of the spread is retained, compared to a reference with completely independent runs -- that is  \emph{retained share of spread} is
$
R(N,t,b)=\frac{\rho_{\infty,\mathrm{branch}}^{\mathrm G}(N,t,b)}{\rho_{\infty,\mathrm{ind}}^{\mathrm G}(N,t)}.
$
where $\rho_{\infty,\mathrm{branch}}$ and $\rho_{\infty,\mathrm{ind}}$ denote the late branching and reference ensemble spreads. 
The fitted law shown in Figure~\ref{fig:branching-law} suggests a clear power-law scaling for the share of spread that is lost (not retained):
\begin{equation}
 1-R(N,T,b)\simeq0.857\left(\frac bT\right)^{0.445}.
 \label{eq:branch-loss-law}
\end{equation}
For example, branching halfway through training is predicted to lose approximately $63\%$ of the independent-seed spread; extrapolating to branching with only $10\%$ of the run remaining gives a loss of approximately $82\%$. 
These results show that late branching substantially reduces ensemble diversity, highlighting a cost of sharing training history to construct ensembles efficiently \citep{wasay2020mothernets}. This power-law can also be used to predict the average spread for a true seed at a lower cost through late branching, before committing to a large multi-seed experiment. 
Figure~\ref{fig:branching-law}c shows that one way to estimate this diversity with model access is using weight correlations, which consistently show higher correlation with later branching (on a logarithmic scale) across model sizes; see also \citet{fort2019}. These observations about the connection between spread and late branching suggest a potential limitation for BDL methods that approximate the posterior using the final portions of the training trajectory \citep{swag} or more generally rely on modelling a loss ``basin''.
Appendix~\ref{sec:branching-details} gives the details and additional results.

\paragraph{Diversity Penalty.} One may try to increase the spread using a diversity penalty, as in negative correlation learning \citep{liu1999negative} and its geometric-pooling extension \citep{webb2021ensemble}. In Appendix~\ref{sec:diversity-branching} we report such an experiment, where we see that the diversity penalty indeed increases the spread at every model size, but this is at the cost of higher member losses, largely cancelling the additional gain in total ensemble loss. This means that while the ensemble now captures more uncertainty, this addition in uncertainty is mainly attributed to reduced performance of individual members, which makes the penalty pointless.

\paragraph{LoRA ensembles.}
LoRA ensembles \citep{loraensembles} and Bayesian adapters \citep{llora,bayeslora26,clora2025} also share a pretrained backbone, so we measure spread from independently trained adapters on a late checkpoint. 
Here, we compare the spread of LoRA in late branching to independent seeds, and see that LoRA pairs branched at $b=125$ have a further $2$ to $4$-fold reduction relative to full-parameter branches from the same checkpoint (Figure~\ref{fig:branching-law}b). Higher rank improves individual-model loss, but even at a high rank of $d/2$, less than half of full-branch spread is retained (Appendix~\ref{sec:lora-details}; Figure~\ref{fig:branching-lora}). 
\citet{muhlematter2026} report favourable diversity and calibration for LoRA ensembles in classification. However, their shared-backbone fine-tuning baseline differs from our independent-pretraining reference.

\section{Discussion and limitations}
Here we discuss the theoretical and empirical upshot from our study.
It is fair to say that many of its fragments are already present in the past literature, yet an aggregate picture and its implications, to our knowledge, have been missing.
On the theoretical side, we argue that reducible loss is the correct average measure 
of epistemic uncertainty for calibrated LLMs -- this calibration condition matters because, in Equation~\ref{eq:calibration}, the equality between the mutual information term (which has a clear epistemic interpretation) and the reducible loss depends on it. Moreover, we identify the spread, rather than BALD, and in particular the \cvis{geometric spread}, as a suitable measure of ensemble disagreement because of its direct connection to the epistemic loss. 

In this context, the existence of two clear scaling laws with distinct ranges for their exponents has important implications. As models grow, a growing fraction of \ctot{true epistemic uncertainty} for a given context becomes \cvis{visible} through ensemble diversity, and ensembling becomes competitive even for better prediction, independent of its use for uncertainty estimation.
Finally, a third power law suggests that cheap approximations through late branching and LoRA significantly reduce the spread. Our findings motivate training larger-scale seed ensembles to verify the power-law scaling of spread and also more extensive ensembling experiments to examine the late-branching power law in more diverse settings. 

The bulk of our theory and also the trends we find in experiments apply to arbitrary ensembles, beyond seed ensembles. What changes in ensembling models with different capabilities is that the mean member loss will no longer reflect the performance of an individual member, and the improvement over the mean by ensemble spread may not imply an improvement over the best ensemble member. The same argument applies both when considering estimates of epistemic uncertainty as reducible loss, and when comparing ensembling to single model scaling.

The most important limitation of our results and predictions is uncertainty in the \cale{irreducible loss}. 
However, note that this floor estimate \emph{only} affects the capacity law (i.e., Chinchilla scaling law), and it has no bearing on the two scaling laws we identify -- that is, the spread and the late branching power-laws. As long as the Chinchilla scaling exponent $\eta$ for the model size is above our estimated exponent $\gamma\approx 0.08$--$0.20$ for the spread law, our predicted trends stay valid. This and other limitations are further elaborated in the appendix.

An interesting theoretical direction for future work is the source of two scaling laws associated with ensembles: 1) while we relate the mild slope of the power-law 
scaling of the spread to slow resolution of difficult tokens, understanding the exact mechanism by which a power law is produced remains open;
 2) and
so does understanding the mechanism that produces the late-branching 
 power law of Figure~\ref{fig:branching-law}.

\label{page:main-end}
\pdfbookmark[1]{References}{preprint.references}
\bibliographystyle{plainnat}
\bibliography{references}
\clearpage
\pdfbookmark[1]{Appendix contents}{preprint.appendix}
\appendix
\setcounter{topnumber}{4}
\setcounter{bottomnumber}{3}
\setcounter{totalnumber}{6}
\renewcommand{\topfraction}{0.92}
\renewcommand{\bottomfraction}{0.85} 
\renewcommand{\textfraction}{0.08}
\renewcommand{\floatpagefraction}{0.70}
\makeatletter
\setlength{\@fptop}{0pt}
\setlength{\@fpsep}{12pt}
\setlength{\@fpbot}{0pt plus 1fil}
\makeatother
\raggedbottom

\section*{Appendix contents}
\begingroup
\small
\setlength{\tabcolsep}{5pt}
\renewcommand{\arraystretch}{1.25}
\noindent
\begin{tabular}{@{}p{.35\linewidth}p{.55\linewidth}r@{}}
\toprule
\textbf{Section} & \textbf{What is covered} & \textbf{Page} \\
\midrule
\raggedright\hyperref[sec:calibration-evidence]{\textbf{\ref*{sec:calibration-evidence}. \nameref*{sec:calibration-evidence}}}
& \raggedright Held-out recalibration gains for character models, Pythia, DataDecide, Qwen, and Delphi, plus confidence calibration and the limits of these tests.
& \pageref{sec:calibration-evidence} \\
\addlinespace[5pt]
\raggedright\hyperref[sec:experimental-details]{\textbf{\ref*{sec:experimental-details}. \nameref*{sec:experimental-details}}}
& \raggedright Details of model families used in experiments and seed selection; evaluation corpora and units; spread fits and sensitivity checks; character-model training and irreducible-loss references.
& \pageref{sec:experimental-details} \\
\addlinespace[5pt]
\raggedright\hyperref[sec:pooling-full-windows]{\textbf{\ref*{sec:pooling-full-windows}. \nameref*{sec:pooling-full-windows}}}
& \raggedright Geometric spread, arithmetic spread, and BALD; their size-law fits and comparisons within fixed entropy-based difficulty groups.
& \pageref{sec:pooling-full-windows} \\
\addlinespace[5pt]
\raggedright\hyperref[sec:context-details]{\textbf{\ref*{sec:context-details}. \nameref*{sec:context-details}}}
& \raggedright Context-length experiments with fixed targets for Pythia, DataDecide, and Qwen, including evaluation procedures, fitted trends, and variation across passages.
& \pageref{sec:context-details} \\
\addlinespace[5pt]
\raggedright\hyperref[sec:visible-details]{\textbf{\ref*{sec:visible-details}. \nameref*{sec:visible-details}}}
& \raggedright Separate visible and invisible loss laws, sensitivity of the visible fraction to the irreducible loss estimate, and comparison with a direct member-loss law.
& \pageref{sec:visible-details} \\
\addlinespace[5pt]
\raggedright\hyperref[sec:uncertainty-details]{\textbf{\ref*{sec:uncertainty-details}. \nameref*{sec:uncertainty-details}}}
& \raggedright Reference-based context-level targets, entropy and spread scores, hidden-state probe, prediction evaluation, and recovered fractions across contexts.
& \pageref{sec:uncertainty-details} \\
\addlinespace[5pt]
\raggedright\hyperref[sec:implication-details]{\textbf{\ref*{sec:implication-details}. \nameref*{sec:implication-details}}}
& \raggedright Proof of the crossover claim, loss fits under matched parameter and training-cost budgets, and measured DataDecide comparisons.
& \pageref{sec:implication-details} \\
\addlinespace[5pt]
\raggedright\hyperref[sec:branching-details]{\textbf{\ref*{sec:branching-details}. \nameref*{sec:branching-details}}}
& \raggedright Full-parameter branching procedures and fits; \hyperref[sec:diversity-branching]{diversity penalties and ensemble-loss tradeoffs (\ref*{sec:diversity-branching})}; LoRA training, ranks, spread, and prediction loss.
& \pageref{sec:branching-details} \\
\addlinespace[5pt]
\raggedright\hyperref[sec:c4-validation]{\textbf{\ref*{sec:c4-validation}. \nameref*{sec:c4-validation}}}
& \raggedright C4 counterparts to the main figures, including the loss floor, context-dependent uncertainty, and matched-budget comparisons.
& \pageref{sec:c4-validation} \\
\bottomrule
\end{tabular}
\endgroup
\clearpage

\section{Calibration}
\label{sec:calibration-evidence}
Section~\ref{sec:theory} relates reducible loss to epistemic uncertainty under calibration. We test for calibration error by fitting prediction-only corrections and measuring their reduction in held-out log loss. The language model stays fixed; maps receive no context or hidden states.

Let $c(q_X)=\Pr(Y=\cdot\mid q_X)$ and $C_{\rm cal}(q)=\EE\KL(c(q_X)\Vert q_X)$. A map $a$ that uses only the prediction satisfies
\[
 L(q)-L(a\circ q)=C_{\rm cal}(q)-\EE\KL(c(q_X)\Vert a(q_X))\leq C_{\rm cal}(q).
\]
Thus achieved gains lower-bound calibration error; small gains do not upper-bound the remaining error. We test character-model members and geometric pools, and individual public checkpoints. The character models' 256-byte vocabulary makes flexible full-distribution correction more tractable.

\subsection{Character-level models}
\label{sec:byte-calibration}
For the architecture in Appendix~\ref{sec:byte-training}, we fit each member and its finite geometric pool separately, without the population rescaling used for spread. Even- and odd-numbered held-out FineWeb-Edu sequences define fitting and scoring folds; their roles are exchanged and scoring results averaged.
 
Temperature scaling uses one scalar; vector scaling gives each log probability its own scale and offset; a residual multilayer perceptron (MLP) adds a learned correction before normalization. The MLP's zero-initialized output layer starts at the identity map. Table~\ref{tab:calibration-maps} gives fitting settings, Table~\ref{tab:calibration-refined} reports every evaluated setting, and Figure~\ref{fig:calibration-refined-full} summarizes size and data dependence.

\begin{table}[tbp]
\centering\small
\caption{\textbf{Recalibration maps using only predictions.} All maps minimize next-byte cross-entropy on the fitting fold. The MLP uses a GELU hidden activation.}
\label{tab:calibration-maps}
\begin{tabular}{@{}lll@{}}
\toprule
Map & Parameters & Fitting settings \\
\midrule
Temperature & 1 & Bounded minimization, $\log T\in[-2,2]$ \\
Vector & 512 & L-BFGS, at most 150 iterations; $w=1,b=0$ initially \\
MLP & 131,584 & 256 hidden units; input $0.1\max(\log q,-30)$ \\
 & & AdamW, 25 epochs, batch 8,192 \\
 & & Learning rate $10^{-3}$, cosine decay; weight decay $10^{-2}$ \\
\bottomrule
\end{tabular}
\end{table}

\begin{preprintfigure}[tbp]
\centering\includegraphics[width=\linewidth]{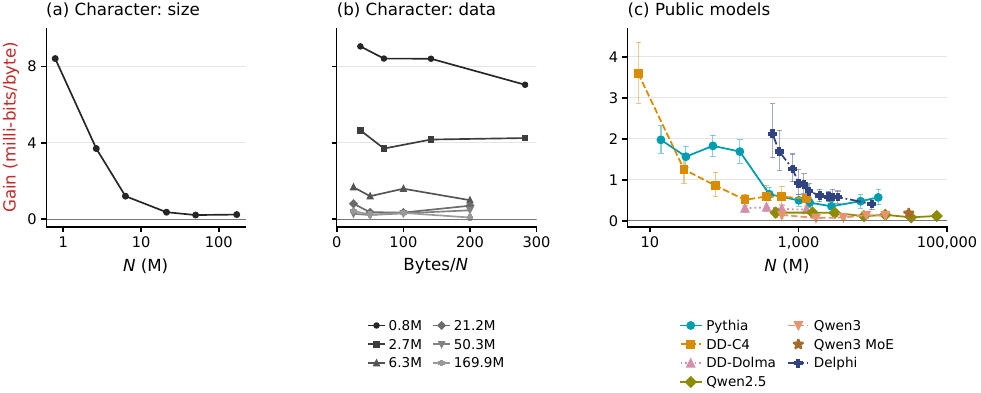}
\caption{\textbf{Recalibration across model and data size.} Held-out MLP gains in milli-bits/byte. (a) Character-model geometric pools at the second retained budget (50--71 bytes per analytic parameter). (b) All character settings at actual $D/N$; the legend gives analytic sizes. (c) Individual public checkpoints from Table~\ref{tab:public-full-recalibration}; DD denotes DataDecide and $N$ counts total parameters, including all MoE experts. Panel (c) uses a smaller vertical range; bars are conditional $95\%$ source-bootstrap intervals with fitted maps fixed. All gains average exchanged fitting/scoring folds. Public fitting and interval limitations are detailed in Appendix~\ref{sec:public-full-recalibration}.}
\label{fig:calibration-refined-full}
\end{preprintfigure}
\begin{table}[tbp]
\centering\small
\setlength{\tabcolsep}{5pt}
\caption{Calibration evaluation of our character-level models on 2,097,152 held-out bytes. $N=12hd^2$ is the analytic parameter count used for character-model scaling, with hidden width $d$ and $h=d/32$ layers. Each entry is a signed held-out loss reduction in milli-bits/byte. Columns T, V, and M denote temperature, vector, and MLP recalibration of the geometric ensemble; the member column averages separate MLP recalibrations. The final column divides the ensemble MLP gain by its reducible loss at $E=0.593341$ bits/byte. All 24 preselected cells are retained.}
\label{tab:calibration-refined}
\begin{tabular}{rrrrrrr}
\toprule
$N$ (M) & Bytes/$N$ & Ens. T & Ens. V & Ens. M & Member M & Gain / reducible loss \\
\midrule
0.79 & 35.4 & 0.154 & 5.452 & 9.047 & 8.677 & 0.671\% \\
0.79 & 70.7 & 0.465 & 5.253 & 8.409 & 8.232 & 0.652\% \\
0.79 & 141.4 & 0.251 & 5.370 & 8.401 & 8.029 & 0.677\% \\
0.79 & 282.8 & 0.280 & 4.197 & 7.038 & 7.174 & 0.588\% \\
2.65 & 35.4 & 0.466 & 3.358 & 4.656 & 4.335 & 0.425\% \\
2.65 & 70.7 & 0.288 & 2.466 & 3.700 & 3.468 & 0.365\% \\
2.65 & 141.4 & 0.161 & 2.973 & 4.172 & 4.148 & 0.432\% \\
2.65 & 282.8 & 0.074 & 2.837 & 4.250 & 4.379 & 0.454\% \\
6.29 & 25.0 & 0.212 & 1.522 & 1.677 & 1.420 & 0.179\% \\
6.29 & 50.0 & 0.371 & 1.603 & 1.210 & 0.990 & 0.138\% \\
6.29 & 100.0 & 0.159 & 1.539 & 1.598 & 1.401 & 0.192\% \\
6.29 & 200.0 & 0.346 & 1.374 & 0.998 & 0.830 & 0.126\% \\
21.23 & 25.0 & 0.140 & 0.879 & 0.812 & 0.641 & 0.107\% \\
21.23 & 50.0 & 0.234 & 0.676 & 0.370 & 0.272 & 0.052\% \\
21.23 & 100.0 & 0.154 & 0.603 & 0.352 & 0.195 & 0.052\% \\
21.23 & 200.0 & 0.105 & 0.850 & 0.709 & 0.579 & 0.111\% \\
50.33 & 25.0 & 0.174 & 0.581 & 0.285 & 0.178 & 0.044\% \\
50.33 & 50.0 & 0.193 & 0.510 & 0.221 & 0.124 & 0.037\% \\
50.33 & 100.0 & 0.128 & 0.475 & 0.310 & 0.220 & 0.055\% \\
50.33 & 200.0 & 0.075 & 0.564 & 0.470 & 0.381 & 0.087\% \\
169.87 & 25.0 & 0.083 & 0.544 & 0.431 & 0.379 & 0.083\% \\
169.87 & 50.0 & 0.128 & 0.434 & 0.246 & 0.175 & 0.051\% \\
169.87 & 100.0 & 0.072 & 0.421 & 0.333 & 0.279 & 0.074\% \\
169.87 & 200.0 & 0.121 & 0.276 & 0.091 & 0.052 & 0.022\% \\
\bottomrule
\end{tabular}
\end{table}

\subsection{A common recalibration test for public models}
\label{sec:public-recalibration}
We apply the same two-parameter map to all 45 public checkpoints in Table~\ref{tab:public-recalibration-gains}: ten Pythia, eleven DataDecide, thirteen Qwen, and eleven Delphi models. We use the released default checkpoint or seed 0 at each size. Let $k=\arg\max_y q(y)$, $p=q(k)$, and $g(p)=\sigma(a\operatorname{logit}(p)+b)$, with logistic function $\sigma$. The corrected distribution is
\[
 q'(k)=g(p),\qquad
 q'(y)=q(y)\frac{1-g(p)}{1-p}\quad(y\ne k).
\]
We minimize binary cross-entropy for $Y=k$ with $a\in[0.05,20]$ and $b\in[-10,10]$, using exchanged even/odd source-text folds. Probabilities stored as exactly one remain unchanged. For $z=\mathbf{1}\{Y=k\}$, the next-token log-loss reduction is
\[
 \log\frac{q'(Y)}{q(Y)}
 =z\log\frac{g(p)}p+(1-z)\log\frac{1-g(p)}{1-p}.
\]
Thus saved top-token probabilities and correctness indicators suffice to score the correction. All models use float32 inference on the same 96 recent-text sources (Appendix~\ref{sec:evaluation-corpora}): 92,160 predictions over 432,878 bytes for Pythia and DataDecide, 89,485 over 429,055 bytes for Qwen, and 87,683 over 431,488 bytes for Delphi. These counts differ with tokenization and scored-prefix truncation. This map preserves non-top relative probabilities and cannot detect their calibration errors.

\begin{table}[tbp]
\centering\small
\setlength{\tabcolsep}{4pt}
\caption{\textbf{A common recalibration test for public models.} Gains are signed held-out next-token log-loss reductions in milli-bits/byte from the two-parameter top-token mass map in Appendix~\ref{sec:public-recalibration}. The adjacent columns express the same gains as percentages of original log loss. Each model uses alternating source texts for fitting and scoring, then exchanges the folds. All 45 checkpoints and signed results are retained. C4 and Dolma name DataDecide training corpora; its labels count total parameters.}
\label{tab:public-recalibration-gains}
\begin{tabular}{@{}lrrlrr@{}}
\toprule
Model & Gain & \% loss & Model & Gain & \% loss \\
\midrule
Pythia-14M & +0.226 & +0.0176 & DataDecide C4 7.0M & +0.040 & +0.0026 \\
Pythia-31M & +0.014 & +0.0012 & DataDecide C4 28.8M & +0.046 & +0.0041 \\
Pythia-70M & +0.107 & +0.0098 & DataDecide C4 76.4M & +0.001 & +0.0001 \\
Pythia-160M & +0.004 & +0.0004 & DataDecide C4 190.5M & +0.016 & +0.0018 \\
Pythia-410M & +0.099 & +0.0118 & DataDecide C4 371.5M & +0.040 & +0.0049 \\
Pythia-1B & +0.150 & +0.0187 & DataDecide C4 597.7M & +0.060 & +0.0076 \\
Pythia-1.4B & +0.129 & +0.0167 & DataDecide C4 1279.9M & +0.051 & +0.0070 \\
Pythia-2.8B & +0.131 & +0.0178 & DataDecide Dolma 190.5M & +0.030 & +0.0034 \\
Pythia-6.9B & +0.209 & +0.0291 & DataDecide Dolma 371.5M & +0.041 & +0.0050 \\
Pythia-12B & +0.310 & +0.0440 & DataDecide Dolma 597.7M & +0.049 & +0.0063 \\
Qwen2.5-0.5B & +0.174 & +0.0216 & DataDecide Dolma 1279.9M & +0.031 & +0.0042 \\
Qwen2.5-1.5B & +0.132 & +0.0187 & Delphi-447M & +0.133 & +0.0139 \\
Qwen2.5-3B & +0.073 & +0.0108 & Delphi-550M & +0.125 & +0.0139 \\
Qwen2.5-7B & +0.070 & +0.0109 & Delphi-837M & +0.083 & +0.0096 \\
Qwen2.5-14B & +0.034 & +0.0056 & Delphi-998M & +0.144 & +0.0171 \\
Qwen2.5-32B & +0.044 & +0.0073 & Delphi-1.2B & +0.122 & +0.0150 \\
Qwen2.5-72B & +0.007 & +0.0011 & Delphi-1.4B & +0.070 & +0.0089 \\
Qwen3-0.6B & +0.054 & +0.0069 & Delphi-1.9B & +0.060 & +0.0078 \\
Qwen3-1.7B & +0.057 & +0.0080 & Delphi-2.5B & +0.035 & +0.0047 \\
Qwen3-4B & +0.015 & +0.0023 & Delphi-2.9B & +0.041 & +0.0056 \\
Qwen3-8B & +0.043 & +0.0069 & Delphi-3.4B & +0.045 & +0.0064 \\
Qwen3-14B & +0.021 & +0.0034 & Delphi-9.7B & -0.019 & -0.0029 \\
Qwen3-30B-A3B & +0.031 & +0.0052 &  &  &  \\
\bottomrule
\end{tabular}
\end{table}

\subsection{Full-distribution recalibration of public models}
\label{sec:public-full-recalibration}
We fit temperature, vector, and residual-MLP maps to the same checkpoints and score the same targets as above. Table~\ref{tab:public-full-recalibration} reports all signed gains; Figure~\ref{fig:calibration-refined-full}c plots the MLP results by size.

Each outer fold scores 48 untouched texts. Temperature uses either the other 48 ($T_{48}$) or 140 fitting texts ($T_{140}$), adding 92 separate texts. Flexible maps use 120 inner-fitting and 20 inner-validation texts (eight original and twelve additional). Validation selects identity, temperature, or the flexible map, its regularization, and stopping epoch; the selected recipe is refitted on all 140 texts. Vector and MLP are reported separately, without selection on scoring results.

Writing $\ell=\log q$ and $\beta=1/T$, the temperature map is $q_T=\operatorname{softmax}(\beta\ell)$, with $\beta\in[e^{-2},e^2]$. The vector map is
\[
 q_{\rm vec}=\operatorname{softmax}\!\left(\beta e^a\odot\ell+b-\bar b\right),
\]
initialized with $a=b=0$ at the fitted temperature. Its penalty is $\lambda(\|a\|_2^2+\|b-\bar b\|_2^2)/(2n_{\rm fit})$, with fitting-token count $n_{\rm fit}$ and $\lambda\in\{1,10,100\}$. The residual MLP is
\[
 q_{\rm MLP}=\operatorname{softmax}\!\left(\beta\ell+f_\theta\!\left(\frac{0.1\max(\ell,-30)}{\sqrt V}\right)\right),
\]
where $V$ is vocabulary size and $f_\theta$ has 32 GELU hidden units, initial first-layer weights multiplied by $\sqrt V$, and a zero-initialized output layer. Its penalty is $\lambda\,\EE\KL(q_{\rm MLP}\Vert q_T)$, with $\lambda\in\{0.1,1,10\}$. Both maps minimize penalized cross-entropy with $\beta$ fixed at its fitting-set estimate.

Both use learning rate $10^{-3}$, a 25-epoch cosine schedule, batches of 8,192 tokens accumulated from microbatches of 256, and seed zero. Vector scaling uses Adam without weight decay; the MLP uses AdamW with weight decay $10^{-2}$. Validation checks epochs $1,2,4,8,16,25$; refitting keeps the schedule and selected epoch.

Gradient fitting uses float16 log-probability caches converted to float32 and renormalized. Refined temperature fitting, validation, and scoring use original float32 predictions with float64 loss reductions; each gain compares corrections of the same predictions. Seven original Pythia $T_{48}$ evaluations instead reuse float16 archives; a four-text diagnostic against fresh float32 predictions provides no all-text error bound.

Conditional $95\%$ intervals use 1,000 whole-source bootstrap resamples (seed 2026110601), dividing resampled total gain by total bytes. They hold fitted maps fixed, exclude fitting and selection uncertainty, and are not simultaneous across checkpoints.

\begingroup\footnotesize
\setlength{\tabcolsep}{3pt}
\setlength{\LTcapwidth}{\linewidth}
\begin{longtable}{@{}lrrrrr@{}}
\caption{\textbf{Full-distribution recalibration of public models.} Signed held-out log-loss reductions in milli-bits/byte on the original 96 source texts. $T_{48}$ and $T_{140}$ use 48 and 140 fitting texts per outer fold, respectively; the latter adds 92 separate texts. Vector and MLP report the correction selected on 20 inner-validation texts, including identity and temperature as candidates, then refitted on all 140 fitting texts. The scoring fold is untouched. Vector and MLP cells give point gains and conditional $95\%$ intervals from whole-source bootstrap resampling with fitted maps fixed, excluding fitting and selection uncertainty; intervals are not simultaneous across models. $N$ counts total parameters in millions. Percentages reported in the text divide the same gains by original total log loss. Small achieved gains lower-bound full-distribution calibration error.}\label{tab:public-full-recalibration}\\
\toprule
Model & $N$ (M) & $T_{48}$ & $T_{140}$ & Vector & MLP \\
\midrule
\endfirsthead
\multicolumn{6}{l}{Table \thetable{} continued} \\
\toprule
Model & $N$ (M) & $T_{48}$ & $T_{140}$ & Vector & MLP \\
\midrule
\endhead
\midrule
\multicolumn{6}{r}{Continued on next page} \\
\endfoot
\bottomrule
\endlastfoot
\multicolumn{6}{l}{\textit{Pythia}} \\*
14M & 14.1 & -0.006 & -0.001 & +2.192 [+1.484, +2.896] & +1.975 [+1.652, +2.313] \\
31M & 30.5 & +0.113 & +0.110 & +2.144 [+1.533, +2.748] & +1.564 [+1.307, +1.817] \\
70M & 70.4 & -0.010 & -0.014 & +3.614 [+2.856, +4.312] & +1.828 [+1.569, +2.089] \\
160M & 162.3 & +0.414 & +0.407 & +2.753 [+2.139, +3.361] & +1.689 [+1.404, +1.977] \\
410M & 405.3 & +0.100 & +0.103 & +0.948 [+0.632, +1.254] & +0.647 [+0.484, +0.800] \\
1B & 1,011.8 & +0.073 & +0.089 & +0.643 [+0.404, +0.861] & +0.497 [+0.355, +0.632] \\
1.4B & 1,414.6 & +0.093 & +0.100 & +0.574 [+0.366, +0.772] & +0.441 [+0.319, +0.563] \\
2.8B & 2,775.2 & +0.080 & +0.082 & +0.699 [+0.427, +0.971] & +0.354 [+0.227, +0.477] \\
6.9B & 6,857.3 & +0.154 & +0.159 & +0.718 [+0.505, +0.929] & +0.476 [+0.303, +0.650] \\
12B & 11,846.1 & +0.265 & +0.263 & +0.784 [+0.567, +1.013] & +0.570 [+0.387, +0.764] \\
\addlinespace[3pt]
\multicolumn{6}{l}{\textit{DataDecide C4}} \\*
4M & 7.0 & +0.016 & -0.013 & +2.311 [+1.225, +3.352] & +3.594 [+2.871, +4.361] \\
20M & 28.8 & +0.050 & +0.047 & +1.055 [+0.314, +1.801] & +1.252 [+0.911, +1.631] \\
60M & 76.4 & +0.005 & +0.006 & +0.906 [+0.332, +1.471] & +0.856 [+0.582, +1.184] \\
150M & 190.5 & -0.006 & -0.001 & +0.819 [+0.271, +1.375] & +0.511 [+0.372, +0.643] \\
300M & 371.5 & -0.002 & -0.002 & +0.930 [+0.434, +1.429] & +0.594 [+0.367, +0.862] \\
530M & 597.7 & +0.001 & +0.014 & +1.029 [+0.562, +1.501] & +0.588 [+0.368, +0.846] \\
1B & 1,279.9 & +0.036 & +0.042 & +1.182 [+0.690, +1.680] & +0.541 [+0.341, +0.777] \\
\addlinespace[3pt]
\multicolumn{6}{l}{\textit{DataDecide Dolma}} \\*
150M & 190.5 & -0.029 & -0.011 & +0.317 [+0.156, +0.471] & +0.306 [+0.194, +0.416] \\
300M & 371.5 & -0.004 & -0.001 & +0.265 [+0.132, +0.391] & +0.327 [+0.217, +0.424] \\
530M & 597.7 & -0.005 & -0.001 & +0.273 [+0.054, +0.482] & +0.290 [+0.179, +0.394] \\
1B & 1,279.9 & -0.000 & +0.004 & +0.208 [+0.119, +0.300] & +0.270 [+0.176, +0.361] \\
\addlinespace[3pt]
\multicolumn{6}{l}{\textit{Qwen2.5}} \\*
0.5B & 494.0 & +0.072 & +0.080 & +0.233 [+0.090, +0.380] & +0.198 [+0.092, +0.307] \\
1.5B & 1,543.7 & +0.120 & +0.123 & +0.226 [+0.099, +0.344] & +0.199 [+0.112, +0.286] \\
3B & 3,085.9 & +0.110 & +0.110 & +0.243 [+0.113, +0.356] & +0.192 [+0.102, +0.270] \\
7B & 7,615.6 & +0.080 & +0.081 & +0.192 [+0.081, +0.298] & +0.111 [+0.054, +0.165] \\
14B & 14,770.0 & +0.045 & +0.042 & +0.187 [+0.081, +0.285] & +0.145 [+0.073, +0.216] \\
32B & 32,763.9 & +0.053 & +0.053 & +0.189 [+0.085, +0.281] & +0.084 [+0.031, +0.135] \\
72B & 72,706.2 & +0.021 & +0.023 & +0.124 [+0.024, +0.214] & +0.116 [+0.059, +0.171] \\
\addlinespace[3pt]
\multicolumn{6}{l}{\textit{Qwen3-Base}} \\*
0.6B & 596.0 & +0.045 & +0.044 & +0.169 [+0.071, +0.263] & +0.143 [+0.046, +0.236] \\
1.7B & 1,720.6 & +0.029 & +0.030 & +0.094 [+0.019, +0.165] & +0.068 [+0.002, +0.133] \\
4B & 4,022.5 & +0.040 & +0.040 & +0.151 [+0.074, +0.227] & +0.081 [+0.036, +0.125] \\
8B & 8,190.7 & +0.036 & +0.036 & +0.169 [+0.092, +0.242] & +0.136 [+0.067, +0.206] \\
14B & 14,768.3 & +0.053 & +0.053 & +0.162 [+0.087, +0.232] & +0.129 [+0.068, +0.196] \\
\addlinespace[3pt]
\multicolumn{6}{l}{\textit{Qwen3 MoE}} \\*
30B-A3B & 30,532.1 & +0.052 & +0.052 & +0.192 [+0.110, +0.270] & +0.180 [+0.098, +0.257] \\
\addlinespace[3pt]
\multicolumn{6}{l}{\textit{Delphi}} \\*
447M & 447.2 & +0.039 & +0.038 & +2.972 [+2.223, +3.787] & +2.123 [+1.540, +2.861] \\
550M & 550.3 & +0.008 & +0.017 & +2.625 [+1.988, +3.367] & +1.688 [+1.234, +2.207] \\
837M & 837.0 & +0.020 & +0.018 & +2.302 [+1.751, +2.916] & +1.264 [+0.956, +1.629] \\
998M & 998.0 & +0.004 & +0.013 & +1.978 [+1.450, +2.537] & +0.917 [+0.646, +1.244] \\
1.2B & 1,180.0 & +0.037 & +0.033 & +1.734 [+1.254, +2.248] & +0.885 [+0.667, +1.144] \\
1.4B & 1,384.6 & -0.009 & +0.003 & +1.606 [+1.177, +2.068] & +0.721 [+0.545, +0.899] \\
1.9B & 1,934.7 & +0.010 & +0.006 & +1.535 [+1.116, +1.976] & +0.615 [+0.463, +0.773] \\
2.5B & 2,544.6 & +0.022 & +0.018 & +1.551 [+1.147, +1.968] & +0.588 [+0.460, +0.721] \\
2.9B & 2,895.2 & +0.025 & +0.014 & +1.391 [+1.007, +1.793] & +0.580 [+0.399, +0.765] \\
3.4B & 3,383.1 & +0.006 & +0.004 & +1.390 [+1.019, +1.774] & +0.579 [+0.431, +0.726] \\
9.7B & 9,714.7 & -0.004 & -0.002 & +0.864 [+0.560, +1.155] & +0.409 [+0.265, +0.555] \\
\end{longtable}
\endgroup
\suppressfloats[t]

\subsubsection{Pythia models}
\label{sec:pythia-calibration}
We use the default release at final step 143,000. Table~\ref{tab:pythia-calibration-corrections} additionally retains an earlier seven-model study using the character models' 256-hidden-unit MLP, with separate validation selecting identity or a learning rate. The shared study above uses the regularized 32-hidden-unit map across all ten sizes.
\begin{table}[tbp]
\centering\small
\setlength{\tabcolsep}{4pt}
\caption{Corrections based only on Pythia's predictions. Gains are signed held-out loss reductions in milli-bits/byte. The MLP recipe used for our character-level models is retained even when it worsens loss. Nested validation chooses identity or a learning rate separately within each fitting fold; outer scoring text never chooses the correction. The last columns give that choice for each fold.}
\label{tab:pythia-calibration-corrections}
\begin{tabular}{rrrrcc}
\toprule
$N$ (M) & Temperature & Original MLP & Validated MLP & Fold 0 & Fold 1 \\
\midrule
14.1 & -0.006 & +0.528 & +0.528 & $10^{-3}$ & $10^{-3}$ \\
30.5 & +0.113 & -238.249 & +0.000 & Identity & Identity \\
70.4 & -0.010 & -245.615 & +0.000 & Identity & Identity \\
162.3 & +0.414 & +0.361 & +0.135 & Identity & $10^{-3}$ \\
405.3 & +0.100 & +0.203 & +0.000 & Identity & Identity \\
1011.8 & +0.073 & -191.980 & +0.000 & Identity & Identity \\
1414.6 & +0.093 & -90.873 & +0.000 & Identity & Identity \\
\bottomrule
\end{tabular}
\end{table}

\subsubsection{DataDecide models}
\label{sec:datadecide-calibration}
We use the default seed at the matched C4- and Dolma-trained checkpoints of the seed comparisons (Table~\ref{tab:primary-seeds}). Their steps are listed in Appendix~\ref{sec:seed-populations}; several precede the scheduled final step.

\subsubsection{Qwen models}
\label{sec:qwen-calibration}
We evaluate final Qwen2.5 and Qwen3 base models, including Qwen3-30B-A3B. The larger models use fresh float32 evaluations with directly recorded predicted token IDs.

\subsection{Confidence calibration}
For Pythia and Qwen, we also report top-token confidence ECE: the prediction-weighted mean absolute confidence--accuracy gap across 15 equal-width bins (Table~\ref{tab:calibration-confidence-summary}). Paired $95\%$ intervals resample the same 96 source texts 1,000 times, retaining all predictions from each text. Table~\ref{tab:qwen-calibration-changes} reports within-family Qwen comparisons. \begin{table}[tbp]
\centering\small
\caption{\textbf{Confidence calibration of public models.} Ranges give 15-bin expected calibration error (ECE), in percentage points, across final individual models. Pythia uses 92,160 predictions per model; Qwen uses 89,485 on the same 96 source texts. Sizes follow the model release names.}
\label{tab:calibration-confidence-summary}
\begin{tabular}{@{}lrlr@{}}
\toprule
Family & Models & Model sizes & ECE range \\
\midrule
Pythia & 10 & 14M--12B & 0.436--1.642 \\
Qwen2.5 dense & 7 & 0.5B--72B & 0.566--1.329 \\
Qwen3 dense & 5 & 0.6B--14B & 0.626--0.912 \\
Qwen3 MoE & 1 & 30B-A3B & 0.743 \\
\bottomrule
\end{tabular}
\end{table}

\begin{table}[tbp]
\centering\small
\setlength{\tabcolsep}{5pt}
\caption{Within-family changes in Qwen confidence ECE, larger minus smaller model, in percentage points. Ranges use paired resampling of the same 96 source texts; they are descriptive and do not adjust for multiple comparisons.}
\label{tab:qwen-calibration-changes}
\begin{tabular}{lrr}
\toprule
Comparison & Change & Paired 95\% range \\
\midrule
Qwen2.5-0.5B $\to$ Qwen2.5-14B & -0.689 & [-0.911, -0.319] \\
Qwen2.5-0.5B $\to$ Qwen2.5-72B & -0.764 & [-0.948, -0.396] \\
Qwen2.5-14B $\to$ Qwen2.5-72B & -0.074 & [-0.294, +0.157] \\
Qwen3-0.6B $\to$ Qwen3-14B & -0.247 & [-0.420, +0.107] \\
\bottomrule
\end{tabular}
\end{table}

\section{Models, evaluation, and size--data scaling}
\label{sec:experimental-details}
This appendix gives the evaluation and fitting details for Sections~\ref{sec:setup} and~\ref{sec:seed-spread}. Public-model sizes count all parameters. Unless stated otherwise, losses and spreads use natural logarithms (as in log-likelihood) divided by scored bytes (averaged over characters). We perform $S/(S-1)$ rescaling of relevant quantities following Section~\ref{sec:ensembles}.

\subsection{Ensemble families}
\label{sec:model-families}
\textbf{Independent seeds.}
We use independent-seed ensembles from Pythia (14M--410M, 6--10 retained members), DataDecide (7M--1.28B, three members), and Marin Delphi (447M--9.7B, three members at eleven sizes) \citep{pythia,polypythias,datadecide,held2026delphi,marin2026seeds}. Their training corpora are the Pile, a collection from 22 sources \citep{pile}; C4, cleaned web text \citep{raffel2020c4}, or Dolma, a mixture of web text, books, code, and other sources \citep{dolma}; and a shared web, code, and mathematics mixture, respectively. DataDecide uses the openly released OLMo transformer family \citep{olmo}. We also include a separate public 6.9B Pythia seed-labelled pair from the deduplicated-Pile v0 release.

\textbf{Release reruns and small recipe changes.}
We combine Pythia v0/v1 release pairs trained on the standard or deduplicated Pile, a 6.9B mixed-release trio, and five DataDecide training recipes at each size. We also include Hubble near-replicates, whose training text differs by about $0.02$--$0.08\%$ of tokens, and OLMo-7B paired with its separately trained Twin at a matched two-trillion-token budget. Their members span roughly 70M--12B total parameters.

\textbf{Larger training or architecture changes.}
We pair dense models of similar size across the Qwen2.5 and Qwen3 generations \citep{qwen3} to create ensembles, and also compare Pythia models trained on the standard and deduplicated Pile. We also include the Qwen2.5-32B/Qwen3-30B-A3B dense/mixture-of-experts (MoE) pair, separately from the five dense Qwen pairs. Our largest ensemble reaches 671B parameters per member with the DeepSeek-V3/V3.1 pairing \citep{deepseekv3}, whose members share early training; we also include Nemotron-3 Nano/3.5 Lightning \citep{nemotronnano,nemotronlightning}, and the Gemma-4 31B/26B-A4B dense/MoE pair \citep{gemma4} as ensembles. We pool aligned token distributions and score loss and spread per byte. Table~\ref{tab:ensemble-families} lists the 72 public-model comparisons in Figure~\ref{fig:champion}b.

\textbf{Our character-level model family.}
We train byte-level transformers on FineWeb-Edu, web text selected for educational quality \citep{fineweb}, with 0.86M--170M total parameters. The primary spread study uses five sizes at 50--400 bytes per analytic parameter, with two retained seeds at the smallest size and three at the others (Figure~\ref{fig:fresh-data-spread}a; Figure~\ref{fig:character-spread-invisible}). These analyses use held-out FineWeb-Edu text. Figure~\ref{fig:champion} evaluates the same checkpoints on recent text. Training our own models gives every data budget its own learning-rate cooldown; this is missing in existing public checkpoints -- e.g., Pythia's intermediate checkpoints have no separate cooldown for each budget. Also a vocabulary of 256 byte values permits sub-million-parameter models. This scale makes late-branching experiments possible within our compute budgets. We have carefully optimized the hyperparameters to achieve a clean scaling law for this model family. 
Appendix~\ref{sec:byte-details} gives the training procedure. 

\subsection{Pretrained seed populations}
\label{sec:seed-populations}
The primary seed comparisons share training text and allocations within each ensemble: the standard, non-deduplicated Pile for Pythia, C4 or Dolma for DataDecide, and a mixture of Nemotron-CC, StarCoderData, and ProofPile~2 for Delphi \citep{pythia,polypythias,datadecide,held2026delphi,marin2026seeds}. Table~\ref{tab:primary-seeds} lists every retained size and member population. Across sizes, Pythia uses a common 300B-token budget; DataDecide follows its released size-dependent schedule at matched steps within each seed comparison; and Delphi ranges from 1.22B to 160.37B training tokens. The size laws describe each family's observed allocation schedule, including joint size--data variation when training budgets grow with size. DataDecide's matched checkpoint steps, in increasing C4 size order, are 5,725, 14,584, 29,042, 37,500, 45,000, 51,250, and 69,369. Dolma uses the same steps at the four largest sizes; several checkpoints precede the scheduled final step.

We exclude a Pythia member if its recent-text loss exceeds the suite median by more than $0.01$ nats/byte. At every size, we evaluate the geometric pool of all retained members, including the six-member 14M pool and nine-member 31M pool. The measured finite-pool loss is $L(q_S^{\mathrm G})=\bar L_S-\rho_S^{\mathrm G}$; only the displayed spread receives the $S/(S-1)$ population rescaling. Filtering can reduce the observed variation; Appendix~\ref{sec:selection-sensitivity} tests its effect.

The separate large Pythia pair uses \href{https://huggingface.co/EleutherAI/pythia-6.9b-deduped-v0}{\texttt{pythia-6.9b-deduped-v0}} and its \href{https://huggingface.co/EleutherAI/pythia-6.9b-deduped-v0-seed42}{\texttt{seed42}} counterpart. Both belong to an older deduplicated-Pile release. The seed-42 checkpoint has no model card, and archived evaluations did not record revision hashes. We therefore treat this pair as a separate public comparison and exclude it from the primary size, difficulty, and recovered-share fits.

\begin{table}[tbp]
\centering
\caption{Recent-text seed measurements, including the separately labelled public 6.9B deduplicated-v0 pair. That pair is excluded from primary fits. Loss and corrected geometric spread are in nats/byte; fractions use the V4-Pro-anchored assumed floor $E=0.33372841$. Every row evaluates the full retained pool, including six members at 14M and nine at 31M. The finite-pool loss $L(q_S^{\mathrm G})$ is measured directly; only spread receives the $S/(S-1)$ correction.}
\label{tab:primary-seeds}
\small
\begin{tabular}{lrrrrrr}
\toprule
Family & $N$ (M) & $S$ & $\bar L_S$ & $L(q_S^{\mathrm G})$ & $\widehat\rho_\infty^{\mathrm G}$ & $\widehat f_\infty^{\mathrm G}$ \\
\midrule
Pythia & 14.1 & 6 & 0.91467 & 0.87543 & 0.04708 & 8.10\% \\
Pythia & 30.5 & 9 & 0.82338 & 0.78469 & 0.04352 & 8.89\% \\
Pythia & 70.4 & 10 & 0.75479 & 0.71678 & 0.04224 & 10.03\% \\
Pythia & 162.3 & 9 & 0.65751 & 0.62693 & 0.03440 & 10.62\% \\
Pythia & 405.3 & 8 & 0.58780 & 0.56178 & 0.02974 & 11.70\% \\
Pythia dedup v0 & 6857.3 & 2 & 0.49224 & 0.48028 & 0.02391 & 15.09\% \\
DataDecide C4 & 7.0 & 3 & 1.06419 & 1.03862 & 0.03836 & 5.25\% \\
DataDecide C4 & 28.8 & 3 & 0.78230 & 0.75651 & 0.03868 & 8.62\% \\
DataDecide C4 & 76.4 & 3 & 0.68822 & 0.66579 & 0.03364 & 9.49\% \\
DataDecide C4 & 190.5 & 3 & 0.61666 & 0.59844 & 0.02732 & 9.66\% \\
DataDecide C4 & 371.5 & 3 & 0.57274 & 0.55625 & 0.02474 & 10.35\% \\
DataDecide C4 & 597.7 & 3 & 0.54724 & 0.53105 & 0.02428 & 11.37\% \\
DataDecide C4 & 1279.9 & 3 & 0.50981 & 0.49594 & 0.02082 & 11.82\% \\
DataDecide Dolma & 190.5 & 3 & 0.61719 & 0.59931 & 0.02681 & 9.46\% \\
DataDecide Dolma & 371.5 & 3 & 0.56958 & 0.55390 & 0.02352 & 9.97\% \\
DataDecide Dolma & 597.7 & 3 & 0.54308 & 0.52778 & 0.02295 & 10.96\% \\
DataDecide Dolma & 1279.9 & 3 & 0.50481 & 0.49158 & 0.01984 & 11.60\% \\
Delphi & 447.2 & 3 & 0.66510 & 0.64449 & 0.03091 & 9.33\% \\
Delphi & 550.3 & 3 & 0.62298 & 0.60436 & 0.02793 & 9.66\% \\
Delphi & 837.0 & 3 & 0.59963 & 0.58203 & 0.02640 & 9.93\% \\
Delphi & 998.0 & 3 & 0.58321 & 0.56647 & 0.02511 & 10.07\% \\
Delphi & 1180.0 & 3 & 0.56536 & 0.54939 & 0.02396 & 10.35\% \\
Delphi & 1384.6 & 3 & 0.55064 & 0.53537 & 0.02291 & 10.56\% \\
Delphi & 1934.7 & 3 & 0.53123 & 0.51686 & 0.02155 & 10.91\% \\
Delphi & 2544.6 & 3 & 0.51976 & 0.50599 & 0.02066 & 11.10\% \\
Delphi & 2895.2 & 3 & 0.50448 & 0.49138 & 0.01964 & 11.50\% \\
Delphi & 3383.1 & 3 & 0.49200 & 0.47933 & 0.01900 & 12.01\% \\
Delphi & 9714.7 & 3 & 0.44627 & 0.43545 & 0.01622 & 14.41\% \\
\bottomrule
\end{tabular}
\end{table}

\subsection{Evaluation data and units}
\label{sec:evaluation-corpora}
Figure~\ref{fig:evaluation-corpora} compares spread on recent web text and C4 validation. Appendix~\ref{sec:c4-validation} gives the C4 counterparts to the main figures. Model labels such as DataDecide-C4 name the training corpus; both model ladders are evaluated on both text samples.

The recent sample comes from August 2026 Common Crawl, collection \texttt{CC-MAIN-2026-34}. We require a June--August date expression and apply C4-like cleaning to retain long English prose. To limit publisher differences, we keep one document per domain group represented in C4 validation or sampled training shards. A fixed shuffle selects the evaluation documents.
These choices aim to reduce likely training overlap and differences from C4. They do not guarantee publication dates or exclude older quotations and boilerplate. Corpus differences therefore do not isolate memorization.

Recent source segments contain the first 1,025 GPT-NeoX tokens of each selected document; the C4 sample also contains 96 segments. Each public model retokenizes the decoded text and processes up to 1,024 tokens, using the first 64 as context before scoring. Losses and geometric spreads are divided by scored bytes. Compatible ensemble members predict aligned token identities; comparisons across tokenizers can score different prefixes of the same source segment. Difficulty analyses remain in nats/token within a shared token space.

\subsection{Broader public-model comparisons}
Table~\ref{tab:ensemble-families} specifies the public-model ensembles in Figure~\ref{fig:champion}b. Circles denote seed comparisons, squares release reruns and small recipe changes, and triangles larger training or architecture changes. The latter groups include variation beyond a random seed. Their population rescaling is descriptive; the actual finite spread is obtained by multiplying by $(S-1)/S$.

\begin{table}[tbp]
\centering
\caption{\textbf{Ensemble inventory for Figure~\ref{fig:champion}.} Sizes count total parameters, including embeddings; unequal-member pools use the geometric mean of member counts, not their sum. For mixture-of-experts models these are total, not active, parameter counts. Every Pythia seed suite uses its full retained pool, including six members at 14M and nine at 31M. The three groups contain 28, 27, and 17 public-model comparisons, respectively.}
\label{tab:ensemble-families}
\small
\begin{tabular}{@{}p{0.31\linewidth}p{0.42\linewidth}r@{}}
\toprule
Model family & Ensemble construction & Plotted size $N$ \\
\midrule
\multicolumn{3}{@{}l}{\textit{Seed comparisons}} \\
Pythia, standard Pile & 6--10 retained seeds per size & 14.1--405M \\
Pythia, deduplicated v0 & Separate public seed-labelled pair & 6.86B \\
DataDecide-C4 & Three seeds per size & 6.96M--1.28B \\
DataDecide-Dolma & Three seeds per size & 191M--1.28B \\
Marin Delphi & Three seeds per size & 447M--9.7B \\
\midrule
\multicolumn{3}{@{}l}{\textit{Release reruns, recipe variants, and near-replicates}} \\
Pythia release reruns & v0/v1 pairs, standard or deduplicated Pile & 70.4M--11.85B \\
Pythia mixed releases & Three members at one size & 6.86B \\
DataDecide recipes & Five training recipes per size & 191M--1.28B \\
Hubble & 2--12 near-replicates per set & 1.18--8.27B \\
OLMo-7B / Twin & Pair at a matched 2T-token budget & 6.89B \\
\midrule
\multicolumn{3}{@{}l}{\textit{Larger training or architecture changes}} \\
Pythia Pile / deduplicated & Pairs changing training data & 70.4M--11.85B \\
Qwen2.5 / Qwen3, dense & Five cross-generation pairs & 543M--14.77B \\
Qwen2.5 / Qwen3, dense/MoE & One cross-generation pair & 31.63B \\
Gemma-4 & Dense/MoE pair & 27.83B \\
Nemotron-3 / 3.5 & Pair changing training procedure & 31.58B \\
DeepSeek-V3 / V3.1 & Pair with shared early training & 671.03B \\
\bottomrule
\end{tabular}
\end{table}

When members differ in size, the plotted count is their geometric mean. Sparse models use total rather than active parameters. Small horizontal offsets separate overlapping symbols; fits use unshifted counts. A nearly identical-weight Pythia standard/deduplicated pair is excluded; comparisons across release versions remain outside the seed-only fits.

\subsection{Spread fits and their uncertainty}
\label{sec:spread-fits}
We regress log corrected spread on log parameter count using the family conventions in Table~\ref{tab:fit-conventions}. Table~\ref{tab:size-fits} gives the primary seed laws. DataDecide fits omit its smallest sizes, where the slope changes. Figure~\ref{fig:champion}b fits the primary DataDecide C4 and Dolma seed comparisons separately. The Qwen fit uses the five dense cross-generation pairs. Table~\ref{tab:ensemble-measure-scaling} gives the corresponding fitted exponents and the arithmetic-spread and BALD results. Figure~\ref{fig:evaluation-corpora} compares the fitted trends on both evaluation corpora.

\begin{table}[tbp]
\centering\small
\caption{\textbf{Family conventions for size fits.} Setting weights apply to spread laws and are divided equally among the observed checkpoints within each setting for individual-model loss fits. Equal weights are constant across settings; interval weights describe coverage in log size. Component-law fits use one mean member loss per ensemble setting and the narrower cohorts in Appendix~\ref{sec:component-share}. Parameter counts are total parameters.}
\label{tab:fit-conventions}
\setlength{\tabcolsep}{4pt}
\begin{tabular}{@{}p{.19\linewidth}p{.23\linewidth}p{.52\linewidth}@{}}
\toprule
Family & Setting weights $v_i$ & Individual-model loss observations \\
\midrule
Pythia & Equal across sizes & One released checkpoint per size (10 checkpoints) \\
DataDecide C4 & Equal across sizes & Default checkpoint at each size (7 checkpoints) \\
DataDecide Dolma & Equal across sizes & Default checkpoint at each size (4 checkpoints) \\
Dense Qwen2.5/3 & Equal across sizes & Each generation fitted separately (12 checkpoints in total) \\
Delphi & Log-size interval coverage & All 33 released checkpoints; three seeds at each size \\
\bottomrule
\end{tabular}
\end{table}

For interval coverage, order the distinct sizes as $x_i=\log N_i$, $i=1,\ldots,m$. Set $b_0=x_1$, $b_m=x_m$, and $b_i=(x_i+x_{i+1})/2$ for $1\leq i<m$, and define $v_i=(b_i-b_{i-1})/(x_m-x_1)$. Thus the size weights sum to one. Dividing a setting's weight among its observed members prevents a larger retained seed population from receiving extra loss-fit weight. Within a family, multiplying every weight by the same constant leaves the fitted coefficients unchanged.

\paragraph{Individual-model loss fits.}
Figure~\ref{fig:champion}a uses each checkpoint's own loss and total parameter count. The six fitted public-model ladders contain the 66 checkpoints listed by family in Table~\ref{tab:fit-conventions}. Eleven other checkpoints appear as measurements only. At fixed $E$, we regress $\log(L-E)$ on $\log(N/10^9)$ using those weights, then restore $E$. The central floor and range, $0.31727$ and $[0.26019,0.33373]$ nats/byte, come from the separate 28-setting component-law profile in Appendix~\ref{sec:component-share}. Each colored band envelopes the loss laws refitted across that range; dashed curves extrapolate beyond the family's largest measured size to $1$T. These bands propagate floor sensitivity rather than provide checkpoint confidence intervals. No population rescaling is applied to individual checkpoint losses.

\begin{table}[tbp]
\centering
\caption{Independent true-seed size fits. Table~\ref{tab:ensemble-measure-scaling} gives public-model reference laws and coefficients for all three pooling measures. Size ranges count total parameters. The fitted form is $\widehat\rho_\infty^{\mathrm G}=c(N/10^9)^{-\gamma}$. Fits use the family weights in Appendix~\ref{sec:spread-fits}. Standard errors and log-RMSE describe weighted regression residuals conditional on the selected models and text; they are not complete sampling intervals.}
\label{tab:size-fits}
\small
\begin{tabular}{lrrrrr}
\toprule
Family & Sizes & $N$ (M) & $\gamma$ & SE & Log-RMSE \\
\midrule
Pythia & 5 & 14.1--405 & 0.138 & 0.020 & 0.041 \\
DataDecide C4 & 5 & 76.4--1280 & 0.162 & 0.016 & 0.027 \\
DataDecide Dolma & 4 & 190.5--1280 & 0.152 & 0.018 & 0.017 \\
Delphi & 11 & 447.2--9715 & 0.200 & 0.008 & 0.022 \\
\bottomrule
\end{tabular}
\end{table}

\begin{preprintfigure}[tbp]
\centering\includegraphics[width=\linewidth]{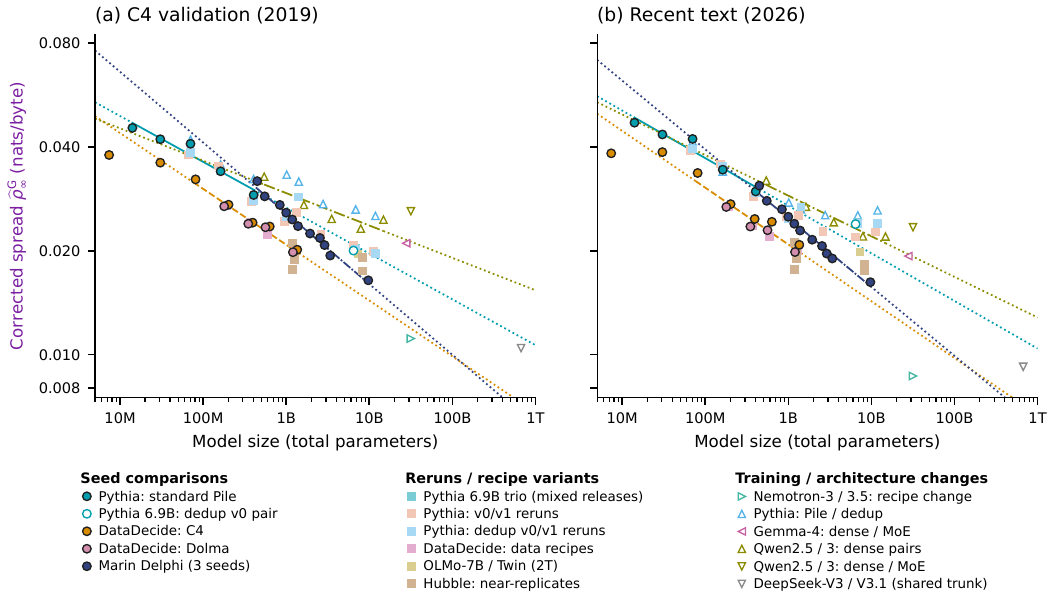}
\caption{\textbf{Spread scaling on C4 validation and recent text.} The 72 public comparisons from Figure~\ref{fig:champion}b are evaluated on each corpus. Marker classes and size offsets follow the main comparison. Curves are refitted on each corpus; dotted segments extrapolate. The DataDecide line combines seed and recipe comparisons. Pythia filtering is applied separately to each corpus, and every size uses its full retained pool. Fits use the conventions in Table~\ref{tab:fit-conventions} for both corpora. No irreducible-loss estimate enters these measurements.}
\label{fig:evaluation-corpora}
\end{preprintfigure}

\paragraph{Specialized evaluation domains.}
We evaluate 32 fixed passages per domain: molecular SMILES strings from \href{https://deepchemdata.s3-us-west-1.amazonaws.com/datasets/Lipophilicity.csv}{MoleculeNet/ChEMBL Lipophilicity} \citep{wu2018moleculenet}, January 2013 \href{https://database.lichess.org/}{Lichess} games in PGN notation \citep{lichess2013database}, \href{https://ifdo.ca/~seymour/nottingham/nottingham.html}{Nottingham} folk tunes in ABC notation \citep{foxley2011nottingham}, and Lean formal-mathematics source from \href{https://github.com/leanprover-community/mathlib4/tree/6201aea845cabc150786187e18e0ff60c9094ab8}{mathlib4} \citep{mathlib2020}.
The original 96-passage C4 sample provides the reference in Figure~\ref{fig:ood-seed-spread}; these specialized sources are domain shifts, without assuming they were absent from pretraining.

\begin{preprintfigure}[htbp]
\centering\includegraphics[width=\linewidth]{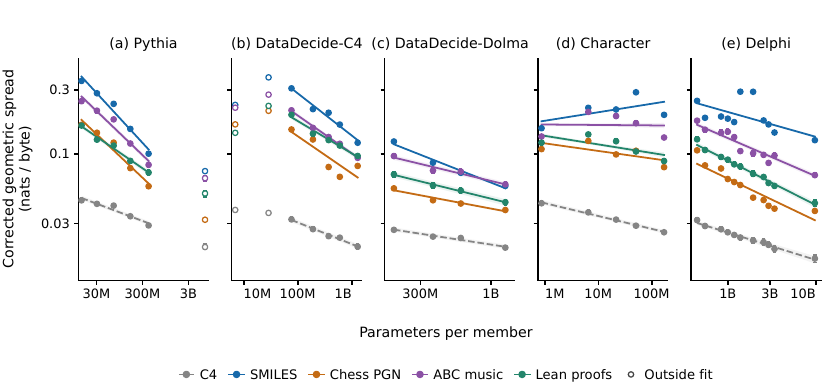}
\caption{\textbf{Seed-ensemble spread across evaluation domains.} (a) Pythia, (b,c) DataDecide trained on C4 and Dolma, (d) character models at 100 training bytes per analytic parameter, and (e) all eleven Delphi sizes. Points use each size's full retained seed pool and the $S/(S-1)$ correction, with membership fixed across corpora, including C4 (gray). Separate log--log fits use equal size weights except for Delphi's log-size interval weights, and stop within their observed fitting ranges. Open markers are excluded from fits: the separate 6.9B Pythia pair and the two smallest DataDecide-C4 sizes. Bars and bands are $95\%$ intervals from 3,000 passage-bootstrap resamples, conditional on checkpoints.}
\label{fig:ood-seed-spread}
\end{preprintfigure}

\subsection{Sensitivity to member and size selection}
\label{sec:selection-sensitivity}
Table~\ref{tab:selection-sensitivity} repeats the fit with and without filtering and changes the included size range. This check averages corrected pair spreads at every Pythia size, whereas the primary analysis uses full retained pools at every size. Its coefficients can therefore differ from Table~\ref{tab:size-fits}. The shallow decline survives, but filtering materially changes the exponent. 
\begin{table}[tbp]
\centering
\caption{Sensitivity of geometric-spread fits to member and size selection, using byte-normalized spreads. DD denotes DataDecide; Recent denotes the August 2026 text sample. Filtered Pythia fits exclude runs whose loss exceeds the suite median by more than $0.01$ nats/byte. The reported decay exponent $\gamma$ is positive. Deletion ranges omit one size at a time; they are sensitivity checks, not confidence intervals.}
\label{tab:selection-sensitivity}
\small
\begin{tabular}{llrrrr}
\toprule
Corpus & Selection & Sizes & $\gamma$ & Log-RMSE & Deletion range \\
\midrule
C4 & Pythia, available & 5 & 0.154 & 0.035 & $[0.138,0.180]$ \\
C4 & Pythia, filtered & 5 & 0.132 & 0.041 & $[0.109,0.152]$ \\
C4 & DD C4, all sizes & 7 & 0.127 & 0.048 & $[0.118,0.154]$ \\
C4 & DD C4, $N\geq60$M & 5 & 0.162 & 0.018 & $[0.152,0.167]$ \\
C4 & DD Dolma & 4 & 0.155 & 0.020 & $[0.126,0.160]$ \\
Recent & Pythia, available & 5 & 0.169 & 0.027 & $[0.161,0.185]$ \\
Recent & Pythia, filtered & 5 & 0.138 & 0.042 & $[0.118,0.158]$ \\
Recent & DD C4, all sizes & 7 & 0.128 & 0.062 & $[0.118,0.163]$ \\
Recent & DD C4, $N\geq60$M & 5 & 0.162 & 0.027 & $[0.137,0.168]$ \\
Recent & DD Dolma & 4 & 0.152 & 0.017 & $[0.140,0.156]$ \\
\bottomrule
\end{tabular}
\end{table}

\subsection{Our character-level models}
\label{sec:byte-details}
\subsubsection{Training design and parameter counts}
\label{sec:byte-training}
Our character-level models are decoder-only transformers trained on FineWeb-Edu, with one byte per token. They use RMSNorm, SwiGLU, rotary position embeddings, query/key normalization, 64-dimensional attention heads, no biases, and untied input and output embeddings. Training processes additional corpus positions as the budget grows.

Let $d$ denote hidden width, $h=d/32$ the number of layers, and $D$ the training bytes. Character scaling and branching use the analytic count $N=12hd^2$, which approximates the block matrices and excludes normalization and vocabulary weights. Table~\ref{tab:byte-training-settings} gives the size and training grid.

Training begins with a linear warmup, followed by a constant learning rate. At each measured budget, a copy of the model and optimizer state completes its own linear cooldown. The constant-rate trajectory then resumes from the saved state. This gives every budget a completed learning-rate schedule, making the data comparison easier to interpret than intermediate checkpoints from one long decay.

\begin{table}[tbp]
\centering\small
\caption{\textbf{Character-model hyperparameters by model size.} Selected batch sizes and peak learning rates for each study. Batches count sequences of 512 bytes. The three rate columns apply to hidden matrices (Muon), input embeddings (AdamW), and the output head and normalization parameters (AdamW). Weight decay applies to hidden matrices and the head.}
\label{tab:byte-training-settings}
\setlength{\tabcolsep}{4pt}
\begin{tabular}{@{}rrrrrrr@{}}
\toprule
Width $d$ & Layers $h$ & $N$ (M) & Batch & Muon rate & Embedding rate & Head/norm rate \\
\midrule
\multicolumn{7}{@{}l}{\textit{Spread and full-parameter branching experiments}} \\
128 & 4 & 0.786 & 16 & 0.006087 & 0.066236 & 0.001775 \\
192 & 6 & 2.654 & 32 & 0.006087 & 0.066236 & 0.001775 \\
256 & 8 & 6.291 & 64 & 0.006087 & 0.066236 & 0.001775 \\
384 & 12 & 21.234 & 128 & 0.006087 & 0.066236 & 0.001775 \\
512 & 16 & 50.332 & 256 & 0.006087 & 0.066236 & 0.001775 \\
768 & 24 & 169.869 & 512 & 0.006087 & 0.066236 & 0.001775 \\
\midrule
\multicolumn{7}{@{}l}{\textit{Loss-law comparison}} \\
128 & 4 & 0.786 & 1 & 0.001522 & 0.016559 & 0.000444 \\
192 & 6 & 2.654 & 2 & 0.000761 & 0.008280 & 0.000222 \\
256 & 8 & 6.291 & 8 & 0.001522 & 0.016559 & 0.000444 \\
384 & 12 & 21.234 & 32 & 0.003044 & 0.033118 & 0.000888 \\
512 & 16 & 50.332 & 256 & 0.006087 & 0.066236 & 0.001775 \\
768 & 24 & 169.869 & 512 & 0.006087 & 0.066236 & 0.001775 \\
\midrule
\multicolumn{3}{@{}l}{Vocabulary / context length} & \multicolumn{4}{l@{}}{256 bytes / 512 bytes} \\
\multicolumn{3}{@{}l}{Warmup / cooldown share} & \multicolumn{4}{l@{}}{18 steps / final $20\%$} \\
\multicolumn{3}{@{}l}{Scaling budgets (bytes/$N$)} & \multicolumn{4}{l@{}}{6.25--400, in factors of $\sqrt{2}$} \\
\multicolumn{3}{@{}l}{Muon momentum} & \multicolumn{4}{l@{}}{0.95} \\
\multicolumn{3}{@{}l}{AdamW $(\beta_1,\beta_2)$ / $\epsilon$} & \multicolumn{4}{l@{}}{$(0.9,0.95)$ / $10^{-10}$} \\
\multicolumn{3}{@{}l}{Weight decay} & \multicolumn{4}{l@{}}{0.001423} \\
\bottomrule
\end{tabular}
\end{table}

Table~\ref{tab:byte-training-settings} lists the chosen batch sizes and learning rates at each model size. These settings are held fixed across training budgets within each study. Appendix~\ref{sec:branching-details} gives the branching and adapter procedures.

\subsubsection{Spread and reducible-loss fits}
\label{sec:byte-scaling}
To compare how quickly spread and reducible loss decrease with size, we fit spread, the invisible portion, and total reducible loss on the same model sizes, training budgets, and held-out FineWeb-Edu bytes used in Figure~\ref{fig:character-spread-invisible}. The spread study retains seeds $0$ and $2$ at width $128$ and three seeds at widths $256$, $384$, $512$, and $768$. We exclude width $192$ and seed $1$ at width $128$ because of anomalous training trajectories. These fits use the analytic count $N=12hd^2$. Refitting spread against total parameters gives $\gamma=0.11844\pm0.00156$; converting bits to nats leaves this exponent unchanged.

For the invisible portion in Figure~\ref{fig:character-spread-invisible}, we first undo the population rescaling to recover the actual finite-pool loss:
\[
 L(q_S^{\mathrm G})=\bar L_S-\frac{S-1}{S}\widehat\rho_\infty^{\mathrm G}.
\]
We then subtract the assumed irreducible loss $E$. Table~\ref{tab:byte-scaling-summary} compares the size exponents at the two choices of $E$: spread decreases more slowly than either the invisible portion or total reducible loss in both cases. Spread itself does not depend on $E$. These fits follow a window of bytes per parameter, so increasing size also increases training data. Successive budgets share training history; regression errors are not independent-run confidence intervals.

\begin{preprintfigure}[tbp]
\centering
\begin{minipage}[c]{.44\linewidth}
\includegraphics[width=\linewidth]{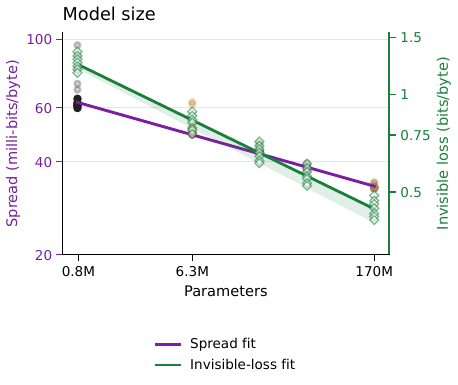}
\end{minipage}\hfill
\begin{minipage}[c]{.53\linewidth}
\captionsetup{font=footnotesize,justification=raggedright,singlelinecheck=false,skip=0pt}
\caption{\textbf{Character-model spread decreases more slowly than invisible loss.} Corrected geometric spread (left axis, milli-bits/byte) and finite-pool invisible loss $L(q_S^{\mathrm G})-E$ (right axis, bits/byte) on 524,288 held-out FineWeb-Edu bytes. Fits use the same 35 settings across five sizes at 50--400 bytes per analytic parameter; fainter spread points show the remaining training budgets. Green shading varies the assumed irreducible loss $E$ from $0.593341$ to $0.63764990$ bits/byte and shows floor sensitivity.}
\label{fig:character-spread-invisible}
\end{minipage}
\end{preprintfigure}

\begin{table}[tbp]
\centering\small
\caption{\textbf{Spread decreases more slowly than reducible loss.} Fits use the same 35 settings as Figure~\ref{fig:character-spread-invisible}: five model sizes and 50--400 bytes per analytic parameter, evaluated on 524,288 held-out bytes. For $Q(N)\propto N^{-\eta}$, a larger exponent means a faster decrease. Entries give the exponent $\pm$ its regression standard error. The two columns vary the assumed irreducible loss $E$ in bits/byte. Spread does not depend on $E$, so its two entries agree.}
\label{tab:byte-scaling-summary}
\begin{tabular}{@{}lcc@{}}
\toprule
Quantity $Q$ & $E=0.593341$ & $E=0.637650$ \\
\midrule
Geometric spread $\widehat\rho_\infty^{\mathrm G}$ & $0.11654\pm0.00163$ & $0.11654\pm0.00163$ \\
Invisible portion $L(q_S^{\mathrm G})-E$ & $0.19140\pm0.00512$ & $0.20396\pm0.00547$ \\
Total reducible loss $\bar L_S-E$ & $0.18709\pm0.00488$ & $0.19887\pm0.00521$ \\
\bottomrule
\end{tabular}
\end{table}

\subsubsection{Empirical constraints on the irreducible loss}
\label{sec:byte-floor}
The working irreducible loss comes from extrapolating publicly released models' losses under the character study's context settings. Kimi-K2-Base \citep{kimik2base} supplies the best achieved reference loss under those settings (Table~\ref{tab:byte-floor-settings}). We vary the assumed irreducible loss between the first two values; DeepSeek-V4-Pro-Base, which has a higher loss under these settings, is shown for comparison. 

\begin{table}[tbp]
\centering\small
\caption{\textbf{Irreducible-loss sensitivity for the character study.} All values use the 512-byte context settings. The achieved upper endpoint is rounded downward for refitting.}
\label{tab:byte-floor-settings}
\begin{tabular}{@{}lr@{}}
\toprule
Choice & $E$ (bits/byte) \\
\midrule
Working value from joint extrapolation & 0.593341 \\
Best achieved reference: Kimi-K2-Base & 0.63764990 \\
DeepSeek-V4-Pro-Base & 0.68195265 \\
\bottomrule
\end{tabular}
\end{table}

Note that spread itself does not depend on irreducible loss; this value mainly appears in calculating the visible fraction. Appendix~\ref{sec:scaling-comparison} uses the same irreducible-loss choices for each loss-law hypothesis.

\section{Pooling, entropy, and fixed difficulty groups}
\label{sec:pooling-full-windows}

\subsection{Comparing ensemble measures}
We compare the measures defined in Section~\ref{sec:ensembles} on the same members and recent-text positions. Figure~\ref{fig:pooling} shows Pythia, DataDecide-C4, and Delphi seed ensembles. The shallow size trend persists for arithmetic spread and BALD.

\begin{preprintfigure}[tbp]
\centering\includegraphics[width=\linewidth]{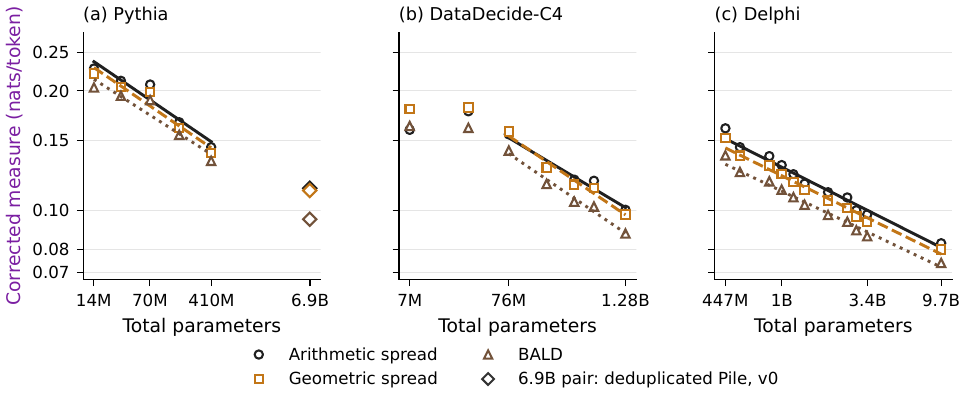}
\caption{\textbf{The shallow size trend persists under both loss pools and BALD.} Corrected measures use matching members and recent-text positions within each family. Points are measurements; lines fit the five standard-Pile Pythia suites, DataDecide-C4 sizes above 60M parameters, and eleven Delphi triples, using Table~\ref{tab:fit-conventions}. Open diamonds show the separate 6.9B deduplicated-v0 pair, excluded from the fits. Lines stop at their fitting ranges.}
\label{fig:pooling}
\end{preprintfigure}

\subsection{Scaling under all three ensemble measures}
\label{sec:measure-family-scaling}
Table~\ref{tab:ensemble-measure-scaling} gives public-model reference laws and fits on matching member populations. Every measure is normalized by scored bytes and multiplied by $S/(S-1)$. We fit $d(N)=c(N/10^9)^{-\gamma}$ by least squares in log space, using the family weights in Appendix~\ref{sec:spread-fits}. Standard errors describe regression residuals conditional on the selected models and text.

Every fit uses the same 96 recent-text windows for all three measures. Geometric and arithmetic spread use all 96 windows for each of the 72 comparisons. Full-window BALD is available for 65. Seven comparisons are left out of matched-measure fits because BALD needs full-vocabulary predictions, which are large to store and were saved for only two windows of these comparisons: the standard-Pile Pythia reruns at 70M, 160M, 410M, and 6.9B, and the Nemotron, Gemma, and DeepSeek pairs.

\begin{table}[tbp]
\centering
\footnotesize
\setlength{\tabcolsep}{4pt}
\renewcommand{\arraystretch}{1.15}
\caption{\textbf{Scaling coefficients for public-model reference laws and primary seed families.} The law is $d(N)=c(N/10^9)^{-\gamma}$. Each cell gives $c$ (nats/byte) on the first line and $\gamma$ on the second, with conditional regression standard errors where estimable; $n$ counts eligible comparisons. The first block includes a pooled DataDecide reference. The second block uses matching 96-window cohorts and full-pool estimators across the three measures. Figure~\ref{fig:champion}b uses the separate DataDecide seed fits. Fits use the family weights in Appendix~\ref{sec:spread-fits}. The dense Qwen reference is a descriptive cross-generation comparison, not an independent-seed estimate. Details and exclusions are given in Section~\ref{sec:measure-family-scaling}.}
\label{tab:ensemble-measure-scaling}
\begin{tabular}{@{}lrlll@{}}
\toprule
Model family / reference & $n$ & Geometric & Arithmetic & BALD \\
\midrule
\multicolumn{5}{l}{\textit{Public-model reference laws}}\\*
Pythia reference & 5 & $\begin{gathered}0.02703\pm 0.00157\\0.138\pm 0.020\end{gathered}$ & $\begin{gathered}0.02802\pm 0.00183\\0.138\pm 0.023\end{gathered}$ & $\begin{gathered}0.02628\pm 0.00176\\0.129\pm 0.023\end{gathered}$ \\
\addlinespace[3pt]
DataDecide reference & 13 & $\begin{gathered}0.02087\pm 0.00031\\0.165\pm 0.012\end{gathered}$ & $\begin{gathered}0.02138\pm 0.00036\\0.153\pm 0.014\end{gathered}$ & $\begin{gathered}0.01924\pm 0.00019\\0.162\pm 0.008\end{gathered}$ \\
\addlinespace[3pt]
Dense Qwen reference & 5 & $\begin{gathered}0.02891\pm 0.00073\\0.118\pm 0.015\end{gathered}$ & $\begin{gathered}0.02904\pm 0.00076\\0.120\pm 0.016\end{gathered}$ & $\begin{gathered}0.02435\pm 0.00049\\0.117\pm 0.012\end{gathered}$ \\
\addlinespace[3pt]
Delphi reference & 11 & $\begin{gathered}0.02487\pm 0.00023\\0.200\pm 0.008\end{gathered}$ & $\begin{gathered}0.02613\pm 0.00027\\0.205\pm 0.009\end{gathered}$ & $\begin{gathered}0.02275\pm 0.00020\\0.194\pm 0.007\end{gathered}$ \\
\addlinespace[3pt]
\midrule
\multicolumn{5}{l}{\textit{Matched family laws: seeds}}\\*
Pythia, standard-Pile seeds & 5 & $\begin{gathered}0.02703\pm 0.00157\\0.138\pm 0.020\end{gathered}$ & $\begin{gathered}0.02802\pm 0.00183\\0.138\pm 0.023\end{gathered}$ & $\begin{gathered}0.02628\pm 0.00176\\0.129\pm 0.023\end{gathered}$ \\
\addlinespace[3pt]
DataDecide-C4 seeds & 5 & $\begin{gathered}0.02163\pm 0.00050\\0.162\pm 0.016\end{gathered}$ & $\begin{gathered}0.02244\pm 0.00059\\0.143\pm 0.018\end{gathered}$ & $\begin{gathered}0.01946\pm 0.00036\\0.163\pm 0.013\end{gathered}$ \\
\addlinespace[3pt]
DataDecide-Dolma seeds & 4 & $\begin{gathered}0.02072\pm 0.00037\\0.152\pm 0.018\end{gathered}$ & $\begin{gathered}0.02079\pm 0.00033\\0.147\pm 0.016\end{gathered}$ & $\begin{gathered}0.01899\pm 0.00032\\0.152\pm 0.017\end{gathered}$ \\
\addlinespace[3pt]
Delphi seeds & 11 & $\begin{gathered}0.02487\pm 0.00023\\0.200\pm 0.008\end{gathered}$ & $\begin{gathered}0.02613\pm 0.00027\\0.205\pm 0.009\end{gathered}$ & $\begin{gathered}0.02275\pm 0.00020\\0.194\pm 0.007\end{gathered}$ \\
\addlinespace[3pt]
\bottomrule
\end{tabular}
\end{table}

\FloatBarrier

\subsection{Keeping difficulty groups fixed}
OLMo-7B token entropy assigns each recent-text position to one of six fixed percentile groups. We then compare the same tokens across Pythia sizes. Holding membership fixed avoids making a token easier simply by moving it to another group as models improve. Reference entropy is a difficulty proxy, rather than measured irreducible randomness.
For each group $b$, we fit $\log d_b(N)=\log A_b-\gamma_b\log(N/10^9)$ with equal weight per ensemble. Figure~\ref{fig:difficulty-pooling-fits} shows the easy--hard contrast for all three measures. 

\begin{preprintfigure}[tbp]
\centering\includegraphics[width=\linewidth]{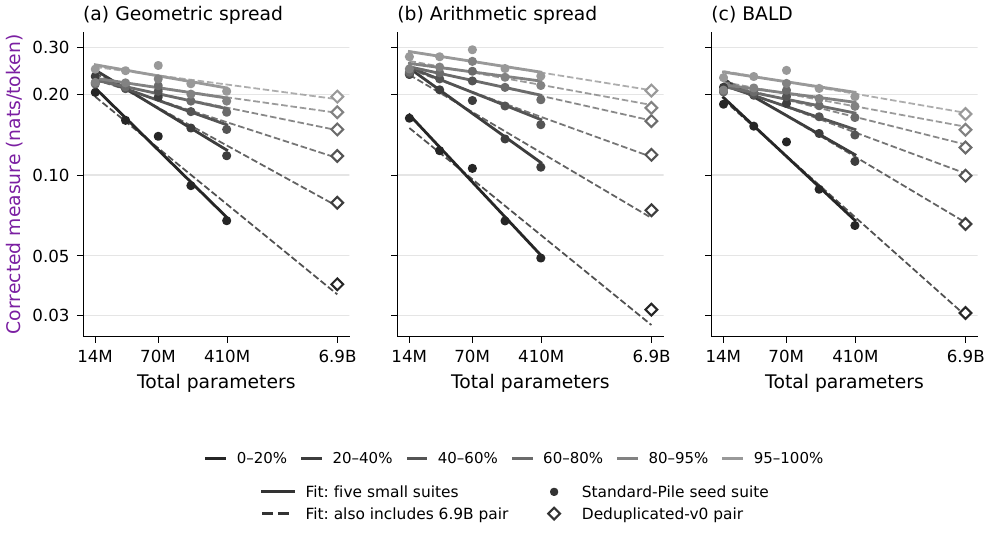}
\caption{\textbf{Disagreement decreases faster on low-entropy tokens under all three measures.} Pythia seed ensembles are evaluated on the same recent-text positions, divided into six fixed percentile groups by OLMo-7B prediction entropy (dark to light: lowest to highest entropy). Panels show group means of (a) geometric spread, (b) arithmetic spread, and (c) BALD, multiplied by $S/(S-1)$ for $S$ ensemble members, in nats/token versus total parameters on log--log axes. Filled circles are the five standard-Pile suites at 14M--410M parameters; open diamonds are the separate 6.9B deduplicated-v0 pair. Solid lines fit power laws to the five suites; dashed lines refit with the 6.9B pair, mixing two training recipes. Both fits show a steeper decline for low-entropy tokens and more persistent disagreement for high-entropy tokens.}
\label{fig:difficulty-pooling-fits}
\end{preprintfigure}

\section{Changing the available context}
\label{sec:context-details}
Here we provide details on the role of context length in ensemble spread.
For this, we vary the preceding context while keeping model weights and target passages fixed.
 
\subsection{Models and evaluation}
\label{sec:context-protocol}
We use the Pythia, DataDecide, and Delphi sizes in Table~\ref{tab:primary-seeds}, excluding the separate 6.9B Pythia pair, and the three smallest dense Qwen2.5/Qwen3 pairs (0.5B/0.6B, 1.5B/1.7B, and 3B/4B; Appendix~\ref{sec:model-families}). Pythia uses three members per size: the default release and seeds 1--2, except at 160M, where the default release fails the loss filter (Appendix~\ref{sec:seed-populations}) and we use seed 3 instead. DataDecide uses \texttt{default}, \texttt{small-aux-2}, and \texttt{small-aux-3}, replacing the auxiliary runs with \texttt{large-aux-2} and \texttt{large-aux-3} at the largest size; Delphi uses seeds 0, 42, and 62746.

Distributions use 50,277 assigned Pythia token IDs, 50,280 DataDecide IDs, 151,665 shared Qwen IDs, and 128,256 assigned Delphi IDs. Qwen uses native context limits without positional extension or chat templates.
We sample 128 passages from 77 eligible PG-19 test books \citep{rae2019compressive}. Each contributes 128 consecutive targets, giving 16,384 per setting. Each family retokenizes the same passages and keeps its targets fixed across context lengths. Target blocks are disjoint, although their contexts can overlap. Pretraining overlap is unknown.

We vary the context limit $C$, the maximum number of preceding tokens available for prediction, in powers of two up to 2,048 tokens for Pythia, DataDecide, and Delphi and 32,768 for Qwen. For efficiency, each forward pass scores a block of $B(C)$ consecutive target tokens, where
\begin{equation}
 B(C)=\min\{128,\max[1,\lfloor C/16\rfloor]\}
 \label{eq:context-block}
\end{equation}
Thus each pass scores between 1 and 128 targets. Within a block, the available context grows by one token per target, from $C-B(C)+1$ preceding tokens for the first target to $C$ for the last. We plot the geometric midpoint $c=\sqrt{(C-B(C)+1)C}$ of this range. For $C\leq16$, $B(C)=1$, so every target has exactly $C$ preceding tokens. We recompute truncated prefixes so discarded context cannot remain in a cache.

\subsection{Measures and fitting}
\label{sec:context-measures}
Both spreads and BALD follow the definitions and $S/(S-1)$ rescaling in Section~\ref{sec:ensembles}. Values are in nats/token, with separate axes for each tokenizer. Error bars show one standard deviation across passage means.
For each measure, we fit
\begin{equation}
 \bar d(c)=a+b\log_2(c/32)
 \label{eq:context-fit}
\end{equation}
by unweighted least squares over context limits $C>16$, with free intercept and slope. Here $b$ is the change per context doubling. Fits start at $C=32$, so slopes include part of the initial contraction.

We estimate 95\% slope intervals by resampling the 77 PG-19 test books used for evaluation, with replacement, and refitting the slopes 2,000 times. Each resample keeps a book's passages together and is reused across context lengths and model settings. These intervals capture variation across evaluation text, with the trained models held fixed.

\subsection{Results}
\label{sec:context-results}
Figure~\ref{fig:context-measures} compares all three disagreement measures across the model families. After the initial contraction, geometric spread changes little with additional context for Pythia, Qwen, DataDecide Dolma, and Delphi; DataDecide C4 spread rises slightly. BALD and arithmetic spread behave similarly.
Across the three Qwen pairs, fitted geometric-spread slopes range from $-1.888$ to $-1.116$ milli-nats/token per context doubling. In the same units, Pythia-160M has slope $-0.304$ (paired 95\% interval $[-0.822,0.210]$), while DataDecide-C4-1.28B has $+3.919$ ($[2.480,5.935]$). DataDecide-Dolma geometric slopes range from $-0.852$ to $-0.573$. 

\begin{preprintfigure}[tbp]
\centering
\includegraphics[width=\linewidth]{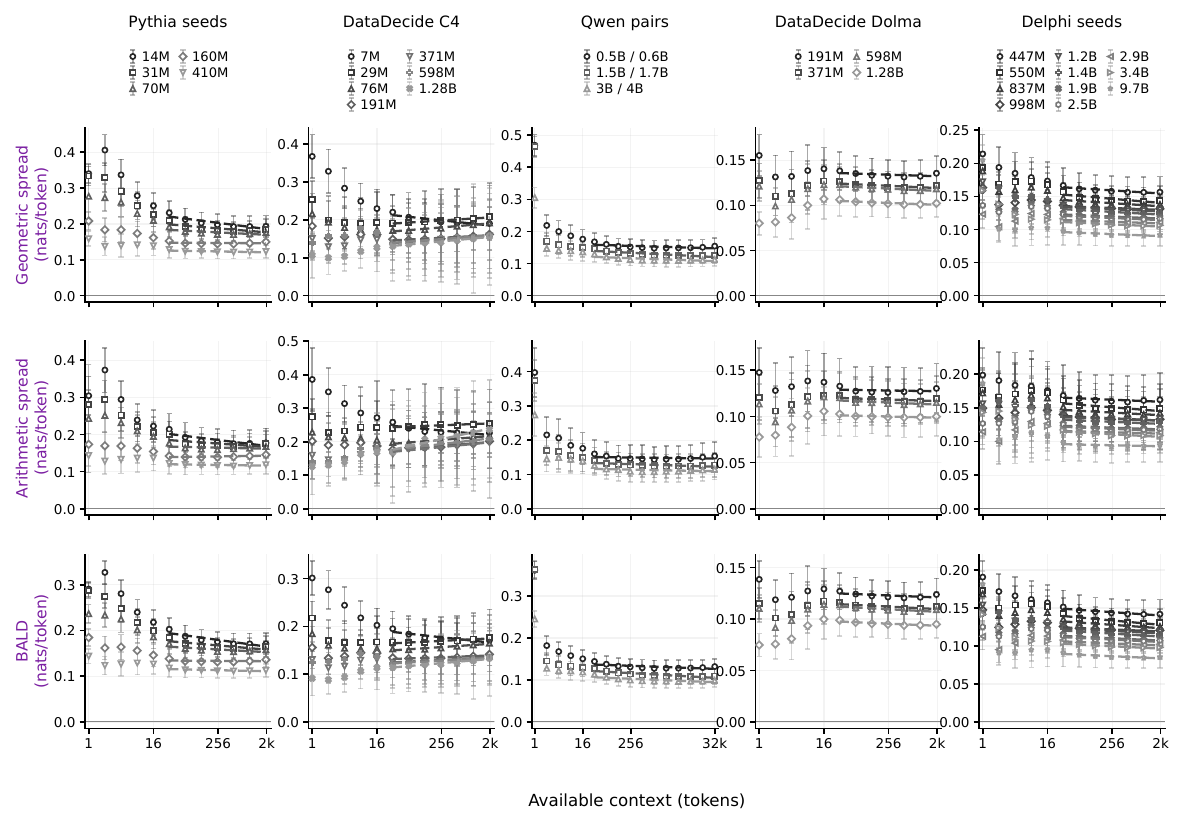}
\caption{\textbf{Ensemble disagreement as context grows.} Rows show geometric spread, arithmetic spread, and BALD; columns show Pythia, DataDecide C4, Qwen, DataDecide Dolma, and Delphi. Points give corrected means over 16,384 fixed PG-19 targets per family, with one passage standard deviation. Dashed lines fit context limits above 16 tokens.}
\label{fig:context-measures}
\label{fig:context-spread}
\label{fig:context-pythia}
\label{fig:context-datadecide}
\end{preprintfigure}

\section{Visible-share scaling and loss-law comparisons}
\label{sec:visible-details}
 
\subsection{Fitting separate visible and invisible laws}
\label{sec:component-share}
We fit the separate-component description in Section~\ref{sec:visible}, specializing Equation~\eqref{eq:component-share} to geometric pooling, $P=\mathrm G$. Its visible component is $\cvis{\rho_\infty^{\mathrm G}(N)=c_{\mathrm G}N^{-\gamma_{\mathrm G}}}$, and its invisible component is $\cinv{L(q_\infty^{\mathrm G})-E=AN^{-\alpha}}$. 
We estimate the shared floor in Figure~\ref{fig:visible} from 28 recent-text settings across five families. These comprise three standard-Pile Pythia suites above 60M, five DataDecide-C4 sets, four DataDecide-Dolma sets, five dense Qwen2.5/3 pairs, and eleven Delphi triples. The separate 6.9B Pythia pair is shown but excluded from these fits. We count total parameters and use their geometric mean for unequal members. These fitting populations are narrower than those in Figure~\ref{fig:champion}b; in particular, the Pythia fit here excludes the 14M and 31M suites, so its spread exponent differs from the headline fit.

For a measured ensemble of size $S$, we use the correction from Section~\ref{sec:ensembles}:
\begin{equation}
 \cvis{\widehat\rho_\infty^{\mathrm G}}=\frac{S}{S-1}\rho_S^{\mathrm G},\qquad
 \widehat f_\infty^{\mathrm G}(E)=\frac{\cvis{\widehat\rho_\infty^{\mathrm G}}}{\ctot{\bar L_S-E}}.
 \label{eq:component-observations}
\end{equation}
The corresponding estimate of $\cinv{L(q_\infty^{\mathrm G})-E}$ is the corrected residual $\cinv{\bar L_S-E-\widehat\rho_\infty^{\mathrm G}}$. It differs from the measured finite ensemble's invisible loss, $L(q_S^{\mathrm G})-E$. For cross-generation Qwen pairs, the correction remains a descriptive rescaling.

Within each public family, we use the normalized setting weights $w_i$ from Table~\ref{tab:fit-conventions} to fit $\rho_\infty^{\mathrm G}(N)$ to $\widehat\rho_\infty^{\mathrm G}$ by least squares in log--log space. We then hold these visible laws fixed and fit a common irreducible loss $E$ and family-specific invisible laws. Let $g(i)$ identify setting $i$'s family and let $\bar L_i$ denote its measured mean member loss, $\bar L_S$. The loss fit minimizes
\begin{equation}
 \begin{aligned}
 \closs{\widehat L_i(E,\theta)}
 &=\cale{E}+\cinv{A_{g(i)}N_i^{-\alpha_{g(i)}}}
   +\cvis{c_{\mathrm G,g(i)}N_i^{-\gamma_{\mathrm G,g(i)}}},\\
 R(E,\theta)&=\sum_{i=1}^{28}w_i\left[\log\closs{\widehat L_i(E,\theta)}-\log\closs{\bar L_i}\right]^2.
 \end{aligned}
 \label{eq:component-floor-fit}
\end{equation}
For a family with $n_g$ retained settings, we normalize the weights as $w_i=n_gv_i/\sum_{j:g(j)=g(i)}v_j$, so they sum to that family's setting count. Every ensemble contributes one mean member loss. The ten coefficients in $\theta$ give an invisible amplitude and exponent for each family.

To assess sensitivity to the unknown floor, we repeat the loss fit at candidate values of $E$. At each candidate, we keep the visible spread laws fixed and refit the invisible amplitude $A$ and exponent $\alpha$ separately for each of the five public families. These ten coefficients can adjust to the amount of loss left after subtracting $E$. The resulting minimum squared log-loss error is $R(E)$, and $R_{\min}=\min_E R(E)$ is the error of the best fit over all candidate floors. This procedure is called profiling $E$.

We retain floors whose refitted errors are sufficiently close to $R_{\min}$ according to a nominal 95\% residual F-profile. The fit has eleven free parameters, one shared floor and ten invisible-law coefficients, leaving $28-11=17$ residual degrees of freedom. The criterion is
\begin{equation}
 R(E)\leq R_{\min}\left[1+\frac{F_{1,17}(0.95)}{17}\right],\qquad
 R(E)=\min_\theta R(E,\theta).
 \label{eq:component-floor-profile}
\end{equation}
Here $F_{1,17}(0.95)$ is the 95th percentile of the F distribution with one and 17 degrees of freedom. Fixing $E$ constrains one parameter. We compare the increase $R(E)-R_{\min}$ with the estimated residual variance $R_{\min}/17$, retaining $R(E)\lesssim1.26R_{\min}$. The nominal calibration assumes independent, approximately Gaussian weighted residuals $\sqrt{w_i}(\log\widehat L_i-\log\bar L_i)$ with common variance, conditional on the fitted spread laws. Because the setting weights include coverage weights rather than estimated inverse error variances, we interpret the profile as a sensitivity range.

The best-fitting floor is $E=0.31727$ nats/byte, and the uncapped profile range is $[0.26019,0.35190]$. We then impose the empirical upper cap from Appendix~\ref{sec:floor-sensitivity}: the irreducible loss should not exceed the lowest loss already achieved on the evaluation distribution. This truncates the retained range to $[0.26019,0.33373]$ nats/byte.

Figure~\ref{fig:visible} propagates this retained floor range into the component fits and visible fractions. At each candidate floor, all five public families use the same $E$ and refit their invisible laws. In panel (b), each shaded band spans the minimum and maximum fitted $f_\infty^{\mathrm G}(N)$ from Equation~\eqref{eq:component-share} at each size over these refits. For each measured marker, the loss $\bar L_S$ and corrected spread $\widehat\rho_\infty^{\mathrm G}$ stay fixed; its interval varies only $E$ in $\widehat f_\infty^{\mathrm G}(E)=\widehat\rho_\infty^{\mathrm G}/(\bar L_S-E)$. A larger floor reduces the denominator and therefore increases this estimated fraction. These bands and intervals describe floor sensitivity conditional on the fitted spread laws and observed ensembles. They exclude spread-fit uncertainty, seed and text sampling, and systematic errors, so they are not full 95\% uncertainty intervals for the visible fractions.

For the character series in Figures~\ref{fig:champion} and~\ref{fig:visible}, we evaluate the same 35 size--budget settings (98 checkpoints) on the 432,878 recent-text target bytes used by Pythia, with up to 512 preceding bytes. Each setting has equal weight and uses total parameter counts. For individual loss fits, its weight is shared equally among its checkpoints. The public settings alone determine the floor profile; the character fit holds that profile fixed. The other character experiments use held-out FineWeb-Edu text.

In Figure~\ref{fig:visible}a, the visible markers are $\widehat\rho_\infty^{\mathrm G}$ and the invisible markers are $\bar L_S-E-\widehat\rho_\infty^{\mathrm G}$. Each component and its fitted curve are normalized by that component's measured value at the smallest fitted size within its family. For the panel (a) band, we refit the invisible curve at each candidate $E$ and normalize it by the measured invisible loss at the smallest fitted size, evaluated at that same $E$. At each size the band spans the minimum and maximum of these normalized curves. Table~\ref{tab:component-share} reports $\alpha$, $\gamma_{\mathrm G}$, and projected $f_\infty^{\mathrm G}$. All six central fits have $\alpha>\gamma_{\mathrm G}$, so the fitted fraction increases with size under the stated floor choices. Appendix~\ref{sec:floor-sensitivity} shows why this trend depends on $E$. Shaded bands stop at each family's smallest and largest measured fitting sizes; only the central curves extend beyond them in panel (b).

\begin{table}[tbp]
\centering
\caption{\textbf{Separate component fits on recent text.} Exponents describe the invisible loss $L(q_\infty^{\mathrm G})-E$ and visible spread $\rho_\infty^{\mathrm G}$ in Equation~\eqref{eq:component-share}. The 28 recent-text settings determine the shared floor $E=0.31727$ nats/byte. Its uncapped profile is $[0.26019,0.35190]$, truncated to $[0.26019,0.33373]$ by the empirical cap. The 35 character settings use the same fixed floor profile. Projected fractions $f_\infty^{\mathrm G}$ use $10^{12}$ total parameters. Brackets give the range over the retained floor profile. These are extrapolations, not observations.}
\label{tab:component-share}
\begin{tabular}{@{}lrrr@{}}
\toprule
Family & Invisible $\alpha$ & Visible $\gamma_{\mathrm G}$ & $f_\infty^{\mathrm G}(10^{12})$ (\%) \\
\midrule
Pythia & $0.28507$ & $0.19988$ & $19.3\;[12.0,22.6]$ \\
DataDecide-C4 & $0.24043$ & $0.16184$ & $17.0\;[10.2,20.2]$ \\
DataDecide-Dolma & $0.25644$ & $0.15189$ & $19.3\;[11.1,23.4]$ \\
Dense Qwen2.5/3 & $0.27564$ & $0.11784$ & $33.4\;[17.4,42.3]$ \\
Delphi & $0.32614$ & $0.19968$ & $20.1\;[10.6,25.2]$ \\
Character & $0.17036$ & $0.08193$ & $13.1\;[10.5,14.1]$ \\
\bottomrule
\end{tabular}
\end{table}

\subsection{Dependence on the unknown irreducible loss}
\label{sec:floor-sensitivity}
Table~\ref{tab:evaluation-reference-losses} gives achieved single-model losses. DeepSeek-V4-Pro-Base achieves the lowest recent-text loss, 0.33372841 nats/byte. This sets the empirical cap for the recent-text floor.
Kimi-K2-Base and DeepSeek-V4-Pro-Base use their official native FP8 checkpoints, without chat templates, on the same source segments and context limits as the other recent-text models. We retain tokenizer default behavior and normalize by decoded scored bytes. Normalizing over all model output logits gives the same loss as normalizing over only the first $|\mathcal V_{\text{tok}}|$ logits (the procedure used for the other models) for both checkpoints.

\begin{table}[tbp]
\centering\small
\caption{Achieved individual-model losses on the two evaluation corpora, in nats per scored byte. Each corpus uses the same 96 source segments across these models, subject to tokenizer-dependent truncation. These are measured cross-entropies, not conditional-entropy estimates; the corpora have different text distributions. Kimi-K2-Base and DeepSeek-V4-Pro-Base were evaluated only on recent text; dashes denote unmeasured C4 losses.}
\label{tab:evaluation-reference-losses}
\begin{tabular}{lrr}
\toprule
Model & C4 validation & Recent text \\
\midrule
Qwen2.5-72B & 0.479487 & 0.395215 \\
DeepSeek-V3-Base & 0.430436 & 0.365561 \\
DeepSeek-V3.1-Base & 0.442147 & 0.373619 \\
GLM-4.5-Air-Base & 0.450985 & 0.384772 \\
GLM-4.5-Base & 0.417401 & 0.352689 \\
Kimi-K2-Base & \textemdash & 0.340229 \\
DeepSeek-V4-Pro-Base & \textemdash & 0.333728 \\
\bottomrule
\end{tabular}
\end{table}

Figure~\ref{fig:floor} varies $E$ for 23 public seed sets above 60M parameters. The public fits use the retained seed families and exclude the separate 6.9B Pythia pair. The fitted recovered share rises with size for every family at all three displayed nonzero floors. At $E=0$ the trend reverses for all four public families.

\begin{preprintfigure}[tbp]
\centering\includegraphics[width=\linewidth]{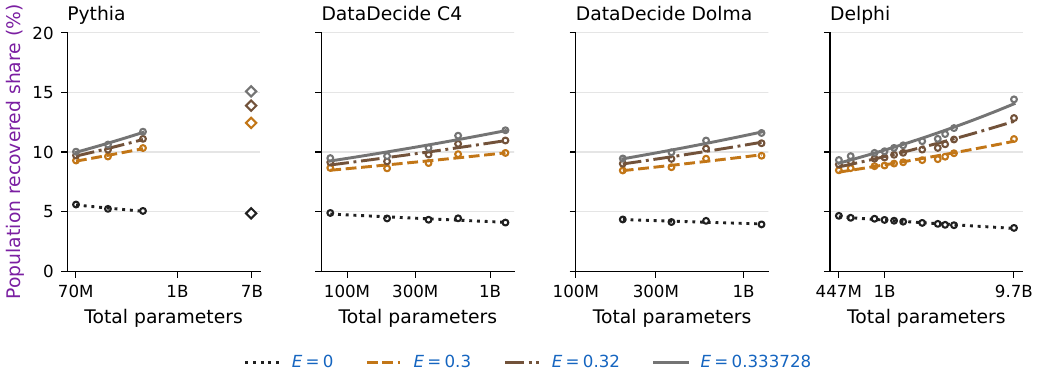}
\caption{\textbf{Recovered fractions depend on the assumed irreducible loss.} Curves fit corrected geometric spread divided by mean member loss minus $E$ on recent text. Public curves use the family weights in Appendix~\ref{sec:spread-fits}; open diamonds show the excluded 6.9B Pythia pair. The four floor choices are sensitivity assumptions.}
\label{fig:floor}
\end{preprintfigure}

Table~\ref{tab:floor-reversal-thresholds} gives the floor at which the fitted log--log slope of $\widehat\rho_\infty^{\mathrm G}/(\bar L_S-E)$ versus $N$ is zero. The character model has a formal negative threshold; its fitted recovered-share slope remains positive for admissible nonnegative floors.

\begin{table}[tbp]
\centering
\caption{\textbf{Reversal thresholds for direct recovered-share fits.}}
\label{tab:floor-reversal-thresholds}
\begin{tabular}{@{}lr@{}}
\toprule
Family & $E_{\mathrm{crit}}$ (nats/byte) \\
\midrule
Character & $-0.35597$ \\
Pythia & 0.18996 \\
DataDecide C4 & 0.20088 \\
DataDecide Dolma & 0.17154 \\
Dense Qwen2.5/Qwen3 pairs & 0.14397 \\
Delphi & 0.20384 \\
Pythia including 6.9B (sensitivity check) & 0.14014 \\
\bottomrule
\end{tabular}

\end{table}

The loss-law comparison below holds an earlier C4 asymptote fixed to compare two descriptions of member loss. For the validation results in Appendix~\ref{sec:c4-validation}, we fit the floor to five families and cap it at the lowest measured C4 loss. The corpus comparison in Figure~\ref{fig:evaluation-corpora} needs no floor assumption. Lower loss on a different corpus alone establishes neither less memorization nor a better entropy estimate.

\subsection{One loss law or a sum of component laws?}
\label{sec:scaling-comparison}
Section~\ref{sec:visible} asks whether total reducible loss or its invisible portion is better described by a power law. We compare both descriptions on the same target: measured mean member loss. 

\paragraph{Fitting and validation.}
Approach A fits the conventional law directly:
\begin{equation}
 \closs{\widehat L_{\mathrm A}(N,D)}=\cale{E}+\ctot{AN^{-a}+BD^{-b}}.
 \label{eq:direct-law-comparison}
\end{equation}
Approach B fits the same form to $R_i=\bar L_i-\widehat\rho_{\infty,i}^{\mathrm G}$, fits spread separately, and adds their predictions:
\begin{equation}
 \begin{aligned}
 \closs{\widehat R(N,D)}&=\cale{E}+\cinv{A'N^{-a'}+B'D^{-b'}},\\
 \cvis{\widehat\rho_\infty^{\mathrm G}(N)}&=cN^{-\gamma},\qquad
 \closs{\widehat L_{\mathrm B}}=\closs{\widehat R}+\cvis{\widehat\rho_\infty^{\mathrm G}}.
 \end{aligned}
 \label{eq:component-law-comparison}
\end{equation}
Final-checkpoint comparisons omit the data term, since it can be absorbed as a constant. The residual $R_i$ estimates a population-pool loss using the approximate correction, rather than an evaluated infinite ensemble. It includes shared data and optimization limitations as well as capacity loss.

Both loss fits minimize a Huber penalty on log residuals, with transition at $0.01$; the spread fit uses log--log least squares. Every setting has equal weight. We score A and the sum in B against the original member loss, using RMSE in its original units. B has two additional fitted spread coefficients. Paired comparisons use the same settings, floor, and held-out sizes. Cross-validation omits an entire character width, Pythia size, or Qwen pair and refits both components; it never adds the omitted setting's measured spread to its prediction. Table~\ref{tab:scaling-law-comparison} gives the floors and includes a freely fitted character-floor control with $0\leq E\leq1.5$.

\paragraph{Evaluation populations.}
The character study uses the loss-law settings in Table~\ref{tab:byte-training-settings}: six widths, each at 13 budgets from 6.25 to 400 bytes per parameter, evaluated on 524,288 held-out bytes. Seeds 0--2 are retained except at width 192, which uses seeds 0--1. Its spread law is fitted on these same runs and training budgets.

Pythia uses ten final checkpoints on the 96 C4 segments, with the standard 1,024-token context and 64-token warm-up. The five smaller sizes use three true seeds; the larger sizes use main/v0 rerun pairs. Here $N$ excludes embeddings. Intermediate checkpoints lack a separate cooldown and give poorly identified data dependence, so the final-step size comparison is the clearest test.

Qwen uses five dense Qwen2.5/Qwen3 pairs on C4 and recent text separately, with the geometric mean of total member counts. These pairs mix generations. Their corrected spread remains a descriptive estimate, and each corpus uses its own stated floor.

\begin{preprintfigure}[tbp]
\centering\includegraphics[width=\linewidth]{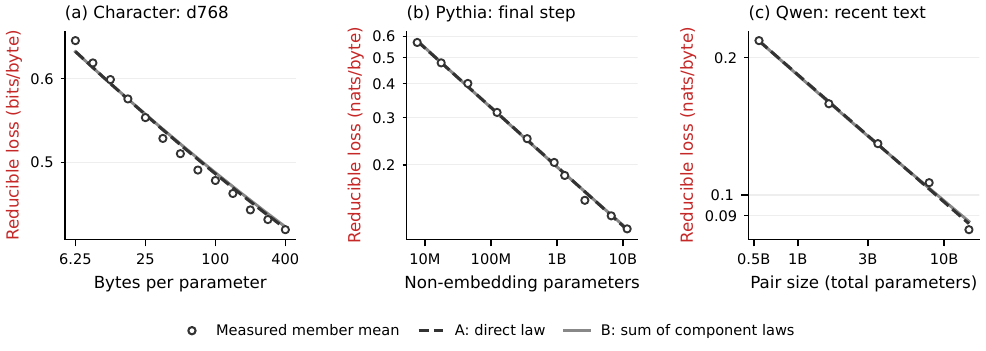}
\caption{\textbf{Direct and component laws make similar member-loss predictions.} (a): Character-level model at hidden width $d=768$ (labelled d768) across training budgets. (b): Pythia's ten final checkpoints, including the larger rerun pairs. (c): the five dense Qwen pairs on recent text. Points are measured mean member losses minus the stated irreducible loss. Curves predict that same quantity: A directly, B by adding its independently fitted pooled log loss and spread. Logarithmic axes show deviations from simple power laws; Table~\ref{tab:scaling-law-comparison} evaluates errors on untransformed member loss and adds held-out-size tests.}
\label{fig:scaling-law-comparison}
\end{preprintfigure}

\begin{table}[tbp]
\centering
\caption{Direct and component fits have similar predictive errors, without a consistent winner. RMSE is evaluated on the same original member-mean loss for A and B. B uses geometric spread with a size-only law. Held-out errors pool all omitted cells; both components are refitted without the held-out size. Character models use 78 FineWeb-Edu settings, with losses in bits/byte. The fits with freely estimated irreducible loss share bounds $0\leq E\leq1.5$. Public-model errors are in nats/byte.}
\label{tab:scaling-law-comparison}
\small
\setlength{\tabcolsep}{4pt}
\begin{tabular}{@{}lrrrrrr@{}}
\toprule
Cohort & $E$ & Cells & A: fit & B: fit & A: held out & B: held out \\
\midrule
\multicolumn{7}{l}{Character models: bits/byte, leave one of six widths out} \\
Character, $N,D$ & 0.593341 & 78 & 0.0193 & 0.0198 & 0.0290 & 0.0294 \\
Character, $N,D$ & 0.637650 & 78 & 0.0200 & 0.0203 & 0.0303 & 0.0310 \\
Character, $N,D$ & free & 78 & 0.0191 & 0.0197 & 0.0304 & 0.0301 \\
\midrule
\multicolumn{7}{l}{Pythia: nats/byte, leave one of ten sizes out} \\
Final, size only & 0.436514 & 10 & 0.0056 & 0.0061 & 0.0075 & 0.0085 \\
\midrule
\multicolumn{7}{l}{Qwen2.5/3: nats/byte, leave one pair out} \\
C4, dense & 0.436514 & 5 & 0.0028 & 0.0035 & 0.0047 & 0.0052 \\
Recent, dense & 0.333728 & 5 & 0.0020 & 0.0022 & 0.0032 & 0.0032 \\
\bottomrule
\end{tabular}
\end{table}

The two descriptions give similar finite-range member-loss predictions, without a consistent advantage in held-out error. Removing the slowly declining spread changes the fitted size exponent even when the summed prediction barely changes. This motivates measuring spread separately, while leaving the form of the invisible portion unresolved. In particular, two power laws for total loss and spread do not force their difference to be a power law, and their unequal exponents cannot hold indefinitely if that difference would become negative.

\section{Context-dependent uncertainty}
\label{sec:uncertainty-details}

\subsection{Target and scores}
\label{sec:uncertainty-target}
Figure~\ref{fig:noise-validation} asks whether a score can identify models' own epistemic error. Since the true next-token distribution is unknown, we use a stronger reference $r_x$ and measure
\begin{equation}
 e_N(x)=\frac1S\sum_{s=1}^S\KL(r_x\Vert q_{N,s,x}).
 \label{eq:reference-context-gap}
\end{equation}
This is the members' mean reducible loss if outcomes follow the reference.
The scores are one member's entropy $h_s(x)=H(q_{N,s,x})$ and geometric spread $g(x)=\tfrac{S}{S-1}[-\log Z_{N,x}]$, using the correction in Section~\ref{sec:ensembles}. Both depend only on predictions at the given context. Arithmetic spread also requires a next token or target distribution, so it is not used as a predictor here. Entropy uses character seed 0, the default Pythia seed (seed 1 at 160M), DataDecide's default seed, Qwen2.5, or Delphi seed 0. The hidden-state probe uses the same member.

\subsection{Models, text, and references}
\label{sec:uncertainty-data}
Table~\ref{tab:uncertainty-design} gives the evaluation populations for Figures~\ref{fig:noise-validation} and~\ref{fig:context-uncertainty-prediction}. Character students use the loss-law settings in Table~\ref{tab:byte-training-settings} and are trained at a common ratio of bytes to analytic parameters; model size and training data therefore grow together. Pythia students instead share a fixed training budget, with memberships as in Appendix~\ref{sec:context-protocol}. DataDecide omits the largest available size because reference quality becomes less certain as the students approach it. Qwen uses the dense base-model pairs with geometric mean parameter counts. Each family uses the reference and captured context populations listed in Table~\ref{tab:uncertainty-design}.

\begin{table}[htbp]
\centering\small
\caption{\textbf{Data for context-dependent uncertainty.} Sizes count total parameters. Each window contributes 32 sampled targets after at least 64 context bytes or tokens. Character models use 512-byte FineWeb-Edu windows; public models use at most 1,024 recent-text tokens. Character students use 50 training bytes per analytic parameter; Pythia uses 300B training tokens.}
\label{tab:uncertainty-design}
\setlength{\tabcolsep}{4pt}
\begin{tabular}{@{}lrlrr@{}}
\toprule
Students & $S$ & Reference & Windows & Train/test contexts \\
\midrule
Character, 0.86--50.7M & 3 & BLT-7B & 256 & 4,096/4,096 \\
Pythia, 14--410M & 3 & MPT-30B & 96 & 1,536/1,280 \\
DataDecide C4, 6.96--597.7M & 3 & OLMo-7B & 96 & 1,536/1,280 \\
Qwen, 0.5/0.6--14/14B & 2 & Qwen2.5-72B & 96 & 1,536/1,280 \\
Delphi, 447M--9.7B & 3 & Delphi 25B & 96 & 1,536/1,280 \\
\bottomrule
\end{tabular}
\end{table}

The character-model column in Figure~\ref{fig:noise-validation} uses held-out FineWeb-Edu text, where alternating source regions define fitting and test halves. Pythia, DataDecide, Qwen, and Delphi use the recent web text described in Appendix~\ref{sec:evaluation-corpora}; the public-model protocol alternates blocks of four complete windows or documents. MPT-30B was previously selected using next-token loss on the first 16 recent-text windows, so we omit these windows from scoring in all recent-text evaluations. The 48 fitting windows remain unchanged, while the test set contains the remaining 40 windows, or 1,280 contexts per family. Within each family and corpus, fractions and correlations in Figure~\ref{fig:noise-validation} and prediction scores in Figure~\ref{fig:context-uncertainty-prediction} use the same test contexts, fixed across model sizes.

Panels (f,l) of Figure~\ref{fig:noise-validation} reuse the public-family recent-text evaluations in columns (b--e,h--k). Pythia, DataDecide, Qwen, and Delphi use the same 40 scoring windows or documents, excluding the reference-selection pilot. Each family scores 32 token positions per window, giving 1,280 held-out contexts. Token positions and raw prefixes depend on the tokenizer; within each family they remain fixed across model sizes and panels. Reference distributions and output vocabularies remain family-specific, so these are descriptive comparisons rather than a pooled cross-family scaling fit.

References are fixed before fitting uncertainty predictors, and the reported test contexts do not enter reference selection. We chose BLT-7B for the character models based on its lower loss than EvaByte on held-out FineWeb-Edu \citep{pagnoni2024blt,evabyte2025}. MPT-30B improves on GPT-NeoX-20B and DCLM-7B on the reserved recent-text pilot \citep{mpt30b2023,dclm2024}. The original OLMo-7B remains the DataDecide reference \citep{olmo}.

We align output vocabularies before computing KL. BLT predictions are conditioned on ordinary byte outputs; public-model predictions retain assigned token IDs and exclude unassigned padding. Qwen retains shared special tokens. Public references use FP32 inference and FP64 reductions; native-forward checks verify the normalized probabilities, and BLT also passes causal-attention checks. Conditioning matters because a byte model cannot predict BLT's special tokens.

\subsection{Correlations and prediction}
\label{sec:component-correlations}
\label{sec:uncertainty-single-member}
Spearman correlations in Figure~\ref{fig:noise-validation}(g--l) are computed separately at each size, without fitting a predictor, on the test contexts in Table~\ref{tab:uncertainty-design}. The visible-fraction panels and the prediction panels in Figure~\ref{fig:context-uncertainty-prediction} use these same contexts within each family. A score can rank contexts well while giving poor numerical estimates, so we evaluate prediction separately from ranking.

\subsubsection{Fitting and evaluation}
\label{sec:uncertainty-protocol}
Each scalar score uses an affine least-squares fit. The hidden-state probe adapts the regression-head procedure of \citet{ahdritz}: an affine head is trained with Adam to predict reference KL. Validation chooses between the middle and final hidden layers and selects the stopping epoch. The final state is taken after output normalization; the middle state follows $\lceil h/2\rceil$ of the model's $h$ transformer blocks. Table~\ref{tab:uncertainty-probe} gives the fitting settings.

\begin{table}[htbp]
\centering\small
\caption{\textbf{Hidden-state probe settings.} Text groups remain intact in validation. The probe reserves one training fold for validation and is not refitted afterward. No test target selects the layer, stopping epoch, or coefficients.}
\label{tab:uncertainty-probe}
\begin{tabular}{@{}lr@{}}
\toprule
Setting & Value \\
\midrule
Loss & Mean squared error \\
Adam learning rate & $10^{-5}$ \\
Batch size & 64 \\
Training folds / validation folds & 3 / 1 \\
Patience / epoch cap & 100 / 100,000 \\
Initialization seed & 42 \\
Character fitting / validation contexts & 3,072 / 1,024 \\
Public-model fitting / validation contexts & 1,152 / 384 \\
\bottomrule
\end{tabular}
\end{table}

We report $R^2=1-\operatorname{MSE}/\operatorname{Var}(e_N)$, with both terms evaluated on the test contexts.

\begin{preprintfigure}[htbp]
\centering\includegraphics[width=\linewidth]{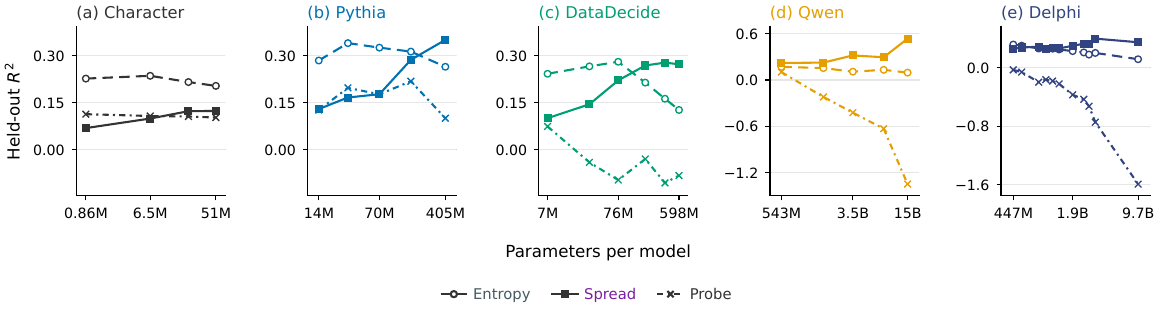}
\caption{\textbf{Predicting context-dependent epistemic uncertainty.} Held-out $R^2$ for predicting reference-based reducible loss from single-model \closs{entropy}, corrected geometric \cvis{spread}, or a hidden-state linear probe. Entropy and spread use affine least-squares fits; the probe is trained with Adam following the protocol adapted from \citet{ahdritz}. (a) Character models on 4,096 FineWeb-Edu test contexts. (b--e) Pythia, DataDecide, Qwen, and Delphi on 1,280 recent-text test contexts per family. Each family uses the same test contexts as Figure~\ref{fig:noise-validation}, fixed across sizes. Negative scores are retained; vertical scales differ across panels.}
\label{fig:context-uncertainty-prediction}
\end{preprintfigure}

Figure~\ref{fig:context-uncertainty-prediction} shows that spread becomes more effective than entropy at larger public-model sizes. At the largest Pythia, DataDecide, Qwen, and Delphi settings, spread yields $R^2=0.351$, $0.273$, $0.536$, and $0.347$, respectively, compared with $0.265$, $0.127$, $0.096$, and $0.116$ for entropy and $0.100$, $-0.083$, $-1.349$, and $-1.591$ for the probe. Character-model entropy remains more accurate than spread at every measured size. Negative probe scores mean worse squared error than predicting the test-set mean. These comparisons condition on the fixed supervision budget and fitting procedure.

\subsection{Recovered fractions across contexts}
\label{sec:context-recovered-share}
Figure~\ref{fig:noise-validation}(a--f) uses the same held-out contexts, model groups, and fixed references as the corresponding correlation panels. The individual-family columns use 4,096 FineWeb-Edu contexts for character models and 1,280 recent-text contexts per public family. The rightmost column reuses the same 1,280 recent-text test contexts per public family as the individual-family columns. Each window contributes 32 positions. All test contexts enter these descriptive fractions; no predictor is fitted, and no context is excluded for having a small positive denominator.

Let $g_N(x)=\tfrac{S}{S-1}[-\log Z_{N,x}]$ denote the corrected \cvis{geometric spread} used in Figure~\ref{fig:context-uncertainty-prediction}. For $e_N(x)>0$, the corrected recovered fraction and finite-ensemble identity are
\begin{equation}
 f_N(x)=\frac{g_N(x)}{e_N(x)},\qquad
 e_N(x)=\frac{S-1}{S}g_N(x)+\KL(r_x\Vert q_{S,N,x}^{\mathrm G}).
 \label{eq:context-recovered-share}
\end{equation}
The solid curve reports the ratio of mean corrected spread to mean reference-based reducible loss:
\begin{equation}
 F_N=\frac{\EE_X g_N(X)}{\EE_X e_N(X)}
 =\frac{\EE_X[e_N(X)f_N(X)]}{\EE_X e_N(X)}.
 \label{eq:context-aggregate-share}
\end{equation}
Thus the aggregate is a reducible-loss-weighted mean of contextwise fractions. If the reference equals the true conditional distribution and the contexts, members, and loss units match the loss evaluation, $\EE_X e_N(X)=\bar L_S-E$, so this ratio agrees with the corrected aggregate fraction in Equation~\eqref{eq:component-observations} at the true irreducible loss.

The light bands are percentile $95\%$ intervals from 2,000 bootstrap resamples of complete 32-position test windows, with the same resampled windows across sizes within each family. Each resample recomputes the ratio of mean spread to mean reducible loss. The intervals condition on the chosen members and reference. Panel (f) of Figure~\ref{fig:noise-validation} repeats the public-family ratios and intervals from panels (b--e). The finite-member correction multiplies spread by $1.5$ for the three-member ensembles and by $2$ for the Qwen pairs, and applies to every plotted fraction and interval. Corrected fractions need not lie below one.

\section{Ensembling versus larger models}
\label{sec:implication-details}

\subsection{Proof of the crossover claim}
\label{sec:ensemble-crossover-proof}
Minimizing $L(N,D)$ at fixed $ND$ gives $\alpha aN^{-\alpha}=\beta bD^{-\beta}$. With $\alpha=\beta$, the size and data terms are equal: $bD^{-\alpha}=aN^{-\alpha}$. Fix $S_2>S_1\geq1$, set $r=S_2/S_1$ and $u_i=1-S_i^{-1}$, and let $(N,D)$ be the compute-optimal member allocation of the $S_2$-member ensemble.

In scenario~(1) of Section~\ref{sec:ensemble-economics}, each of the $S_1$ members uses $(rN,D)$, matching total parameters and training compute because $S_1rN=S_2N$. In scenario~(2), each uses $(\sqrt rN,\sqrt rD)$, preserving compute optimality and matching total training compute because $S_1(\sqrt rN)(\sqrt rD)=S_2ND$. The reductions in mean member loss relative to the $S_2$ ensemble's members are therefore
\[
 G_1=a(1-r^{-\alpha})N^{-\alpha},
 \qquad G_2=2a(1-r^{-\alpha/2})N^{-\alpha}.
\]
The $S_2$ ensemble's additional spread relative to the $S_1$ ensemble is
\[
 H_1=c_P(u_2-u_1r^{-\gamma_P})N^{-\gamma_P},
 \qquad H_2=c_P(u_2-u_1r^{-\gamma_P/2})N^{-\gamma_P}.
\]
Thus the loss of the $S_2$ ensemble minus that of the $S_1$ ensemble is $G_j-H_j$ in scenario $j$. Both coefficients of $N^{-\gamma_P}$ are positive since $u_2>u_1\geq0$ and $r>1$. Since $\alpha>\gamma_P$, the inequality $H_j>G_j$ holds exactly when $N>N_j^\star$, with the finite positive thresholds in Equation~\eqref{eq:threshold}. Finally,
\[
 \left(\frac{N_2^\star}{N_1^\star}\right)^{\alpha-\gamma_P}
 =\frac{2}{1+r^{-\alpha/2}}
  \frac{u_2-u_1r^{-\gamma_P}}{u_2-u_1r^{-\gamma_P/2}}>1.
\]
The first factor exceeds one, and the second is at least one, so $N_2^\star>N_1^\star$. Setting $S_1=1$ gives $u_1=0$ and $r=S_2$, recovering the comparison with a single model. Hence every fixed larger ensemble size eventually outperforms the smaller one under the assumed laws, with the crossover delayed in scenario~(2). This ordering requires the compute-optimal balance; proportional growth of data with size alone is insufficient. $\square$

\subsection{Data and fits for Figure~\ref{fig:economics}}
\label{sec:pair-variation}
We evaluate actual finite geometric pools, without applying the population correction to their losses. Pythia includes every pair and triple of the retained independent runs in Table~\ref{tab:primary-seeds}. The character models and DataDecide C4 use all pairs and the triple at each displayed size. DataDecide retains the full seven-size singleton fit, including the three smaller sizes outside the displayed range. Qwen has one generation pair at each size and a separate singleton reference for each generation. Horizontal coordinates sum the members' actual parameter counts. Character losses use FineWeb-Edu, while public-model losses use recent text, so absolute losses should be compared within each family.

The increasing-data character trajectory uses 100 training bytes per analytic parameter. For its fixed-data panel, Table~\ref{tab:character-fixed-data-ensembles} gives the shared budget and losses. The 0.858M and 6.461M models each use three completed continuations to the exact shared budget, evaluated on the same held-out bytes as the larger models. One intermediate size requires interpolation in $\log D$ between recorded checkpoints; no data-budget extrapolation is used. We retain departures from the smooth trends, including Pythia's 70M models and the character models near 6.5M parameters.

\begin{table}[htbp]
\centering\footnotesize\setlength{\tabcolsep}{3pt}
\begin{tabular}{@{}rlrrr@{}}
\toprule
Member $N$ (M) & Source & $L_1\pm\mathrm{SD}$ & $L_2\pm\mathrm{SD}$ & $L_3$ \\
\midrule
0.858 & Measured & $1.146265\,\pm\,0.001574$ & $1.120932\,\pm\,0.000701$ & $1.112706$ \\
6.461 & Measured & $0.941672\,\pm\,0.000186$ & $0.923487\,\pm\,0.000264$ & $0.917532$ \\
21.441 & Measured & $0.859157\,\pm\,0.000824$ & $0.843909\,\pm\,0.000443$ & $0.838903$ \\
50.678 & Interpolated & $0.812848\,\pm\,0.000357$ & $0.799516\,\pm\,0.000199$ & $0.795133$ \\
170.303 & Measured & $0.764301\,\pm\,0.001086$ & $0.752760\,\pm\,0.000440$ & $0.748968$ \\
\bottomrule
\end{tabular}

\caption{\textbf{Character ensembles at fixed training data.} Each member sees $D_0=8{,}493{,}465{,}600$ bytes. Losses are nats/byte, with sample SD across members or pairs; the full triple has no subset-choice SD. The 50.7M losses are interpolated in $\log D$ between $7{,}117{,}996{,}032$ and $10{,}066{,}329{,}600$ bytes. Overlapping pairs share models, so these SDs describe subset choice rather than independent replications.}
\label{tab:character-fixed-data-ensembles}
\end{table}

\paragraph{Single-model references.}
The upper curves fit $H_S(B)=E_S+A_S(B/10^9)^{-\eta_S}$, where $B$ is total parameters. We use equal weight per size and least squares on log excess loss. Fixed-data gains compare the measured ensemble with $H_1(B)$. For the increasing-data panels, the reference instead fits singleton loss against training cost, $J_1(C)=E+A(C/C_0)^{-\eta}$, and evaluates it at the ensemble's exact cost $C=\sum_sN_sD_s$. Table~\ref{tab:economics-fit-settings} gives the numerical settings. These cost references generally have different parameter counts from the ensemble, so the lower markers are not vertical gaps between the upper curves.

\begin{table}[htbp]
\centering\footnotesize
\caption{\textbf{Reference settings for the ensemble-budget comparisons.} Irreducible-loss settings are in nats/byte. The fixed-data character settings come from the corresponding size--data fits. $A$ and $\eta$ are fitted separately for each upper curve; the cost-reference coefficients are listed explicitly. Character cost uses $C_0=10^{18}$ parameter-bytes; DataDecide uses $10^{19}$ parameter-tokens.}
\label{tab:economics-fit-settings}
\begin{tabular}{@{}lrrr@{}}
\toprule
Curve & $E$ & $A$ & $\eta$ \\
\midrule
Public-model upper curves & 0.33372841 & --- & --- \\
Character increasing-data upper curves & 0.41127265 & --- & --- \\
Character fixed data, $S=1$ & 0.65696935 & --- & --- \\
Character fixed data, $S=2$ & 0.65553698 & --- & --- \\
Character fixed data, $S=3$ & 0.65496251 & --- & --- \\
Character singleton cost reference & 0.41127265 & 0.366901 & 0.092751 \\
DataDecide singleton cost reference & 0.33372841 & 0.240116 & 0.128495 \\
\bottomrule
\end{tabular}
\end{table}

\paragraph{Predicted ensemble gains.}
The lower curves use the spread law separately from the upper ensemble-loss fits. We divide each finite spread by $1-S^{-1}$ and fit a common law $c(x/x_0)^{-\gamma}$ jointly to pairs and triples, with equal weight per size and ensemble count in log space. Together with the corresponding singleton law, this gives
\[
 \widehat\Delta_S(x)
 =(1-S^{-1})c(x/x_0)^{-\gamma}
 -A(1-S^{-\eta})(x/x_0)^{-\eta}.
\]
Here $x$ is member size at fixed data, or member training cost for the increasing-data panels. The first term is the pooling gain; the second is the improvement from spending the same budget on one larger model. A fitted map from member size to cost places the latter curves on the total-parameter axis; markers retain exact recorded costs.

\paragraph{Floor sensitivity.}
Although $E$ cancels explicitly from a difference of fitted losses, changing it changes the refitted amplitudes and exponents. Figure~\ref{fig:economics} propagates the retained public floor range $[0.26019,0.33372841]$ nats/byte and the separate character range $[0.593341,0.63764990]$ bits/byte from Appendices~\ref{sec:component-share} and~\ref{sec:byte-floor}. At each candidate floor, we refit the size and cost laws on their original cohorts with the same objectives. For fixed-data characters, we first refit each ensemble size's size--data law on the original 65 cells, then recompute its finite-data offset $E_{D,S}=E+b_SD_0^{-\beta_S}$ at the shared budget $D_0$ before refitting the five size centers. The central curves retain the settings in Table~\ref{tab:economics-fit-settings}.

Shading spans the pointwise minimum and maximum refitted curves. For each predicted gap, we evaluate the complete expression above at each floor before taking its envelope, holding the independently fitted spread law and allocation map fixed. Vertical intervals instead vary the fitted singleton reference while keeping each measured ensemble loss fixed, using exact recorded $ND$ for the increasing-data panels. Upper loss markers are independent of the floor. The public range comes from a nominal 95\% conditional floor profile, while the character range is a reference-to-cap sensitivity range; these propagated envelopes are not full confidence intervals. They omit residual fit uncertainty at a fixed floor, seed and text sampling, interpolation, and model-form uncertainty.

These are empirical budget comparisons. DataDecide's data allocation is chosen for overtraining, and Qwen generations differ in data and training recipe \citep{datadecide,qwen25technical,qwen3}. They do not locate the theoretical crossover sizes or establish their ordering. Dashed segments extrapolate, and the largest fixed-data character ensembles require a singleton reference beyond the measured size range. The five-size fixed-data character fits have three residual degrees of freedom each, conditional on the finite-data offset at each candidate floor. Matching parameters or $ND$ also does not establish equal runtime or latency.

\paragraph{Numerical crossover examples.}
Uncertainty in the irreducible loss changes the refitted loss amplitudes and exponents, and hence the crossover predictions from Equation~\eqref{eq:threshold} and its empirical counterpart at matched $ND$. Varying $E\in[0.26019,0.33373]$ nats/byte while retaining the fitted spread laws gives a pair-versus-single crossover ($S_1=1,S_2=2$) at approximately $[2.9\times10^{12},\,1.6\times10^{15}]$ parameters per Pythia member at fixed $D$ and $[2.0\times10^{10},\,7.5\times10^{11}]$ per DataDecide C4 member at matched $ND$. For a triple ensemble to outperform a pair ($S_1=2,S_2=3$), the corresponding crossover thresholds are approximately $[1.7\times10^{14},\,9.8\times10^{17}]$ parameters per Pythia triplet member and $[1.3\times10^{12},\,6.2\times10^{14}]$ per DataDecide C4 triplet member, respectively. These are ranges of the threshold, above which the larger ensemble has lower predicted loss, with $N$ always counting parameters per member of the larger ensemble. The DataDecide ranges follow its fitted training allocation and do not establish compute optimality. All ranges describe conditional floor sensitivity and omit the other fitting and extrapolation uncertainties.

\subsection{A measured comparison under both budget constraints}
\label{sec:datadecide-fixed-data}
\label{sec:datadecide-matched-compute}
We compare the same preselected DataDecide C4 pair with two larger checkpoints: one trained on the same data budget, and one given more data at nearly the same total $ND$. Table~\ref{tab:datadecide-budget-comparisons} gives the allocations and losses. Neither member individually beats the fixed-data reference. Pooling creates that advantage, while allowing the larger model more data reduces it to an unresolved near tie. This illustrates the stronger reference in scenario~(2), without identifying a scale crossover.

The pair uses the nominal 300M \texttt{default} and \texttt{small-aux-2} runs at step 45,000. The references are the nominal 750M default at step 25,000 and 600M default at step 40,000. Checkpoints and seeds were fixed before evaluation. All use the common recent-text evaluation in Appendix~\ref{sec:evaluation-corpora}, float32 inference, checkpoint RMS-normalization epsilon $10^{-6}$, and the full output vocabulary. Paired intervals resample 24 contiguous four-document groups 5,000 times and recompute summed loss divided by summed bytes. They quantify text variation conditional on these models.

\begin{table}[htbp]
\centering\footnotesize\setlength{\tabcolsep}{2pt}
\begin{tabular}{@{}lrrrrrl@{}}
\toprule
Constraint & Ref. (M) & $D$ (B) & Cost ratio & Single & Gain & 95\% interval \\
\midrule
Fixed data & 758.6 & 29.4912 & 1.02097 & 0.565021 & +5.616 & $[+3.502,+7.656]$ \\
Matched cost & 597.7 & 36.7002 & 1.00108 & 0.559427 & +0.023 & $[-1.198,+1.392]$ \\
\bottomrule
\end{tabular}

\caption{\textbf{The same C4 pair under two budget constraints.} Members have 371.492M parameters each; their geometric-pool loss is 0.559404 nats/byte. $D$ is reference training tokens in billions. Cost ratio is reference $ND$ divided by summed pair $ND$. Single loss is nats/byte; gain and its paired 95\% interval are milli-nats/byte. Positive gain favors the pair.}
\label{tab:datadecide-budget-comparisons}
\end{table}

The models are at different stages of their published learning-rate schedules, and the references were not separately annealed for these budgets. The comparison therefore concerns available checkpoints rather than models individually optimized for each constraint.

\subsection{Specialized-domain comparisons}
\label{sec:specialized-economics}
Figure~\ref{fig:economics-specialized} repeats the budget comparison on the specialized datasets, which may not be strictly in distribution. 
We see that many ensembles are able to outperform single models, under two different matched budgets in these settings, where spreads are generally higher; see Figure~\ref{fig:ood-seed-spread} for how spreads change in these domains.

\begin{preprintfigure}[p]
\centering
\includegraphics[width=\linewidth,height=.84\textheight,keepaspectratio]{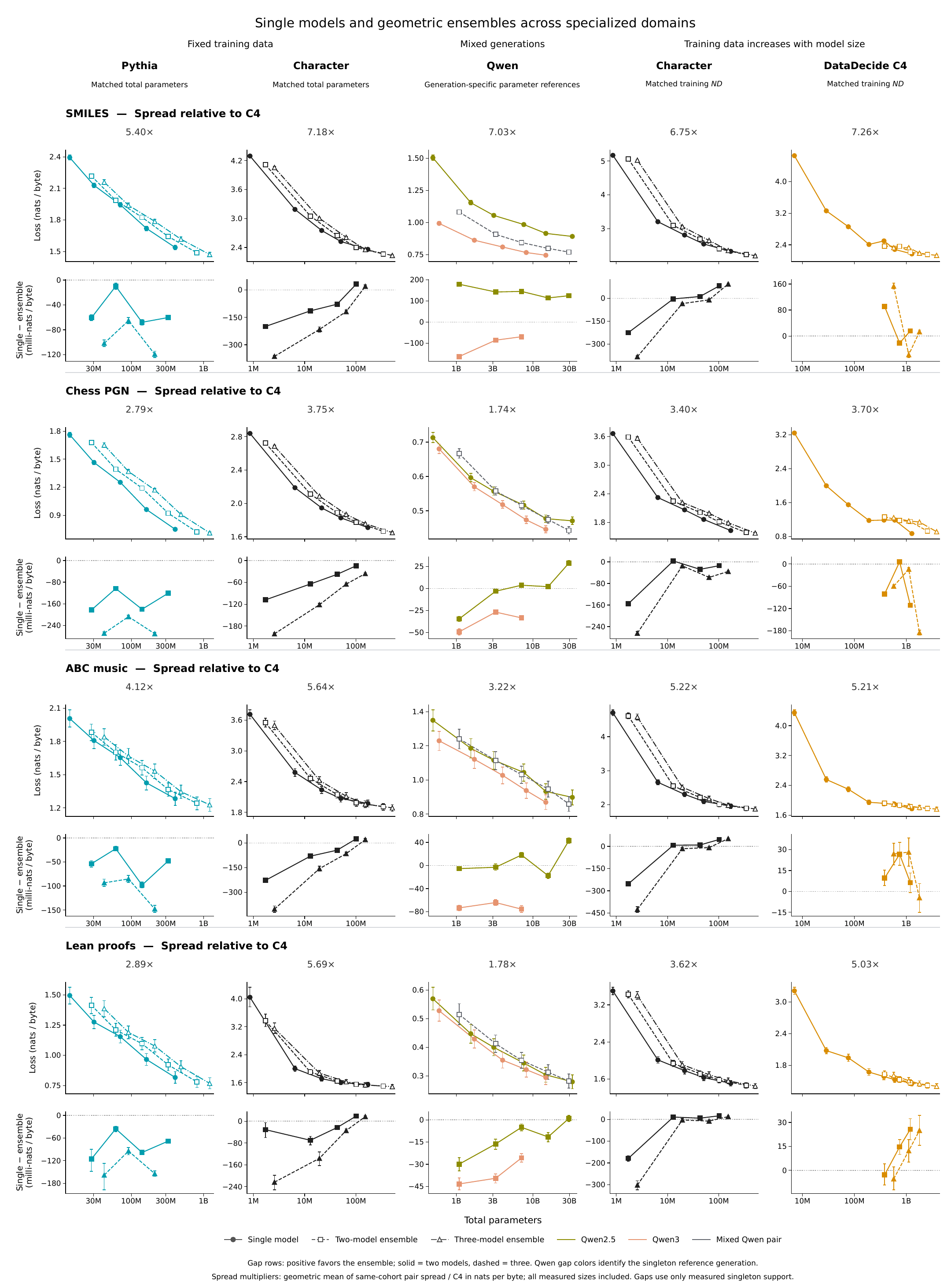}
\caption{\textbf{Single models and geometric ensembles on specialized domains.} Each dataset occupies two rows: measured losses above, and matched-budget single-model loss minus ensemble loss below (positive favors ensembling). Singleton references use within-support linear interpolation in $\log N$ or $\log(ND)$, without a fitted loss floor; bars are paired passage-bootstrap $95\%$ intervals. Multipliers are geometric means, across matching model sizes, of per-byte pair-spread ratios to C4. Qwen references are generation-specific and match parameters only; training budgets and recipes differ.}
\label{fig:economics-specialized}
\end{preprintfigure}
\FloatBarrier

\section{Shared-training and LoRA ensemble studies}
\label{sec:branching-details}

\subsection{Matched data and training states}
We use the spread and branching settings in Table~\ref{tab:byte-training-settings}. Table~\ref{tab:branching-retention} lists the sizes. Let $t=D/N$ be the evaluation budget and $b$ the branch point, both in bytes per parameter. Learning rates, batch sizes, and cooldown schedules are matched between each parent, its continuation, and its independent reference.

After an 18-step warmup, each parent follows a constant-learning-rate trajectory through 160 bytes per parameter. A decay continuation adds one quarter of the preceding trajectory, ending at $T=200$. All branch points in Section~\ref{sec:branching} occur after warmup and before this decay. A branch resumes saved weights and optimizer moments, with no reset or additional warmup. It permutes the remaining stream in blocks chosen to preserve the corpus positions consumed at every evaluation endpoint. Both members therefore see the same data, with different orders after branching.

The parent uses initialization seed 0 and order seed 1; the independent reference uses initialization seed 1 and order seed 2; branches use order seed 7. Both parent and reference have permuted streams from the start. The initialization-only comparison in Figure~\ref{fig:branching-law} instead averages the original seeds' pair spreads on their common sequential stream; it is not the denominator of $R$.

\subsection{Evaluation and fitted laws}
The checks in Tables~\ref{tab:branching-retention} and~\ref{tab:branching-fits} and Figure~\ref{fig:branching-lora} use 1,024 held-out FineWeb-Edu sequences of 512 bytes. Panels (a,b) of Figure~\ref{fig:branching-law} evaluate the same trained checkpoints on 432,878 recent-text bytes. Both evaluations display $2\rho_2^{\mathrm G}$ using the pair correction. This factor cancels in $R$; correlated branches do not become independent population samples through rescaling. Table~\ref{tab:branching-retention} retains every width at the final budget.

\begin{table}[htbp]
\centering\small
\caption{\textbf{Retained pair spread at 200 bytes per parameter.} Evaluation uses held-out FineWeb-Edu text. Branch columns give $R$ from Section~\ref{sec:branching}. The independent column reports $2\rho_{2,\mathrm{ind}}^{\mathrm G}$ in milli-bits/byte.}
\label{tab:branching-retention}
\begin{tabular}{@{}rrrrrrr@{}}
\toprule
Width & $N$ (M) & Independent & $b=1$ & $b=5$ & $b=25$ & $b=125$ \\
\midrule
128 & 0.79 & 76.72 & 1.043 & 0.971 & 0.722 & 0.281 \\
192 & 2.65 & 60.65 & 0.899 & 0.837 & 0.659 & 0.247 \\
256 & 6.29 & 51.21 & 0.949 & 0.885 & 0.700 & 0.284 \\
384 & 21.23 & 43.90 & 0.944 & 0.866 & 0.687 & 0.307 \\
512 & 50.33 & 40.04 & 0.926 & 0.846 & 0.678 & 0.318 \\
768 & 169.87 & 34.12 & 0.905 & 0.830 & 0.688 & 0.383 \\
\bottomrule
\end{tabular}
\end{table}

We fit the size laws by least squares in log space on the same five widths, excluding the smallest. Its earliest branch has more spread than the independent pair, so its lost share cannot enter a log fit. The lost-share law uses all branch points at the retained widths. Table~\ref{tab:branching-fits} gives the fits. Early branches roughly retain the independent size slope; the latest branch has a shallower slope.

\begin{table}[htbp]
\centering\small
\caption{\textbf{Final-budget branching fits on FineWeb-Edu.} Size fits use $2\rho_2^{\mathrm G}=cN^{-\gamma}$ over widths 192--768. Uncertainties are residual regression SEs, not from variation across repeated training experiments. The lost-share fit uses 20 observations and has natural-log RMSE 0.160.}
\label{tab:branching-fits}
\begin{tabular}{@{}lrr@{}}
\toprule
Size-law population & $\gamma$ & SE \\
\midrule
Independent initialization and order & 0.134 & 0.008 \\
Initialization only & 0.127 & 0.005 \\
Branch at $b=1$ & 0.135 & 0.002 \\
Branch at $b=5$ & 0.140 & 0.001 \\
Branch at $b=25$ & 0.129 & 0.003 \\
Branch at $b=125$ & 0.039 & 0.010 \\
\midrule
\multicolumn{3}{@{}l@{}}{Lost-share fit: $1-R=0.846(b/T)^{0.467\pm0.021}$} \\
\bottomrule
\end{tabular}
\end{table}

The square-root approximation describes the measured range. It fails at complete sharing, where identical final states must have zero spread, and zero shared training is untested. Intermediate-budget checks also favor separate dependence on branch point and evaluation budget over their ratio alone. Each width has one parent and one independent reference; branch ratios share their denominator. The fitted observations are therefore not independent training replications.

Sharing a fraction $f=b/T$ saves $f/2$ of the training work for a pair, excluding the extra cooldowns used for intermediate measurements. Within the fitted range, dividing a branched-pair spread by the fitted retained share could estimate independent-pair spread. This correction has not been separately validated, and small retained shares amplify its error.

\subsection{Final-weight correlations}
\label{sec:branching-weight-correlation}
We compare each parent with its four full-parameter branches at the final checkpoint, $T=200$ bytes per analytic parameter, across all six widths. Parameters are aligned by registered name and shape, and all unique learned scalar coordinates are concatenated into vectors $w_{\mathrm P},w_{\mathrm B}\in\mathbb R^M$. Here $M$ is the total learned parameter count, including embeddings, output weights, and normalization gains; the analytic count $N$ still defines the training budget and figure legends. We omit nonparameter buffers and compute
\begin{equation}
r_w=
\frac{\sum_{i=1}^{M}(w_{\mathrm P,i}-\bar w_{\mathrm P})(w_{\mathrm B,i}-\bar w_{\mathrm B})}
{\sqrt{\sum_{i=1}^{M}(w_{\mathrm P,i}-\bar w_{\mathrm P})^2}
 \sqrt{\sum_{i=1}^{M}(w_{\mathrm B,i}-\bar w_{\mathrm B})^2}},
\label{eq:branching-weight-correlation}
\end{equation}
where each mean is taken over the complete concatenated vector. Thus every scalar coordinate enters the calculation; we do not average correlations computed separately for individual tensors. Reductions use float64 arithmetic without coordinate sampling, and no neuron or attention-head permutation alignment is performed.

Figure~\ref{fig:branching-weight-correlation} shows $r_w$ against the shared training fraction and the pair's predictive spread. The spread panel uses the same $432{,}878$ scored recent-text bytes as Figure~\ref{fig:branching-law} and displays $2\rho_2^{\mathrm G}$ in milli-bits/byte. Independent references change both initialization and data order; they have no branch point and are placed at the separate ``Ind.'' position only for display. Their coordinate correlations depend on the chosen parameterization, as do those of the branches.

\begin{preprintfigure}[htbp]
\centering\includegraphics[width=\linewidth]{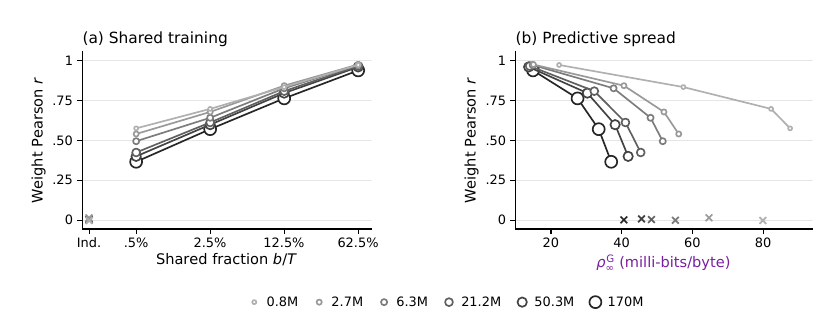}
\caption{\textbf{Later branches retain more similar final weights.} At $T=200$, Pearson correlation of the complete parameter vectors versus (a) the shared training fraction $b/T$ and (b) corrected geometric pair spread on recent text. Open circles denote parent--branch pairs; crosses denote independent initialization and data-order references, shown separately at ``Ind.'' without an assigned $b/T$. Lines join the four measured branch points. Grayscale and marker size identify analytic parameter counts $N$.}
\label{fig:branching-weight-correlation}
\end{preprintfigure}

Large-variance parameter groups can dominate this coordinate correlation. As a sensitivity check, removing embeddings, the output head, and one-dimensional normalization gains gives correlations of $0.147$--$0.216$ at $b=1$ and $0.918$--$0.964$ at $b=125$; the increasing trend remains. With all parameters included, the corresponding ranges are $0.365$--$0.575$ and $0.937$--$0.974$. These are descriptive comparisons with one parent and one continuation per branch point and width. The four branches share a parent, and the results do not estimate variation across repeated training runs.

\subsection{Diversity penalty and ensemble loss}
\label{sec:diversity-branching}
\textbf{Result across model sizes.} At $\lambda=1$, the diversity penalty increases geometric spread by $2.30$--$2.72$ times, while higher mean member loss offsets $93$--$106\%$ of the added spread. Most of the additional gain in total ensemble loss is therefore lost (Table~\ref{tab:diversity-branching-scale}; Figure~\ref{fig:diversity-branching-scale}).

We test whether explicitly rewarding disagreement can recover spread lost through shared training. We first sweep the penalty weight in the width-$192$ byte model (six layers, $2.76$M total parameters), branching at $b=125$ and finishing at $T=200$ bytes per analytic parameter. We then fix $\lambda=1$ and extend the comparison to widths $256$, $384$, and $512$, spanning $2.76$--$50.68$M total parameters. We retain each checkpoint's optimizer moments and the batch sizes, learning rates, cooldown, and data orders described above. The parent continues ordinary cross-entropy training; only the branch receives a diversity reward, evaluated on its training contexts against the parent's current predictions:
\begin{equation}
\mathcal J_\lambda=\widehat L(q_{\mathrm B})-2\lambda\widehat\rho_2^{\mathrm G},
\qquad
\widehat\rho_2^{\mathrm G}
=-\widehat{\EE}_X\log\sum_{y\in\mathcal V}
\sqrt{q_{\mathrm P,X}(y)q_{\mathrm B,X}(y)}.
\label{eq:diversity-branching-objective}
\end{equation}
Here $q_{\mathrm P}$ and $q_{\mathrm B}$ denote parent and branch predictions, and hats indicate training-minibatch averages. The parent's predictions are held constant when differentiating this objective, while its own updates remain unregularized. The negative geometric-spread term rewards predictive diversity over the full vocabulary. Its factor of two makes the branch gradient equal to that of $2[(1-\lambda)\widehat{\bar L}_2+\lambda\widehat L(q_2^{\mathrm G})]$, with the parent held constant, giving a branch-only version of the joint-training objective of \citet{webb2021ensemble}. Thus $\lambda=1$ directly minimizes geometric-pool loss with respect to the branch.

\begin{table}[htbp]
\centering\small
\begin{tabular}{@{}lrrrr@{}}
\toprule
Pair & $\bar L_2$ & $\rho_2^{\mathrm G}$ & $L(q_2^{\mathrm G})$ & $L(q_2^{\mathrm A})$ \\
\midrule
Ordinary late branch ($\lambda=0$) & 1.542097 & 0.007462 & 1.534636 & 1.534759 \\
$\lambda=0.25$ & 1.543177 & 0.008937 & 1.534241 & 1.534463 \\
$\lambda=0.50$ & 1.544696 & 0.011007 & 1.533689 & 1.534087 \\
$\lambda=1.00$ & 1.554386 & 0.020424 & 1.533962 & 1.536031 \\
\midrule
Independent seeds & 1.542116 & 0.030323 & 1.511793 & 1.514174 \\
\bottomrule
\end{tabular}

\caption{\textbf{Diversity rewards increase spread and mean member loss.} All values are bits/byte on the same $1{,}024$ held-out sequences of $512$ bytes. Losses are ordinary cross entropies, without the training reward. Spread is the raw pair quantity $\rho_2^{\mathrm G}$, so $L(q_2^{\mathrm G})=\bar L_2-\rho_2^{\mathrm G}$; the earlier branching plots display $2\rho_2^{\mathrm G}$. The $\lambda=0$ row is a fresh ordinary late-branching control. Independent seeds change both initialization and data order, with the same architecture and training budget.}
\label{tab:diversity-branching}
\end{table}

In the coefficient pilot, mean member loss and spread increase at every tested positive weight (Table~\ref{tab:diversity-branching}). At $\lambda=0.5$, spread rises by $47.5\%$ and mean member loss by $2.599$ milli-bits/byte relative to ordinary late branching, while geometric ensemble loss falls by only $0.946$ milli-bits/byte. At $\lambda=1$, spread is $2.74$ times the control, but arithmetic ensemble loss rises by $1.272$ milli-bits/byte.

The scale study reuses this $\lambda=1$ pair at width $192$ and uses the saved ordinary late-branching pairs as controls at every width. At width $192$, the saved control differs slightly from the fresh control in Table~\ref{tab:diversity-branching}. All models are evaluated on the same held-out bytes. Let $\Delta$ denote the penalized pair's value minus its ordinary late-branching control. The additional geometric ensemble gain is exactly
\begin{equation}
G^{\mathrm G}
=L(q_{2,0}^{\mathrm G})-L(q_{2,1}^{\mathrm G})
=\Delta\rho_2^{\mathrm G}-\Delta\bar L_2,
\label{eq:diversity-additional-gain}
\end{equation}
where subscripts $0$ and $1$ identify the ordinary and penalized pairs. Thus increasing spread improves total ensemble loss only to the extent that the added spread exceeds the increase in mean member loss.

\begin{table}[htbp]
\centering\small
\begin{tabular}{@{}rrrrrrr@{}}
\toprule
Width & Parameters (M) & Spread ratio & $\Delta\bar L_2$ & $\Delta\rho_2^{\mathrm G}$ & $G^{\mathrm G}$ & $G^{\mathrm A}$ \\
\midrule
192 & 2.76 & 2.724 & $+12.180$ & $+12.927$ & $+0.747$ & $-1.214$ \\
256 & 6.46 & 2.550 & $+10.493$ & $+11.278$ & $+0.785$ & $-0.916$ \\
384 & 21.44 & 2.375 & $+9.832$ & $+9.269$ & $-0.563$ & $-1.735$ \\
512 & 50.68 & 2.297 & $+7.840$ & $+8.253$ & $+0.414$ & $-0.758$ \\
\bottomrule
\end{tabular}

\caption{\textbf{Higher member loss offsets the extra spread across model sizes.} Effects of $\lambda=1$ relative to the saved ordinary late-branching control at each width, at $b=125$, $T=200$. Parameter counts are totals; spread ratio is $\rho_{2,1}^{\mathrm G}/\rho_{2,0}^{\mathrm G}$. The last four columns are milli-bits/byte. Positive $G^P=L(q_{2,0}^P)-L(q_{2,1}^P)$ means lower ensemble loss, for $P\in\{\mathrm G,\mathrm A\}$. All evaluations use the same $524{,}288$ held-out bytes.}
\label{tab:diversity-branching-scale}
\end{table}

\begin{preprintfigure}[htbp]
\centering\includegraphics[width=\linewidth]{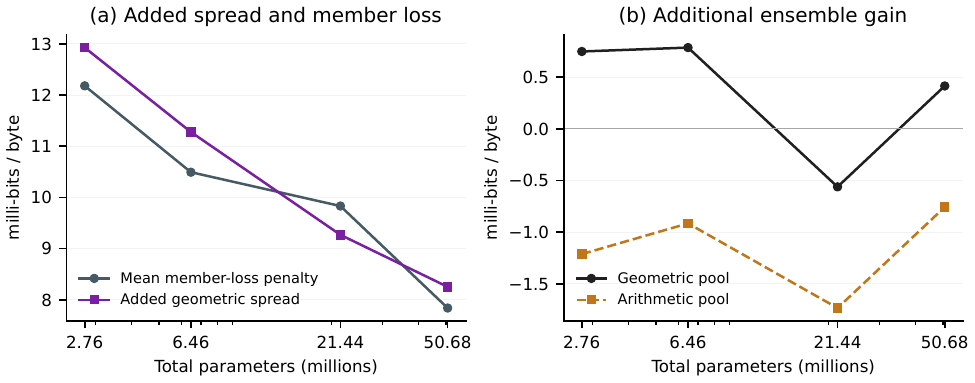}
\caption{\textbf{Most of the additional ensemble gain is lost to higher member loss.} With $\lambda=1$, (a) added raw geometric spread and the mean member-loss penalty nearly coincide; (b) the resulting geometric-pool gain is small and inconsistent across sizes, and arithmetic pooling worsens at every size. Positive gain means lower loss than ordinary late branching. All loss changes are milli-bits/byte.}
\label{fig:diversity-branching-scale}
\end{preprintfigure}

Across the four sizes, the penalty increases spread by factors of $2.30$--$2.72$, while mean member loss rises by $7.840$--$12.180$ milli-bits/byte (Table~\ref{tab:diversity-branching-scale}). This member-loss penalty offsets $93$--$106\%$ of the added geometric spread (Figure~\ref{fig:diversity-branching-scale}). Net geometric gains range from $-0.563$ to $+0.785$ milli-bits/byte, without a consistent improvement with size; arithmetic loss increases by $0.758$--$1.735$ milli-bits/byte at every size. Independent-seed ensembles still have more spread and lower pooled loss throughout. Higher mean member loss therefore largely cancels the additional pooling benefit in total ensemble loss, although pooling still improves each pair relative to its own mean member loss.

These are descriptive comparisons using one parent checkpoint and continuation order per width. Ordinary-parent loss differences from the saved references range from $-0.035$ to $+0.264$ milli-bits/byte, so numerical trajectory variation is also relevant to the small net pooling gains.

\subsection{LoRA training and rank sweep}
\label{sec:lora-details}
We freeze the parents at $b=125$ and train two adapters per rank through the same remaining data to $T=200$. Widths are those retained in the branching fits. Embeddings, output heads, and normalization parameters remain frozen. We add $W+BA$ updates to the combined query/key/value projection, attention output, and all three feed-forward matrices in every block. Entries of $A$ are sampled from $\mathcal N(0,1/n_{\mathrm{in}})$, $B$ starts at zero, and $\alpha/r=1$. Both members therefore begin with the parent's predictions.

The rank choices are $8,d/8,d/2$. Adapter seeds are 1 and 2; remaining-data order seeds are 7 and 8. The first member shares its order with the full-parameter continuation. Adapters use fresh AdamW states, learning rate 0.003, $(\beta_1,\beta_2)=(0.9,0.95)$, $\epsilon=10^{-10}$, zero weight decay, and a 100-step warmup. This rate was selected using a small bracket at the smallest retained width, without retuning at larger widths. Full-parameter branches retain their optimizers without warming up again. The comparison changes both the trainable subspace and optimization treatment, so it does not isolate rank as the cause of the spread difference.

\subsection{LoRA spread and prediction loss}
\label{sec:lora-ensembles}
\begin{preprintfigure}[htbp]
\begin{minipage}[c]{.76\linewidth}
\includegraphics[width=\linewidth]{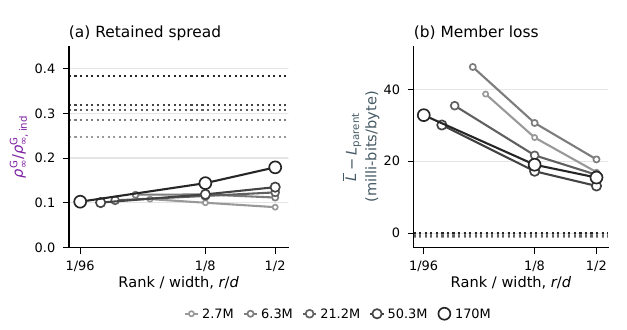}
\end{minipage}\hfill
\begin{minipage}[c]{.22\linewidth}
\captionsetup{font=footnotesize,justification=raggedright,singlelinecheck=false}
\caption{\textbf{LoRA further reduces spread on FineWeb-Edu.} At $b=125$, $T=200$: (a) retained spread; (b) mean member loss above the parent. Dotted lines show full-parameter branches.}
\label{fig:branching-lora}
\end{minipage}
\end{preprintfigure}

Table~\ref{tab:lora-ranks} summarizes the rank sweep. Increasing rank improves member loss at every width, while its effect on spread varies with width. Even the largest rank retains less than half the full-branch spread. The pooled adapters still have higher loss than one full continuation: their small pooling gain does not offset the member-loss penalty. Thus lower spread here does not establish lower total epistemic loss.
 
\begin{table}[htbp]
\centering\small
\caption{\textbf{LoRA rank sweep after shared training.} Evaluation uses held-out FineWeb-Edu text. The three rank columns give LoRA pair spread divided by independent-pair spread. The last column gives the range, across ranks, of geometric-pool loss above one full continuation, in milli-bits/byte. All loss penalties are positive.}
\label{tab:lora-ranks}
\begin{tabular}{@{}rrrrr@{}}
\toprule
Width $d$ & $r=8$ & $r=d/8$ & $r=d/2$ & Pool loss penalty \\
\midrule
192 & 0.108 & 0.100 & 0.090 & 15.30--36.47 \\
256 & 0.118 & 0.119 & 0.112 & 18.22--43.76 \\
384 & 0.105 & 0.115 & 0.123 & 14.31--33.92 \\
512 & 0.100 & 0.119 & 0.135 & 10.48--28.13 \\
768 & 0.102 & 0.144 & 0.179 & 12.34--31.02 \\
\bottomrule
\end{tabular}
\end{table}

Adapters are merged before evaluation, using the same held-out FineWeb-Edu text as the full branches. Each pair contains two adapters at the same rank. Figure~\ref{fig:branching-law}b instead uses recent text, with $r=d/8$ and a band spanning the rank sweep. There is one parent and one pair per rank at each width, so we do not fit a LoRA size law or report training-population confidence intervals. The result concerns this late branch and optimizer recipe, rather than every shared-backbone uncertainty method.

\section{C4 validation results}
\label{sec:c4-validation}
We repeat the public-model analyses in Figures~\ref{fig:champion}, \ref{fig:visible}, \ref{fig:noise-validation}, and~\ref{fig:economics} on the 96 English C4 validation passages described in Appendix~\ref{sec:evaluation-corpora}. The five-family fit to 28 settings gives a C4 irreducible-loss estimate of $E=0.38504$ nats/byte. Its profile interval is $[0.31973,0.42460]$ before applying the lowest measured C4 loss, $0.41740$ nats/byte for GLM-4.5-Base (Table~\ref{tab:evaluation-reference-losses}). We use the resulting range $[0.31973,0.41740]$ for the C4 fits.

\begin{preprintfigure}[htbp]
\centering\includegraphics[width=.90\linewidth]{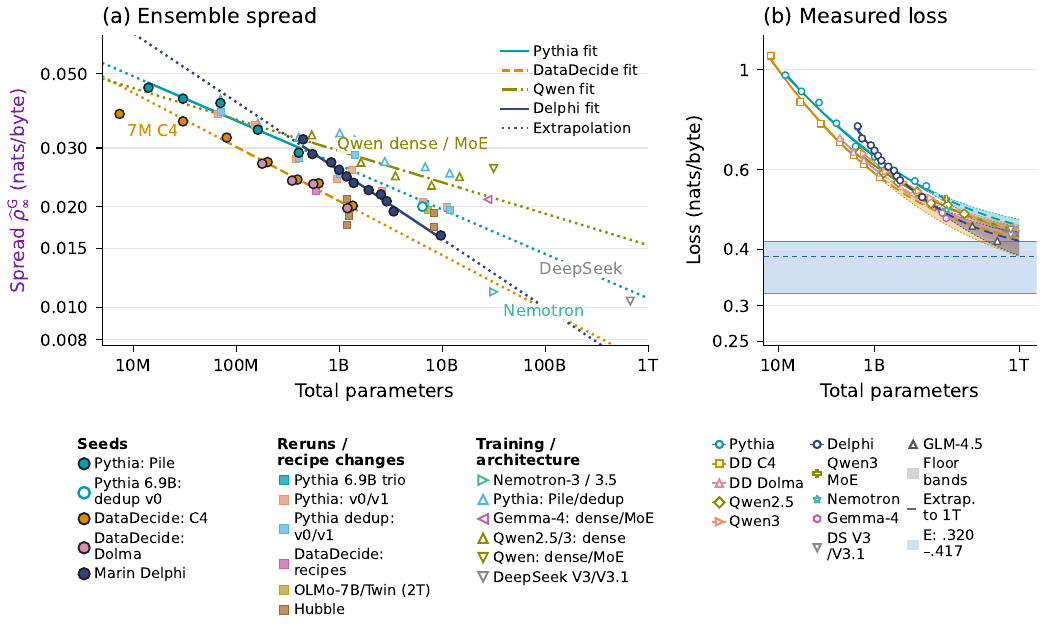}
\caption{\textbf{Ensemble spread and individual-model loss on C4} (cf.\ Figure~\ref{fig:champion}).
(a) Across 72 public comparisons, the fitted spread exponents are $0.133$ for Pythia, $0.162$ for DataDecide, $0.094$ for dense Qwen, $0.205$ for Delphi. (b) Of the 75 checkpoints with C4 scores, 66 enter the six family loss fits. Their exponents range from $0.196$ to $0.307$. Kimi-K2-Base and DeepSeek-V4-Pro-Base have scores only on recent text.}
\label{fig:c4-size}
\end{preprintfigure}

\begin{preprintfigure}[htbp]
\centering\includegraphics[width=.91\linewidth]{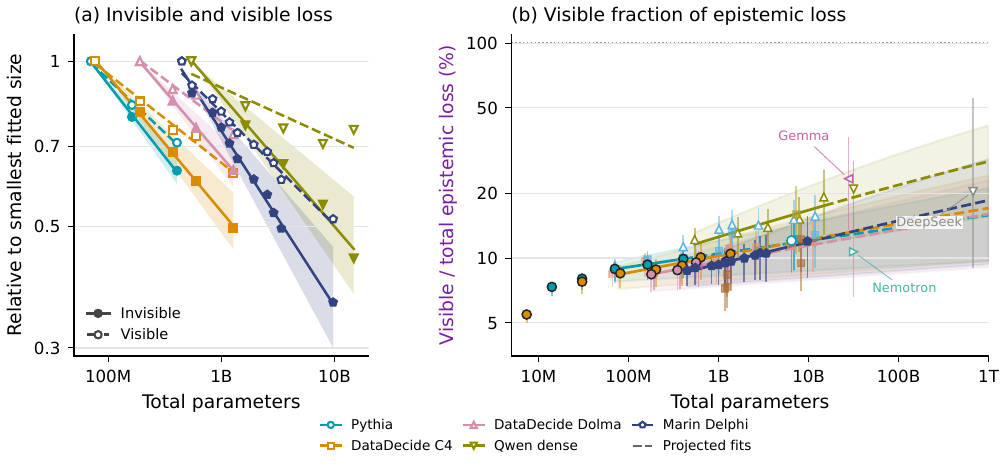}
\caption{\textbf{Visible and invisible loss on C4} (cf.\ Figure~\ref{fig:visible}).
(a) Within each family, the two components are normalized at the smallest fitted size. Solid lines show invisible loss and dashed lines show spread. (b) Invisible loss falls faster than spread in all five fitted series, so the fitted visible fraction rises with size. Shading in both panels spans refits over the C4 floor range.}
\label{fig:c4-components}
\end{preprintfigure}

\begin{preprintfigure}[htbp]
\centering\includegraphics[width=\linewidth]{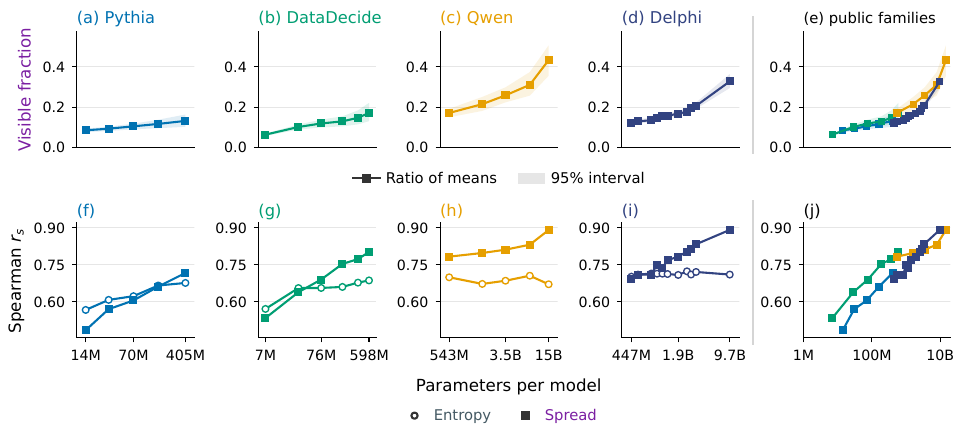}
\caption{\textbf{Context-dependent uncertainty on C4} (cf.\ Figure~\ref{fig:noise-validation}).
Pythia, DataDecide, Qwen, and Delphi share 40 held-out passages. Each family scores 32 positions per passage in its native vocabulary. The top row divides mean corrected spread by mean reference-to-member KL, with $95\%$ intervals from resampling passages. The bottom row shows Spearman correlations of entropy and spread with reference-to-member KL. References follow Table~\ref{tab:uncertainty-design}.}
\label{fig:c4-context}
\end{preprintfigure}

\begin{preprintfigure}[htbp]
\centering\includegraphics[width=\linewidth]{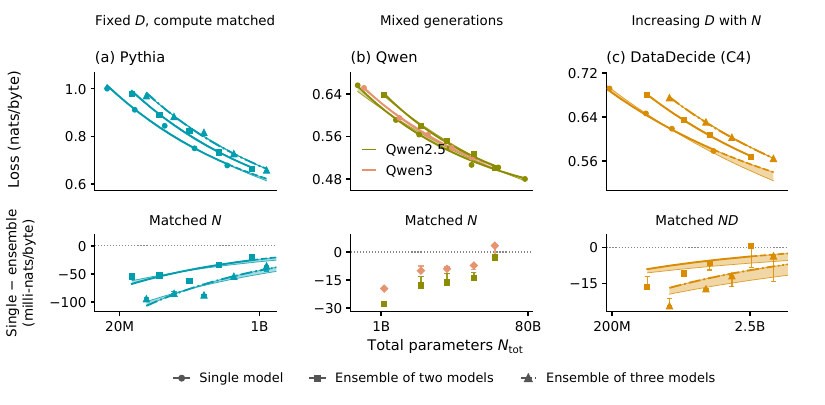}
\caption{\textbf{Ensembles and larger single models at matched budgets on C4} (cf.\ Figure~\ref{fig:economics}).
The upper panels show checkpoint and ensemble losses for Pythia, Qwen, and DataDecide. The lower panels subtract ensemble loss from fitted singleton loss at matched parameters or training compute. Curves and shading use the C4 floor range $[0.31973,0.41740]$ nats/byte, with central curves at the empirical cap.}
\label{fig:c4-economics}
\end{preprintfigure}

\end{document}